\documentclass{article}

\usepackage{PRIMEarxiv}
\usepackage{color}
\usepackage[authoryear]{natbib}
\usepackage[utf8]{inputenc} 
\usepackage[T1]{fontenc}    
\usepackage{hyperref}       
\usepackage{url}            
\usepackage{booktabs}       
\usepackage{amsfonts}       
\usepackage{nicefrac}       
\usepackage{microtype}      
\usepackage{lipsum}
\usepackage{fancyhdr}       
\usepackage{graphicx}       
\graphicspath{{media/}}     
\usepackage{xcolor}
\usepackage{longtable}
\usepackage{diagbox}
\usepackage{framed,multirow}
\usepackage{pifont}
\usepackage{makecell}
\usepackage{framed,multirow}
\usepackage{booktabs}
\usepackage{threeparttable}
\title{A Survey of Deep Learning-based Radiology Report Generation Using Multimodal Data}

\author{
  Xinyi Wang\\
  The University of Nottingham\\
  United Kingdom\\
   \And
  Grazziela Figueredo\\
  The University of Nottingham\\
  United Kingdom\\
   \And
  Ruizhe Li\\
  The University of Nottingham\\
  United Kingdom\\
   \And
  Wei Emma Zhang\\
  The University of Adelaide\\
  Australia\\
  \And
  Weitong Chen\\
  The University of Adelaide\\
  Australia\\
  \And
  Xin Chen \thanks{\textit{\underline{Corresponding author}}: 
Xin Chen, xin.chen@nottingham.ac.uk} \\
  The University of Nottingham\\
  United Kingdom\\
}

\begin{document}
\maketitle

\begin{abstract}
Automatic radiology report generation can alleviate the workload for physicians and minimize regional disparities in medical resources, therefore becoming an important topic in the medical image analysis field. It is a challenging task, as the computational model needs to mimic physicians to obtain information from multi-modal input data (i.e., medical images, clinical information, medical knowledge, etc.), and produce comprehensive and accurate reports. Recently, numerous works have emerged to address this issue using deep-learning-based methods, such as transformers, contrastive learning, and knowledge-base construction. This survey summarizes the key techniques developed in the most recent works and proposes a general workflow for deep-learning-based report generation with five main components, including multi-modality data acquisition, data preparation, feature learning, feature fusion and interaction, and report generation. The state-of-the-art methods for each of these components are highlighted. Additionally, we summarize the latest developments in large model-based methods and model explainability, along with public datasets, evaluation methods, current challenges, and future directions in this field. We have also conducted a quantitative comparison between different methods in the same experimental setting. This is the most up-to-date survey that focuses on multi-modality inputs and data fusion for radiology report generation. The aim is to provide comprehensive and rich information for researchers interested in automatic clinical report generation and medical image analysis, especially when using multimodal inputs, and to assist them in developing new algorithms to advance the field.
\end{abstract}


\section{Introduction \label{sec:Intro}}

Medical images can offer detailed information about the bodies and help physicians screen, diagnose, and monitor medical conditions without requiring invasive techniques \citep{beddiar2023automatic,liao2023deep}. Radiologists summarize the information extracted from medical imaging to radiological reports for clinical decision-making. The manual generation of reports is however labour-intensive, time-consuming, and requires extensive expertise \citep{beddiar2023automatic}. The volume of radiological imaging data is growing significantly faster than the number of trained radiologists. In some cases, each radiologist needs to analyze an image every 3 to 4 seconds during an 8-hour workday \citep{hosny2018artificial,mcdonald2015effects}. \citet{rimmer2017radiologist} revealed that 97\% of the United Kingdom radiology departments faced challenges in meeting diagnostic reporting requirements. \citet{I1topol2019deep} pointed out that the demand for medical image interpretation greatly surpasses the current capacity of physicians in the United States. This makes it challenging for radiologists to provide high-quality reports within the scheduled time. The current demand extends patient waiting time increases the risk of disease transmission \citep{beddiar2023automatic} and compromises patient care. The development of automatic report-generation techniques can help alleviate this problem.

Automatic high-quality report generation is challenging. It is intrinsically a multi-modality problem.  \citep{multimoda1tu2024towards, multimoda2yan2023multimodal}. A high-standard report demands clarity, correctness, conciseness, completeness, consistency, and coherence \citep{kaur2022methods,liao2023deep}. To meet these requirements, radiologists need to combine image data with information from other modalities, such as medical knowledge and clinical history, since a single image lacks sufficient diagnostic details. For example, the same image may yield different diagnoses depending on the associated clinical context. Ideally, automated systems should mimic this process, and experiments also show that diverse modalities can enhance report generation performance \citep{11yang2022knowledge,20li2023dynamic,126zhang2024generalist}.

\begin{figure}[!htbp]
\centering
\includegraphics[width=0.45\columnwidth]{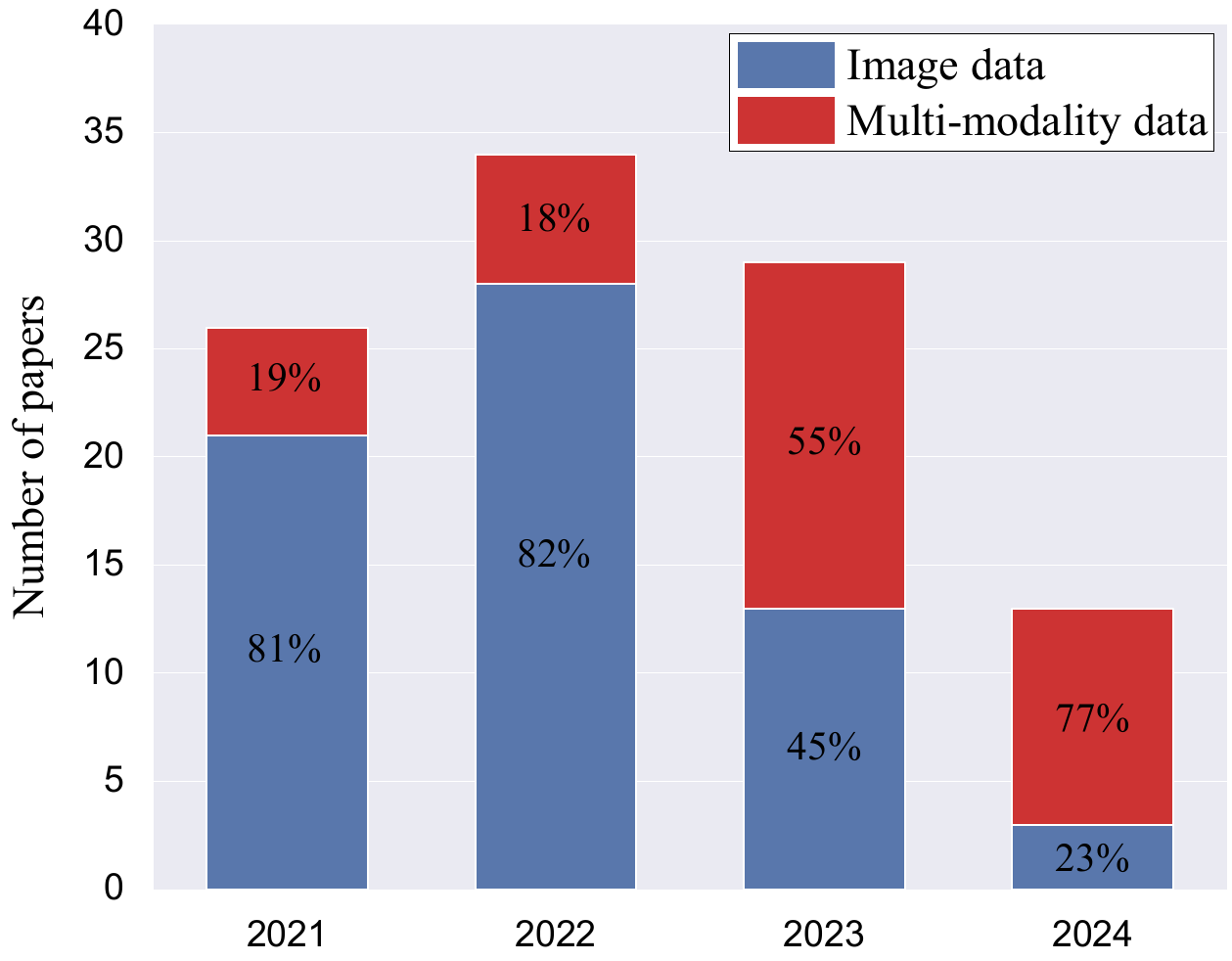}
\caption{\label{fig:Year_inputs}The distributions of reviewed papers using image data and multi-modality data as inputs per year from 2021 to 2024. The percentage denotes the input's prevalence among articles published within the year.}
\end{figure}

Mimicking this process is difficult for automated systems. First, multi-modal data are represented in distinct feature spaces with different sizes, data formats (e.g. discrete vs. continuous), and data structures (e.g. graphs, vectors, etc.)  \citep{102li2023unify}. This disparity complicates the alignment and integration of multi-modal data. Second, models need to remove redundant information while identifying and leveraging complementary information from multi-modal inputs. Third, different modalities exhibit various types of noise. Fourth, data from disparate sources contribute unevenly to report generation, necessitating that models assess and appropriately weigh these contributions. Due to these challenges, previously developed techniques mostly considered images as input, while for the past three years, multi-modality deep learning developed very rapidly. An increasing number of research papers endeavoured to emulate physicians by leveraging multi-modal data for the generation of diagnostic reports. Additionally, the recent rise of large models that utilize prompts has made the use of multi-modal inputs a mainstream trend, as shown in Figure \ref{fig:Year_inputs}.

Most of the previous surveys on this topic \citep{kaur2022methods, beddiar2023automatic, liao2023deep, shamshad2023transformers, liu2023systematic, pang2023survey} did not include non-imaging inputs. \citet{messina2022survey} considered non-imaging inputs but only included 6 papers. In total, the previous surveys included 40 to 66 papers for report generation, primarily focusing on articles published before 2022. In these studies, automated systems typically follow five key steps: multi-modality input acquisition, data preparation, feature learning, feature fusion and interaction, and report generation. Performance-improving techniques are applied at specific stages, making step-wise categorization useful for clarifying their functions. Although earlier surveys have addressed this point, they have not provided sufficient clarity. Additionally, we found that previous surveys have given limited attention to two important issues. While the application of large medical models in report generation has gained increasing recognition, current surveys lack analysis of their usage and fine-tuning. Further, interpretability and uncertainty quantification in clinical decision making remains a concern. While \citet{messina2022survey} study on explainable AI is comprehensive, there is limited survey analysis of papers published after 2022.

This survey differs from previous ones in five main contributions: 
\begin{itemize}
    \item We analyse an additional 22 papers that utilize non-image inputs and focus on the acquisition, analysis, and integration of multi-modal inputs. To the best of our knowledge, it is the first review to investigate state-of-the-art multi-modal data processing techniques for report generation. 
    \item Totally, we examine 100 papers published from 2021 to 2024 to provide a comprehensive study on novel techniques in automatic report generation, of which 90 focus on traditional deep learning methods, and 10 are based on large model approaches.
    \item We propose a general workflow for deep-learning-based report generation with a taxonomy of approaches employed, as shown in Figure \ref{fig:colorpipeline}. Table \ref{tab:overall} in Appendix A summarizes all reviewed papers.
    \item For large model-based approaches, including language models, vision models, and multimodal models, we summarize the acquisition of image-text datasets with prompts, the evolution from large language model to large multimodal model, and how to train large medical models with limited resources.
    \item We present recent advancements in explainability for report generation, including visualization, auxiliary tasks, and uncertainty quantification.
\end{itemize}

\begin{figure*}[!t]
\centering
\includegraphics[width=\columnwidth]{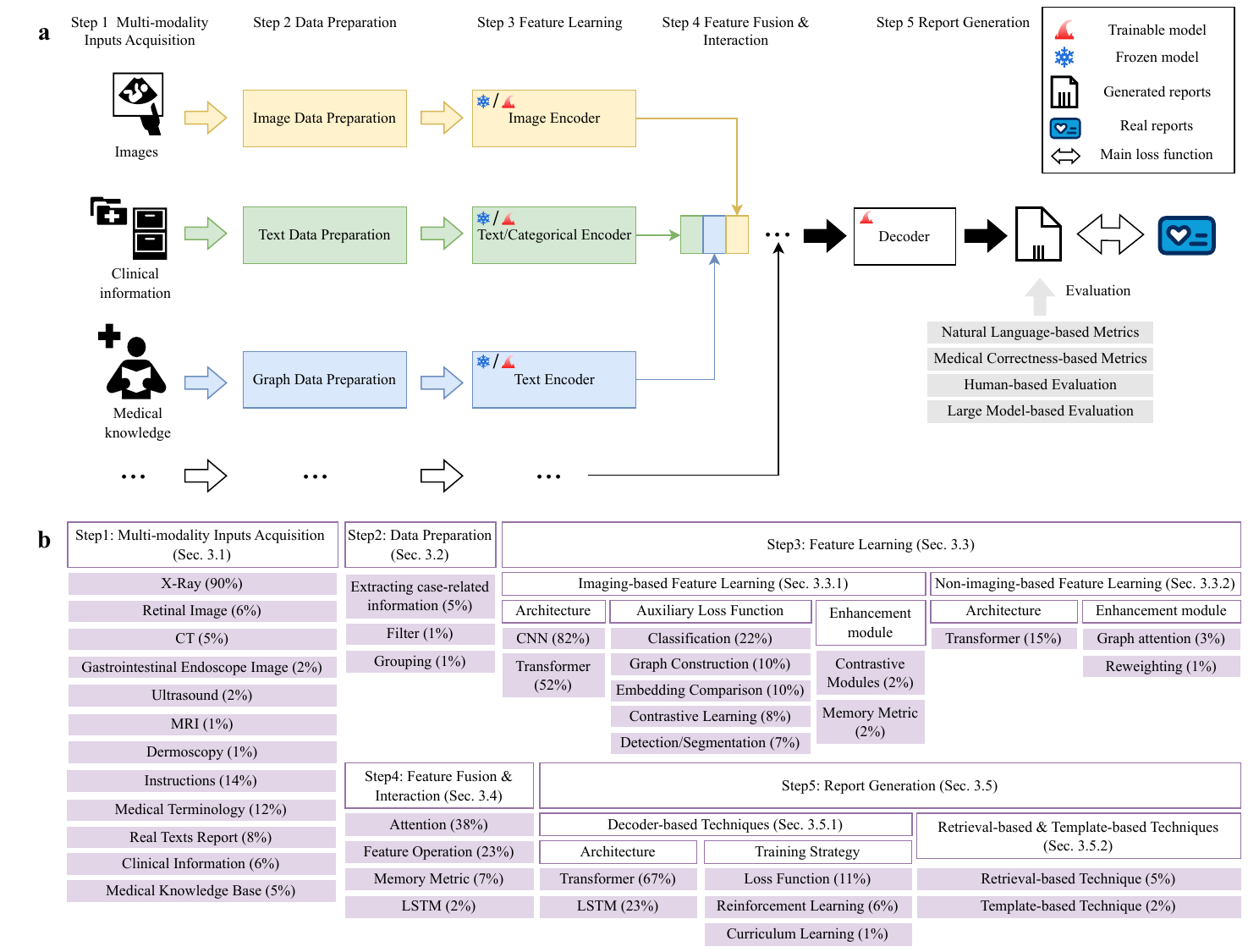}
\caption{\label{fig:colorpipeline}An overview of report generation: workflow and taxonomy of employed approaches. a. The typical workflow of automatic radiology report generation. b. The summary of the usage of techniques in the reviewed papers for each step. (x\%) represents the percentage of these articles relative to the 100 papers reviewed.} 
\end{figure*}

The remainder of the paper is organized as follows. Section \ref{sec:search&selection} introduces the paper search and selection process. Section \ref{sec:methods} first provides a workflow of report generation, then analyses the techniques in each component of the workflow. Section \ref{sec:lmm} focuses on the application of large models in report generation. Section \ref{sec:inter} discusses the latest developments in explainability. Next, in sections \ref{sec:datasets} and \ref{sec:evaluation}, we introduce popular public report generation datasets and evaluation methods, including metrics and expert evaluation. Section \ref{sec:comparison} compares the model performance of several papers in the same experimental setting. Lastly, we discuss challenges and perspectives on this topic in Section \ref{sec:challenges&futurework} and provide our conclusions in Section \ref{sec:conclusions}.

\section{Search and selection of articles\label{sec:search&selection}}
Three search engines (Google Scholar, PubMed, and Springer) and four queries were employed to collect articles. They included ``radiology report generation", (medical OR medicine OR health OR radiology) AND (report OR description OR caption) AND generation, modal AND (medical OR medicine OR health OR radiology) AND (report OR description OR caption) AND generation, and ``medical report generation". Following the searches, the titles and abstracts of each article were read briefly to identify those that met the selection criteria. If there was uncertainty, the article was included to ensure relevant studies were not omitted. The selection criteria were framed around three aspects. First, we included articles published in the years 2021, 2022, 2023, and 2024 due to the significant number of developments using multi-modal technology in recent years. We aim to focus on the latest algorithms not covered in previous surveys. Second, the studies should be original research on the automatic generation of full-text natural language radiology reports. While large model-based methods may generate only parts of reports for specific prompts, they are included in this survey because they have the potential to produce complete reports with different prompts. In contrast, methods that generate short captions of one or two sentences are excluded due to the differing nature of long reports and sentence generation. Third, papers published in journals, conferences, and conference workshop proceedings were included. Additionally, papers from arXiv in 2023 and 2024 with more than 15 citations were selected. Due to the potential impact of emerging medical large models on report generation, relevant articles from arXiv are included regardless of their  citation count.

A total of 151 papers were identified using three search engines. In addition, by tracing the ancestry and descendants of papers, we identified another 30 papers. After removing duplicates, 110 publications were retained. We thoroughly read these works and applied exclusion criteria. First, at least one of the generated languages should be English because it is the common language of academia, which ensures broader accessibility, easier evaluation, and more accurate comparisons of algorithm performance. Second, at least one of the input data should be images. Finally, 102 works were included in the following analysis. 

\section{Methods\label{sec:methods}}
Deep-learning-based radiology report generation typically follows a standard workflow summarized in Figure \ref{fig:colorpipeline} a. Overall, a basic radiology report generation framework consists of 5 steps: (1) multi-modality data acquisition (Section \ref{sec:Inputs}); (2) data preparation (Section \ref{sec:data_prepare}); (3) feature learning (Section \ref{sec:FeatureLearning}); (4) feature fusion and interaction (Section \ref{sec:fusion}); and (5) report generation (Section \ref{sec:generation}). Although traditional deep learning methods and large model-based approaches prioritize different aspects of research for automatic report generation, both adhere to the workflow shown in Figure \ref{fig:colorpipeline} a. Traditional deep learning methods focus on enhancing the effectiveness of each step, while the exploration of large models is in its early stages. Large models focus on large-scale data acquisition, the transition from large language models to large multimodal models, and effective training. This section analyzes the techniques in the 100 works based on the workflow, and the next section discusses the details of large models. We believe that effective techniques from traditional methods can inspire the future development of large models due to their similar workflows.

Typically, medical images, when analyzed with or without other types of data, are first prepared, including image normalization, resizing, etc. (step 2). Subsequently, they are inputted into feature extractors to perform feature learning (step 3), predominantly implemented using a Convolutional Neural Network (CNN) or Transformer architectures, along with auxiliary loss functions or enhancement modules. The feature extractors aim at extracting features relevant to report generation, and additional techniques are utilized to improve the expressiveness of the features. For certain approaches, a feature fusion and interaction module (step 4) is subsequently applied to align cross-modal data and mitigate the negative effects caused by differences between the visual and textual domains. After fusion and interaction, the features are input into the generator to generate the report (step 5). Throughout this process, the network parameters are generally trainable. In some cases, a frozen pre-trained encoder can be used to enhance training stability. For instance, the visual encoder in large-model-based methods \citep{127li2024llava, 119lee2023llm}. Table \ref{tab:overall} in Appendix A presents detailed information for each paper across the five steps and training strategies.

\subsection{Multi-modality inputs acquisition \label{sec:Inputs}}

The input data refers to the data received by the report generation system. During model training and inference, the data can vary; for example, both images and real reports are used as inputs during training, while only images are used during inference \citep{68shetty2023cross}. However, the model learns from the distribution and features of the training data. If the testing data changes the model will likely struggle to generalize, resulting in decreased performance. Therefore, in this section, we define and explain the acquisition of input data used consistently in both the training and testing phases of the reviewed papers.

Input data includes images, medical terminology, medical knowledge base, real text reports, clinical information, questionnaires, and medical descriptions. Figure \ref{fig:data_input} illustrates the types, examples and sources of these data. The following provides a detailed description:

\begin{figure*}[!t]
\centering
\includegraphics[width=\columnwidth]{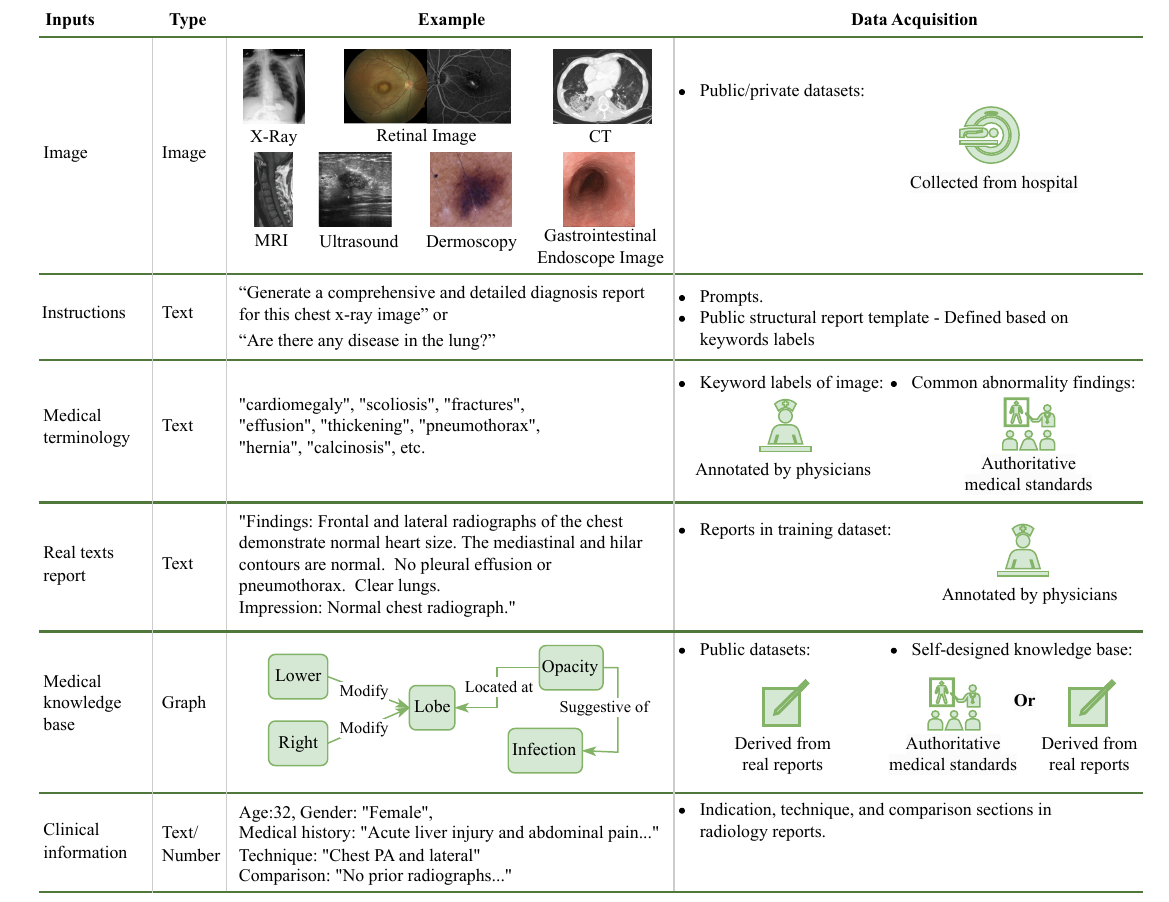}
\caption{\label{fig:data_input}A summary of the types, examples, and sources of multi-modal input data.}
\end{figure*}
\begin{itemize}
    \item \textit{Image data} includes X-ray, magnetic resonance imaging (MRI), computed tomography (CT), ultrasound, gastrointestinal endoscope image, retinal image, and dermoscopy image. Most papers we reviewed focus on generating medical reports for chest X-ray images (82 works). Other than chest diagnosis, retinal image is the second most prevalent image modality (6 works). The retinal image includes lots of categories, e.g., fundus fluorescein angiography and color fundus photography. Other works focus on chest CT (5 works), gastrointestinal endoscope image (2 works), spine MRI (1 work), dermoscopy image (1 work), and breast ultrasound images (1 work).  Although some images are non-radiological, such as ophthalmic images, we still include their report-generation techniques in this survey.
    \item \textit{Instructions} refer to the explanations provided for completing a task or carrying out an operation. It comprises prompts and questionnaires. Prompts like ``Generate a comprehensive and detailed diagnosis report for this chest x-ray image'' are commonly used with large models for report generation  \citep{86tanwani2022repsnet,118pellegrini2023radialog,119lee2023llm,120wang2023r2gengpt,122hyland2023maira,123bannur2024maira,124chen2024chexagent,125guo2024llavaultra,126zhang2024generalist,127li2024llava,129alkhaldi2024minigpt}. Questionnaires are sets of questions designed for report generation \citep{86tanwani2022repsnet, 108pellegrini2023rad}. \citet{108pellegrini2023rad} developed a structured report dataset for Chest X-rays, containing hundreds of questions, in which reports are generated by answering a series of questions, similar to the process of hierarchical visual question answering.
    \item \textit{Medical terminology} refers to medical terms and expressions, which are from keyword labels of images \citep{71huang2021deep,18huang2021deepopht,62huang2022non, 106liu2023observation}, or self-built corpora \citep{1liu2021exploring,34cao2022kdtnet,67cao2023cmt,99xue2024generating,105gu2024complex, 110li2023enhanced}, storing common descriptions found in medical reports. Incorporating medical terminologies can assist the model in generating descriptions that are relevant to medical terminology. \citet{106liu2023observation} further developed a concepts library for medical descriptions based on various terminologies, providing detailed information on symptoms and health status. For example, the term ``Cardiomegaly'' corresponds to ``The heart size is enlarged''.
    \item \textit{Real texts report} refers to radiology reports written by physicians. In the reviewed papers, two methods are used to obtain these reports. The first retrieves them from the training dataset to avoid overlap with the test set, as detailed in Section \ref{sec:data_prepare}. It mimics the behaviour of radiologists referencing similar prior cases when writing reports. The second uses prior reports, referring to the patient's historical data \citep{123bannur2024maira}. As described in Section \ref{sec:data_prepare}, utilizing this data enables monitoring of temporal changes, which is essential for clinical applications.
    \item \textit{Medical knowledge base} mostly records the connections between different organs and diseases, and is presented in a graph format. A graph is a fundamental data structure consisting of a set of nodes and edges, which can easily represent a set of subjects and their connections. A knowledge graph can primarily be obtained in two ways: (1) public datasets such as the RadGraph \citep{Radgraphjain2021radgraph} used by two works \citep{11yang2022knowledge,20li2023dynamic}. Recently, an enhanced version of RadGraph, termed RadGraph-XL \citep{delbrouck2024radgraph}, was introduced, offering improved performance. This update extends its applicability to a broader range of medical imaging modalities (i.e., chest CT, abdomen/pelvis CT, brain MRI, and chest X-ray), anatomical regions, and data sources; (2) self-designed knowledge bases according to authoritative medical standards \citep{21huang2023kiut, 35xu2023vision} or real reports in training dataset \citep{69jia2022few}. Utilizing a knowledge base allows the model to learn medical knowledge akin to a physician.
    \item \textit{Clinical information} encompasses patient demographics (e.g., age and gender), clinical observations, and medical histories. By incorporating clinical information, the model can obtain rich clinical context, which helps it better understand the specific situation and background of the patient. Clinical information is embedded in the indication, technique, and comparison sections of radiology reports (see Figure \ref{fig:DataSample.}). While most studies utilize only the indication section \citep{17zhou2021visual,72dalla2022multimodal,109dalla2023controllable,113dalla2023finding,122hyland2023maira}, MAIRA-2 \citep{123bannur2024maira} leverages all three sections, stating that the technique and comparison sections provide additional information, and demonstrates their positive impact through experiments.
\end{itemize}

\subsection{Data preparation \label{sec:data_prepare}}
Data preparation endeavours to enhance data quality and prepare it for model deployment, typically encompassing data cleansing, transformation, and organization. Conventional data preparation techniques primarily target images and text. Image processing typically involves resizing and cropping, while text processing includes case normalization, removal of non-alphabetic tokens, and tokenization. The methods utilized in each paper are outlined in Table \ref{tab:overall} in Appendix A, but it is worth noting that conventional preparation methods are so ubiquitous that some papers do not mention them. While the information is not recorded in the table, it does not mean the absence of a data preparation process.

Novel data preparation methods in the reviewed papers can be categorized into filtering \citep{12ramesh2022improving}, grouping \citep{82wang2022inclusive}, and case-specific information extraction. \citet{12ramesh2022improving} argued that writing a radiology report necessitates referencing historical information, which inevitably included descriptors such as `again' and `decrease'. However, these terms cannot be inferred from a single image, therefore \citet{12ramesh2022improving} filtered such descriptions in the reports. This exclusion was found to facilitate the model's learning process; however, simply modifying ground truth reports does not accurately reflect clinical practice. In some clinical settings, monitoring temporal changes is essential for the effective diagnosis and management of various diseases, such as pneumonia \citep{kalil2016management, ImaGenomewu2021chest}. At the same time, follow-up reports are often shorter than those for the initial image, focusing only on changes or new findings (e.g., ``Otherwise, little change.''). Without historical context can result in incomplete image descriptions. Current publicly available datasets are typically organized by subject identifiers and study identifiers, representing different examinations for the same patient and arranging them chronologically \citep{MIMICjohnson2019mimic, ImaGenomewu2021chest, chambon2024chexpertplusaugmentinglarge}. Analyzing a sequence of images or inputting prior reports into the model might be more appropriate \citep{123bannur2024maira, 109dalla2023controllable}. Another approach is to train using only the initial image-report pairs to maintain data validity, though this would limit the model’s applicability and significantly reduce the amount of available training data. 

Grouping refers to organizing sentences from ground truth reports into distinct sections, typically relying on keywords from pre-defined knowledge graphs and filtering rules. Each section describes a specific anatomical structure. Grouping aims to enable the generation system to process various types of sentences differently. For instance, \citet{82wang2022inclusive} employed different decoders to generate descriptions for different anatomical structures. Alongside the reviewed papers, a recently released public dataset named ImaGenome \citep{ImaGenomewu2021chest} also includes grouping results in their annotation files (see Section \ref{sec:datasets}). The grouping result becomes more easily accessible.

\begin{figure*}[!t]
\centering
\includegraphics[width=0.9\columnwidth]{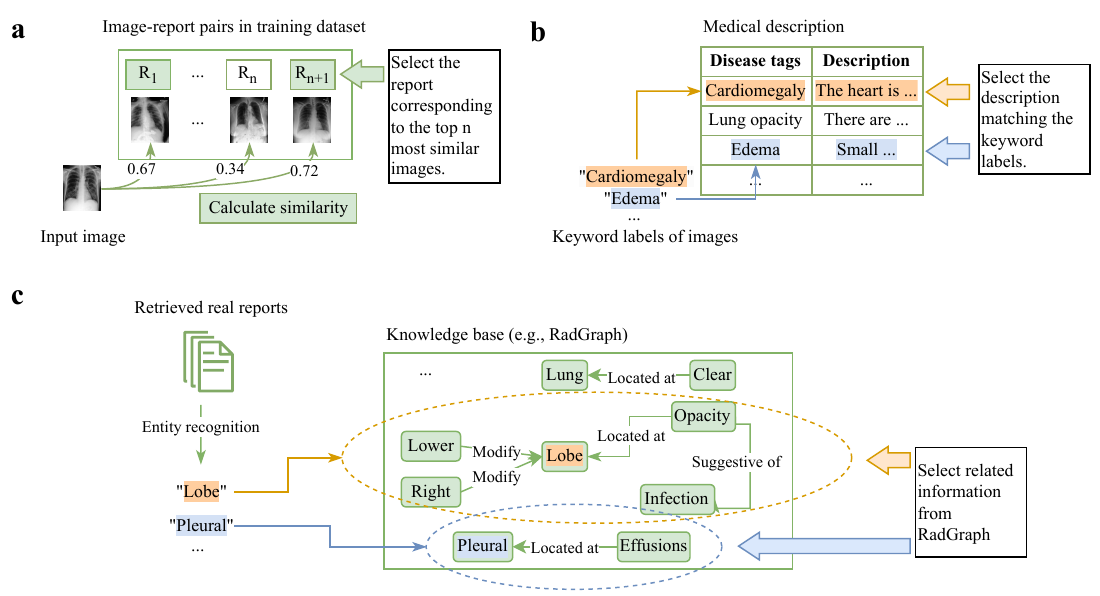}
\caption{\label{fig:data_comb}Three typical  methods for extracting case-related information. (a) Extracting relevant textual reports from the training set by comparing the similarity between the input image and images in the training set. (b) Extracting relevant medical descriptions from a concepts library based on the image's annotated keywords. (c) Automatically extracting entities (e.g., anatomy) from the reports retrieved in (a) and using these entities to search the knowledge base for related knowledge graphs. ``...'' refers to concepts such as these. The figure only lists a few examples from the training dataset, keyword labels of images, medical descriptions, recognized entities in the report, and knowledge base.}
\end{figure*}

When using medical knowledge bases and real text reports, inputting the entire knowledge base or all reports from the training dataset often leads to information redundancy, making it difficult for the network to extract relevant information. Extracting case-specific information can alleviate this problem. Figure~\ref{fig:data_comb} illustrates three highly effective methods in reviewed papers.

The first method is the retrieval of related real text reports \citep{1liu2021exploring,66song2022cross,11yang2022knowledge,20li2023dynamic,106liu2023observation,110li2023enhanced,112jin2024promptmrg}.  This process mimics radiologists consulting previous medical case reports when drafting their own. By calculating the similarity between the input image and the images in the training set, we select the reports corresponding to the top $N$ most similar images and use these reports as inputs. The second method is the retrieval of related medical descriptions. \citet{106liu2023observation} obtained medical description from a pre-defined concepts library based on input terminologies. However, this method still depends on disease tags from radiologists for high performance,  increasing their workload. 

The third method combines retrieved reports in the first method with a medical knowledge base (e.g., RadGraph) to extract case-related medical knowledge \citep{11yang2022knowledge,20li2023dynamic}. An automated tool first extracts entities from the reports, and then all related information from the knowledge base is retrieved as input. Alternatively, \citet{110li2023enhanced} use the annotation tool in RadGraph to extract entities and positional information from retrieved reports. 

Additionally, two studies \citep{35xu2023vision, 21huang2023kiut} used image classification results to process self-built graphs and extract case-related information. This method is less effective than the three mentioned above, partly due to inherent classification errors. Improving classifier performance could enhance the quality of report generation.

\subsection{Feature learning \label{sec:FeatureLearning}}
\subsubsection{Image-based feature learning \label{sec:FeatureLearning-Imaging Data}}
Previous research primarily utilized CNNs as architectures for extracting image features, however, recently, an increasing number of researchers have opted for the use of Transformers due to their improved performance. Simultaneously, numerous studies proposed novel modules to enhance the model capability. In this section, the model architectures are first introduced and subsequently enhancement modules are described. These modules include auxiliary tasks, contrastive learning, and memory metrics. The architecture and modules utilized in each paper are outlined in Table \ref{tab:overall} in Appendix A.

\paragraph{Architectures for image feature extraction}

\begin{figure}[!h]
\centering
\includegraphics[width=0.45\columnwidth]{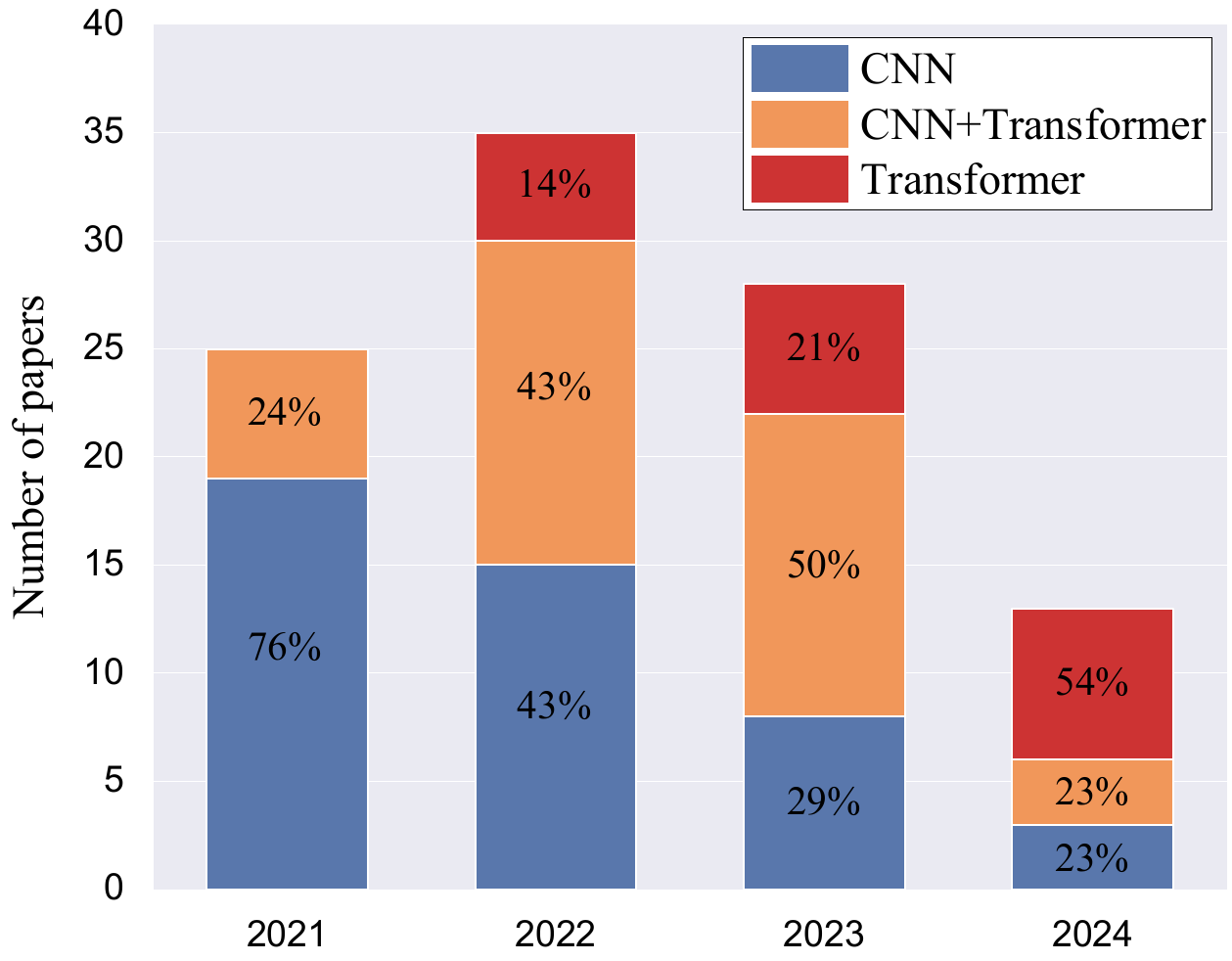}
\caption{\label{fig:Year_encoder} The statistics of the reviewed papers using different architectures to extract image features per year from 2021 to 2024. The percentage denotes the method's prevalence among articles published within the year.}
\end{figure}

The statistics of the architectures used as image feature extractors are shown in Figure \ref{fig:Year_encoder}. It shows a clear trend of more studies adopting Transformer-based architectures for image feature extraction. However, unlike the decoder architecture in Section \ref{sec:decoder}, where the Transformer has entirely replaced the Long Short-Term Memory (LSTM), many studies combine CNNs with Transformers, using CNNs for image processing and feeding the extracted features into the Transformer encoder. However, there is a lack of research comparing this approach with using only CNNs or only Transformers for report generation.

Most studies developed CNN architectures based on classical models, such as the Residual Nets (ResNet) \citep{Resnethe2016deep} (42 papers) and the Dense Convolutional Network (DenseNet) \citep{Densenethuang2017densely} (22 papers). To improve model performance, ten works modified the CNN structure by attention modules, which assigned varying degrees of importance (weights) to different parts of the input by learnable parameters, allowing the model to selectively focus on specific regions of an image, such as small localized lesions. Specifically, the core of CNNs lies in the convolutional kernels, which extracts informative features by  processing spatial and channel information within local receptive fields. The introduction of attention mechanisms is based on the assumption that the model requires a mechanism (i.e., attention) to adaptively recalibrate channel or spatial features by explicitly modeling their interdependencies. This enables the model to focus more effectively on the most relevant features. Traditional attention mechanisms can be classified into channel-wise \citep{93du2022automatic,92wang2022medical,80gajbhiye2022translating,16pahwa2021medskip} and spatial-wise \citep{16pahwa2021medskip,53jia2021radiology}, which allocate different weights to the various channels and spatial positions of the inputs respectively. In addition, \citet{14li2023auxiliary} and \citet{98wang2024camanet} utilized the idea of the class activation map \citep{zhou2016learning} to obtain weights. \citet{74yan2022prior} initially extracted image patch features, clustered them using an unsupervised method, and then weighted the cluster results.  Experimental results show that the attention mechanism allows models to pay more attention to the lesions than to irrelevant backgrounds. With the rise of the Transformers \citep{vaswani2017attention}, multi-head attention has become a potent method for information interaction. In clinical practice, it is widely acknowledged that radiologists report with varying granularity in radiology reports to reflect different degrees of pathological changes \citep{101wang2023self}. Therefore, both global and local information are essential for report generation, with varying importance when generating different content. Dynamic interactions between these elements can facilitate adjustments in their significance, thereby enhancing outcomes. \citet{101wang2023self} extracted regions of interest from the frontal view and then employed multi-head attention to fuse information between the frontal and lateral views with the regions.

For Transformer architectures, most of them leverage a standard Transformer encoder, while two studies \citep{102li2023unify, 119lee2023llm} noted that aligning images with text is challenging due to the continuous nature of images and the discrete nature of text. To address this, they used pre-trained models \citep{ramesh2021zero, esser2021taming} to generate discrete visual tokens. However, despite the ablation study validation \citep{102li2023unify}, the results were still inferior to other report generation methods, likely due to information loss during discretization. Additionally, the pre-trained models were trained on non-medical images, and this domain gap can also impact performance. Other works improved the model performance by modifying the self-attention module. Three works \citep{92wang2022medical,55lin2023contrastive,10wang2023metransformer} added high-order interactions among the query, key, and value of the Transformer attention module to enhance their interactions for extracting more complex information. Two works \citep{7miura2021improving,92wang2022medical} were inspired by the memory-augmented attention \citep{M2Transcornia2020meshed} and extended the keys and values with additional plain learnable vectors to record more information. \citet{10wang2023metransformer} modified the encoder by including additional input tokens. These tokens were named `expert tokens' to emulate the ``multi-expert joint diagnosis” methodology. Ablation studies indicate that these designs enhance the results significantly.

In conclusion, while an increasing number of studies employ Transformer-based architectures for feature extraction, direct comparisons with CNNs remain absent. Most architectural enhancements focus on modifications to attention mechanisms, which have demonstrated effective improvements. Some research explored the discretization of visual features to align with text features; however, it yielded limited success.

\paragraph{Auxiliary loss functions for image feature extraction\label{par:featurelearning_auxiliary_task}}

Auxiliary loss functions aim to provide additional supervision signals to the feature extractor, enabling it to extract information relevant to report generation from images. Generally, the model optimizes the main and auxiliary loss functions simultaneously through a weighted sum, with the weights reflecting the relative importance of each task. The auxiliary loss functions described in this section are used to constrain the features extracted by the visual encoder. Additionally in this section, we include losses that are contributed to the pre-training phase, such as those from contrastive learning. Although these losses do not work directly with the main loss function, they still serve as auxiliary losses to help build a more effective visual encoder. These tasks mainly include classification (22 papers), graph construction (10 papers), embedding comparison (10 papers), contrastive learning (8 papers), and detection/segmentation (7 papers). Each of them is introduced in detail below. 

\textbf{Classification loss function:} The most common auxiliary task used in the reviewed papers is classification referring to assigning images to predefined categories. Ensuring the generated radiology reports contain essential medical facts is crucial. Introducing an auxiliary classification loss helps the model focus on extracting features relevant to these facts during feature extraction. Predefined categories are usually disease labels \citep{24liu2021medical,17zhou2021visual,51wang2022automated,37yang2023radiology,22zhang2023semi,98wang2024camanet,112jin2024promptmrg,47hou2021automatic,115hou2023organ}, which are provided by auto annotation tools (e.g., CheXpert \citep{CheXpertirvin2019chexpert} and CheXbert \citep{CheXbertsmit2020combining}), typically covering 14 categories of diseases (e.g., atelectasis). For instance, \citet{37yang2023radiology} utilized the CheXpert labeling tool to annotate 14 common chest radiological findings and integrated a classification layer following the visual encoder to predict diseases. The model was optimized by minimizing the binary cross-entropy loss between the true labels and the predicted disease outcomes. Some researchers argue that these categories are insufficient to capture all necessary medical facts for report generation, and thus extend them to include other medical tags \citep{80gajbhiye2022translating,96kaur2022cadxreport,92wang2022medical,47hou2021automatic,93du2022automatic,50you2022jpg,8alfarghaly2021automated}, such as anatomical structures (e.g., abdominal) and pathological signs (e.g., cicatrix). The medical tags are provided by manual annotations, e.g., the Medical Subject Headings annotation (MeSH) for IU-Xray dataset, or auto annotation tools, e.g., the Medical Text Indexer web Application Programming Interface \footnote{\url{https://ii.nlm.nih.gov/MTI/index.shtml}} and RadGraph \citep{Radgraphjain2021radgraph}.  Compared to disease labels, medical tags offer a more comprehensive range of information. However, to the best of our knowledge, no literature supports the superiority of medical tags over disease labels. Perhaps due to the extensive scope covered by medical tags, deep learning models face challenges in acquiring such rich knowledge. \citet{17zhou2021visual} incorporated 32 additional labels for lesion location, size, and shape (e.g., ``upper/lower” and  ``patchy”) into the disease label set, observing a slight improvement in model performance.

Other notable categories used in the reviewed papers include matching status \citep{20li2023dynamic}, local properties \citep{43yang2021automatic}, report cluster results \citep{46li2022self}, and fix answer categories \citep{86tanwani2022repsnet}. \citet{20li2023dynamic} proposed a binary classification task to predict the image-report matching status. The image features, extracted by the image encoder, are input into a cross-attention mechanism alongside the reports. The output then passes through a linear layer to predict the matching probability, which is supervised with a cross-entropy loss. The objective of this task is to align the visual features more closely with the text representation. \citet{43yang2021automatic} devised localized property labels for breast ultrasound images, such as tumor morphology, to facilitate the identification of properties that are challenging to be discerned in low-resolution images. \citet{46li2022self} first conducted unsupervised clustering on the ground truth report, subsequently utilizing the resultant clusters as labels. The study validated the independence of the clustering results to ensure the diversity of the generated themes. The ablation study confirmed the effectiveness of these methods, yet the final results were inferior to those achieved using more direct additional supervisory information, such as disease labels and image-text matching. \citet{86tanwani2022repsnet} considered the report generation as a question-answer task, and the classifier was designed for fixed answer categories. 

In conclusion, the classification loss function helps the model focus on extracting medically relevant features by providing additional supervision based on medical facts. Other classification tasks include image-text matching, which aligns textual and visual features; local property classification, which aids in identifying details in low-resolution images; and question-answering tasks. Providing more direct disease-related supervision is beneficial for improving results, however, experimental results revealed that overly abundant information may complicate the network's learning and information extraction, potentially diminishing its effectiveness. In addition, for a detection auxiliary task, classifiers need to be applied to identify attributes of detected regions (e.g., `right lung'). This paragraph eschews such cases to circumvent redundancy. 

\textbf{Graph construction loss function:} Graph construction aims to integrate prior knowledge into the report generation process, encompassing key organs, diseases, and their interrelations. It enables the network to extract nodes' features from the image and defines their relationships (i.e., edges) as parameters in graph convolutional networks. A classical method was proposed by \citet{3zhang2020radiology} and yielded promising outcomes. A knowledge graph was constructed firstly based on insights provided by domain experts, where nodes represented major abnormalities and major organs, and bidirectional connections linked nodes that were related to each other. To initialize node features, a spatial attention module was introduced after the CNN backbone using 1$\times$1 convolution layers and softmax layers. The number of channels matched the number of nodes. The nodes' initial embedding was derived as attention-weighted feature maps. Then graph convolution layers were employed to disseminate information throughout the graph, followed by two branches for classification and report generation. First, the classification branch was trained, and subsequently, parameters in both the CNN backbone and the graph convolution layers were frozen, only the report generation decoder was trained. Six works \citep{1liu2021exploring,2liu2021auto,34cao2022kdtnet,30wang2022prior,42yan2022memory,22zhang2023semi} utilized this method \citep{3zhang2020radiology}.  \citet{30wang2022prior} expanded the graph \citep{3zhang2020radiology} by incorporating information from a radiology terms corpus named Radiology Lexicon (RadLex)\footnote{\url{http://www.radlex.org/}} \citep{RadLexlanglotz2006radlex}. As the number of graph nodes increased, the model performance initially improved, peaked at 40 nodes, and then declined, with a noticeable decrease at 60 nodes. \citet{2liu2021auto} constructed a large graph based on the Medical Information Mart for Intensive Care Chest X-ray (MIMIC-CXR) dataset. The nodes represented frequent clinical abnormalities and the edges represented the co-occurrence situation of different abnormalities. In addition, \citet{14li2023auxiliary} used the disease prediction results to obtain the node features. The nodes were classification probabilities, with learnable edge weights. 

Another graph reconstruction method aims at reconstructing triplets that are in the form of (entity1, relationship, entity2), such as (opacity, suggestive of, infection). \citet{72dalla2022multimodal} proposed a triplet extraction method for chest X-rays, where entities were first extracted using RadGraph \citep{Radgraphjain2021radgraph} and then merged via ScispaCy \citep{neumann2019scispacy} for simplified annotation. \citet{31li2022cross} used an AI-accelerated human-in-the-loop method for entity recognition and relation linking in ophthalmic images \citep{wu2021ai}. Three works \citep{72dalla2022multimodal,113dalla2023finding,31li2022cross} firstly predicted the triplets and then generated reports based on them. Triplet prediction is supervised by cross-entropy loss in token prediction. The experimental results indicate that integrating the extracted triplet with image features in the subsequent network significantly enhances performance \citep{31li2022cross, 113dalla2023finding}. The manually constructed graphs mentioned above are highly accurate and pre-defined, allowing the model to mimic human learning from textbooks. However, these graphs are specific to particular diseases; for instance, a graph constructed for chest X-rays cannot be directly applied to retinal diagnostics. Furthermore, the need for manual annotations increases physicians' workload. To address these limitations, \citet{37yang2023radiology} propose replacing pre-defined graphs with a learnable memory metric (see Section \ref{sec:fusion}).

In conclusion, utilizing graph convolutional networks is a mainstream approach for knowledge integration during training, whereas triplet generation is infrequently utilized. Both methods depend on pre-defined information, leading to limited scalability. 

\textbf{Embedding comparison loss function:} Embedding comparison refers to constraining the consistency of different features in intermediate layers, thereby guiding the learning process. The comparison in reviewed papers is mainly applied between features extracted from images and real reports \citep{29najdenkoska2021variational,29_2najdenkoska2022uncertainty,17zhou2021visual,9yang2021joint,90chen2022vmeknet,25wang2021self, 51wang2022automated,37yang2023radiology}, aligning learned visual features with supervised textual features in the same space to facilitate subsequent report generation. Experimental results show that the supervision signals from real text enable the extracted visual features to carry richer semantic information, facilitating more effective translation into radiology reports. Four works \citep{25wang2021self, 17zhou2021visual,51wang2022automated,37yang2023radiology} utilized a triple loss function to compel the image-text paired features to be closer to a latent space than the unpaired ones. \citet{29najdenkoska2021variational,29_2najdenkoska2022uncertainty} inspired by the Auto-Encoding Variational Bayes \citep{kingma2013auto}. They used real reports to obtain a latent space during training and generated reports based on this space. The image extractors were enabled to capture features from images that closely resemble those found in real reports. Other two works \citep{9yang2021joint,90chen2022vmeknet} used the Term Frequency and Inverse Document Frequency (TF-IDF) to extract important information from real reports as supervision signals. TF-IDF is a statistical measure assessing a word's importance by considering its frequency in a specific document and its rarity across the entire document set. 

In addition, to produce reports for abnormalities not seen during training, \citet{48sun2022lesion} initially linearly projected visual features to semantic features, and extracted semantic features of labels by the pre-trained Bidirectional Encoder Representations from Transformers for Biomedical Text Mining (BioBert) model \citep{BioBertlee2020biobert}. Consistent constrain was applied between two similarities: 1) the similarity between pairwise elements in the semantic features from visual features; and 2) the similarity between the semantic features from visual features and the semantic features from labels. This operation ensures the similarity between visual features and label features, aligning them while preserving medical facts within the visual features. \citet{22zhang2023semi} integrated semi-supervised learning into report generation using teacher-student networks. They first applied different types of noise to an input image to create two variations, which were then fed into the two networks. An auxiliary loss function was employed to ensure consistency in the extracted visual features.

In conclusion, using information from real text reports as supervision signals is highly effective for feature extraction. Visual features that are aligned with the text and rich in semantic content can enhance subsequent report generation. Additionally, embedding-based approaches can inspire future research. For example, if large models demonstrate superior performance, a teacher-student framework could enable smaller models to replicate similar learning outcomes.

\textbf{Contrastive learning loss function:} Contrastive learning is a self-supervised learning method, which allows models to minimize the distance among positive pairs and maximize it for negative ones. Contrastive learning mainly leverages the similarities and differences between samples other than annotations, facilitating the extraction of stronger feature representations. It can be used for image extractor pretraining \citep{92wang2022medical,55lin2023contrastive,84wu2022multimodal,107wang2023fine} or to optimize the model in conjunction with the main loss (i.e., report generation) \citep{86tanwani2022repsnet,33wang2022cross,20li2023dynamic, 106liu2023observation}. For pretraining,  \citet{55lin2023contrastive} utilized a classical contrastive learning method named Momentum Contrast \citep{he2020momentum}, where different views or augmented versions of the same image were considered as positive pairs. Another representative contrastive learning work is the Contrastive Language-Image Pre-training (CLIP) model \citep{CLIPradford2021learning}. It connects textual and visual information by directly training on a vast dataset consisting of image-text pairs. \citet{92wang2022medical} directly employed it for image feature extraction and \citet{84wu2022multimodal} applied the idea of CLIP to train the feature extractors on the training dataset.  \citet{107wang2023fine} however argued that previous works treated th e entire report as input, overlooking the distinct information contained within individual sentences. This oversight could result in incorrect matching of image-text pairs. Therefore, they proposed phenotype-based contrastive learning. This method involved randomly initializing a set of vectors as phenotypes, allowing sentences and visual embeddings to interact with them, and finally conducting contrastive learning between the processed embeddings. The results outperformed previous contrastive learning methods in report generation.

When contrastive learning be part of the training loss, it can be applied between visual and textual features (i.e. image-text pairs) \citep{86tanwani2022repsnet,20li2023dynamic,106liu2023observation}, or be applied based on labels, treating samples with shared labels as positives and those without any common labels as negatives \citep{33wang2022cross}. Ablation studies showed a significant improvement when using contrastive learning as an auxiliary loss function.

In conclusion, the additional supervisory signal from contrastive learning arises from variations and similarities among distinct samples. It can be employed for both pre-training and training phases. Given the rich information in medical images and reports, an image pair (or an image-text pair) can form positive pairs for one medical fact while serving as negative pairs for another. Thus, the effective design of positive and negative samples is crucial. Currently, sentence-level matching has demonstrated strong performance for image-text pairs, and utilizing image labels is an effective approach for image pairs.

\textbf{Detection/segmentation loss function:} Object detection locates and identifies objects or patterns within an image, focusing on determining their presence and position. Segmentation divides an image into meaningful regions by identifying and separating objects based on specific characteristics. Both processes improve the model's comprehension of medical images through object recognition and region extraction. Abnormalities in medical images are often confined to small, localized regions. Supervision by detection or segmentation auxiliary loss enables backbones to yield dense features specific to these regions, rather than the sparse features typically extracted. The detected or segmented regions can be anatomical regions \citep{13tanida2023interactive,113dalla2023finding,109dalla2023controllable, 101wang2023self, 19han2021unifying, 105gu2024complex} and abnormal regions \citep{48sun2022lesion}. Detection is commonly applied to X-ray imaging using the public Chest ImaGenome dataset, while segmentation is only used for spine MRI on a private dataset. The ImaGenome provides atlas-based detection annotations, which offer bounding box labels for anatomical structures (see Section \ref{sec:datasets}). This makes them easier to acquire than segmentation annotations.

Overall, auxiliary loss functions operate in three ways. The primary approach provides disease-related supervisory signals (e.g., classification, graph construction, and detection) to facilitate the model's identification of specific lesions. The second approach aligns image data with relevant textual information (e.g., embedding comparison), enriching image features with semantic information to make decoding into a report easier. The third approach is contrastive learning, which leverages image similarities and differences to extract more robust features. The careful design of the sample pairs is crucial. In addition, the outputs of auxiliary tasks supervised by auxiliary loss function can provide valuable information such as disease labels, therefore, entering them into the following generation network is a common choice \citep{8alfarghaly2021automated, 47hou2021automatic,87singh2021show,43yang2021automatic,6you2021aligntransformer,17zhou2021visual,93du2022automatic,69jia2022few,96kaur2022cadxreport,48sun2022lesion,33wang2022cross,92wang2022medical,74yan2022prior,50you2022jpg,13tanida2023interactive, 31li2022cross,112jin2024promptmrg}. For example, \citet{17zhou2021visual} sent the semantic word embeddings of the predicted findings from the classifier to the report generation decoder.

\paragraph{Enhancement modules for image feature extraction\label{enhance}}
Designing enhancement modules involves creating components to improve the interaction of specific information, thereby enhancing feature extraction efficiency. In the reviewed papers, two main types of modules are employed. Firstly, contrastive attention guides the model's focus on specific regions. \citet{52ma2021contrastive} designed a contrastive attention model to extract abnormal region features by comparing input samples with normal cases. Similar features shared between the input and normal cases were subtracted from the input image feature, and the remaining feature was then concatenated with the original feature. \citet{66song2022cross} argued that \citet{52ma2021contrastive} did not consider historical similar cases; therefore, they proposed a module based on similarity retrieval technique to obtain similar images from the training dataset. Unique information in the image is emphasized by enlarging different features between inputs and the  retrieved images.

The second is memory metric. Using a memory metric for image feature extraction assumes the presence of similar features in various medical images. Memory metrics are employed to record and transmit the similarity information during training \citep{90chen2022vmeknet,42yan2022memory}. Typically, an $n \times n$ matrix is randomly initialized, where $n$ represents the number of metric rows. Then, at each training step, the matrix is updated based on the visual features and the previous metrics. 

\subsubsection{Non-imaging-based feature learning \label{sec:FeatureLearning-Nonimage}}

The types of non-image data include text, graph, and numerical data (see Figure \ref{fig:data_input}). Numerical data is typically fed into the network in the next step after one-hot encoding, without a specifically designed encoder. Therefore, in this section, we focus on feature extraction for text and graph inputs. 

After tokenization mentioned in Section \ref{sec:data_prepare}, text data is divided into basic units (such as words, subwords, or characters). Then the general embedding technique (e.g., lookup table) is applied to convert these tokens into continuous vector representations. Next, many studies directly fuse these vectors with image features, while others incorporate a text encoder to enhance the representational capabilities of textual features. Transformer-based models and their variants are mainstream methods, widely applied to terminology \citep{24liu2021medical,34cao2022kdtnet,67cao2023cmt,1liu2021exploring,99xue2024generating,106liu2023observation,110li2023enhanced}, real text reports \citep{1liu2021exploring, 106liu2023observation,110li2023enhanced,112jin2024promptmrg}, and questionnaires \citep{86tanwani2022repsnet,108pellegrini2023rad}. These models are typically trainable, and some studies also utilize frozen models that are pre-trained on medical datasets to extract clinical features more effectively. For example, \citet{110li2023enhanced} utilized the Clinical Bidirectional Encoder Representations from Transformers (ClinicalBERT) \citep{alsentzer2019publicly} to extract clinical concept features. There is no experimental evidence proving that the introduction of pre-trained models enhances performance; however, reducing the number of trainable parameters can accelerate convergence and facilitate training. In addition, in medical reports, there is an imbalance between normal and abnormal descriptions, which may cause the model to prioritize normal descriptions. To address this issue, \citet{110li2023enhanced} employed TF-IDF to re-weight the input terminology embeddings. During training, the weights are derived from ground truth reports, while during inference, they are obtained from the reports of similar training images.

In knowledge graph feature extraction, extracting node embeddings with pre-trained transformer models is common \citep{21huang2023kiut,20li2023dynamic,35xu2023vision}; the critical challenge lies in effectively utilizing edge information (i.e., relationships). The pre-defined graph connecting organs and diseases indicates their association but does not specify the relationships. Traditional graph convolutional networks utilize fixed weights, such as the adjacency matrix, to represent the association and update the node features by averaging the features of neighboring nodes. However, this approach fails to capture the importance of different neighbors to the node. To address this, the graph attention mechanism is introduced and computes the attention weights between the node and its neighbors, then performs a weighted aggregation of the neighbors' features to update the node's feature \citep{21huang2023kiut, 35xu2023vision, 20li2023dynamic}. In addition, \citet{20li2023dynamic} added identifiers to indicate node levels (i.e., root, organ, and disease). Combined with the case-specific extraction method (see Section 3.2), this approach yields significant performance improvements. 

\citet{11yang2022knowledge} aims to represent more complex relationships between medical entities, utilizing a general graph embedding model called RotatE \citep{Rotatesun2019rotate}. Relationships are represented as rotations from entity1 to entity2 within a complex vector space. In this way, four common relationships (i.e., symmetry, antisymmetry, inversion, and composition) can be represented. However, these four relationships are not practical for medical data. Relationships can be more specified in the medical field; for instance, RadGraph defines relationships between medical entities as ``suggestive of'', ``located at'', and ``modify''. \citet{11yang2022knowledge} transformed it into sentences (e.g., ``pneumothorax suggestive of bleeding") to extract semantic features using a pre-trained transformer model. 

In conclusion, feature extraction from text data typically relies on transformer models, which lack distinctive designs. For graph data, leveraging relationships is a crucial topic. Exploring ways to incorporate more specific relationships presents a promising research direction.

\subsection{Multi-modal feature fusion and interaction \label{sec:fusion}}

Feature fusion and interaction refer to the integration of multi-modal data from inputs or auxiliary tasks. It is important for report generation using multi-modal inputs. As discussed in Section \ref{sec:Intro}, aligning multimodal data involves four main challenges: distinct feature spaces, identifying complementary and redundant information, various types of noise, and uneven contributions. Fusion and interaction methods can help alleviate the above problems. Additionally, the correspondence between image regions and report sentences can be learned during this process.

The most straightforward approach is feature-level operation including the concatenation, summation, or multiplication of multimodal features (14 works). However, the feature-level operation could be too simple to enable sufficient interaction. Therefore, neural network-based methods are leveraged. \citet{71huang2021deep} used an LSTM network to integrate keywords and image features, starting with image features as input and sequentially introducing keywords at each time step. This method is relatively cumbersome and less effective than attention mechanisms. The multi-head attention mechanism is a widely adopted method for feature fusion (28 works), involving three inputs: query, key, and value. Its core idea is to compute the similarity between the query and key, and use this correlation to weight the value, enabling the identification of alignable features across different modalities while filtering out irrelevant information. The attention mechanism can be applied to concatenated feature data through self-attention, allowing for an arbitrary number of input modalities. For example, \citet{72dalla2022multimodal} concatenated clinical information tokens with visual features and used the combined features as the query, key, and value. \citet{108pellegrini2023rad} merged question features with image features, feeding the concatenated features into the BERT model. However, it does not differentiate between specific modalities. Alternatively, another approach is that one modality (e.g., images) can act as the query, while another modality (e.g., textual keywords) serves as keys and values, allowing the query modality to guide the model in learning to identify relevant information from other modalities. For instance, \citet{11yang2022knowledge} proposed a Knowledge-enhanced multi-head attention mechanism, using visual features as queries and knowledge graph entity embeddings as keys and values for feature fusion, with relation embeddings incorporated. Ablation studies showed that including relation embeddings improves performance. However, there is no detailed experimental analysis of the difference between concatenating features as Q, K, V inputs in attention and using one as Q and the others as K and V.

We would like to highlight the memory metric-based method proposed by \citet{5chen2021cross} to facilitate feature interaction. Theoretically, certain parts of the textual and visual features should convey the same information. However, differences in feature spaces make direct mapping between text and image features challenging, requiring an intermediate medium to facilitate the process. \citet{5chen2021cross} employed a $N\times D$ metric for this purpose, where $N$ refers to the number of memory vectors, and $D$ refers to the dimension of each memory vector. First, the metric was initialized randomly. Image features, text features from generated tokens, and memory metric features then underwent linear transformation to align their feature spaces. Subsequently, distances between the image features and memory metric features, as well as text features and memory metric features, were calculated. The top $K$ metric features with the closest distances to the image or text were selected, respectively. These selected features were then weighted based on these distances and were fed back into an encoder-decoder structure. This metric, serving as a medium, ensured that the selected visual and textual features belong to the same feature space (i.e., memory metric), loosely aligning the visual and textual features and reducing the problems arising from distinct feature spaces. Additionally, the authors tested separate textual and visual memory metrics, which also improved performance. This suggests that regularity patterns may exist within radiology images and reports, and using the memory metric can better capture these patterns to enhance model performance. This point is further supported by \citet{37yang2023radiology}, who introduce a memory metric that dynamically captures knowledge during training. This metric is then used as a knowledge base during inference to address the limitation of manually constructing knowledge graphs, as discussed in Section \ref{par:featurelearning_auxiliary_task}. The memory metric is updated at each step by incorporating textual and visual features, as well as metrics from the previous step, using an attention mechanism. 

Two studies \citep{32qin2022reinforced,50you2022jpg} followed \citet{5chen2021cross}. \citet{33wang2022cross} modified it in two ways: 1) they initialized the memory metric by visual and textual features; 2) the cross-modal interaction occurred only among cases with the same label. These two modifications resulted in a notable improvement. \citet{102li2023unify} contended that the methods mentioned lack explicit constraints for cross-modal alignments. They considered orthonormal bases as the memory metric and input them along with visual or textual features, into multi-head attention modules. Then the outputs of attention modules were processed by a self-defined gate mechanism. A triplet matching loss was utilized to align the processed visual and textual features. This method slightly improved the results. 

Overall, the multimodal fusion and interaction methods reviewed primarily focus on addressing challenges associated with distinct feature spaces. To balance the contributions of different modalities, a common strategy involves automatically learning the weights of contributions during training. However, regarding the other two challenges, many approaches rely heavily on similarity calculations between inputs to identify informative features. This reliance limits their ability to capture complementary information that is critical for maximizing the strengths of each modality. Furthermore, there is often a lack of specialized mechanisms to effectively address various types of noise.

\subsection{Report generation\label{sec:generation}}

The last step is report generation, which utilizes extracted features from earlier steps to produce the final reports. The generation methods mainly include decoder-based techniques (Section \ref{sec:decoder}), retrieval-based techniques (Section \ref{sec: retrieval&temp}), and template-based techniques (Section \ref{sec: retrieval&temp}).

\subsubsection{Decoder-based techniques for report generation\label{sec:decoder}}

A typical method for training decoder-based report generation is autoregressive training. In this process, image (or fused) features and generated tokens (or real text preceding the token awaiting generation) are input into the model to compute the probability distribution for the next token. New tokens are generated by sampling from this distribution, and a loss function is applied between the output distribution and the ground truth. Two main challenges arise in this process: first, how to mitigate information forgetting and extract useful information from long texts to generate new tokens. Second, designing an effective loss function is crucial for guiding text generation and avoiding biases, such as favouring high-frequency words over important medical keywords. The approaches to these two challenges are primarily discussed in Sections \ref{par:ar_de} and \ref{sec: loss}. Additionally, this section presents two other training strategies: reinforcement learning (Section \ref{sec: Rein}) and curriculum learning (Section \ref{sec: TrainingStrategy}).

\paragraph{Architectures for decoder-based techniques \label{par:ar_de}}
The decoder decodes the extracted representation of inputs and generates a descriptive report. The mainstream architectures include LSTM \citep{hochreiter1997long} and Transformer. Compared to LSTM, the Transformer processes the entire sequence simultaneously rather than sequentially. Therefore, the Transformer allows for more efficient parallelization during training and can capture long-range dependencies. In the 100 reviewed papers, the Transformer tends to replace LSTM. Sixty-six works utilized the Transformer as a decoder and only 23 of them utilized LSTMs (12 works) or hierarchical LSTMs (11 works). For the papers published in 2023 and 2024, all encoder-decoder structures used the Transformer as their decoders. There are two ways to modify the decoder and improve model performance by more effectively capturing the information within the text: shortcut connections and memory-driven transformer.

 \textit{Shortcut connections:} Connecting different layers in networks can be considered as a promising way to enhance the flow of information in both forward and backward propagation \citep{mirikharaji2023survey}. The U-connection \citep{21huang2023kiut} and meshed connection \citep{7miura2021improving,56lee2022cross, M2Transcornia2020meshed} are added between encoder and decoder, resulting in a similar performance enhancement \citep{21huang2023kiut}.

\textit{Memory-driven Transformer:} We would like to highlight the Memory-driven Transformer (R2Gen) proposed by \citet{58chen2020generating}. It has been increasingly popular in recent years. The R2Gen introduces a memory module and a memory-driven conditional layer normalization module into the Transformer decoder architecture. The design of the memory module hypothesizes that diverse images exhibit similar patterns in their radiological reports, thereby serving as valuable references for each other. Building a memory matrix can capture this pattern and transfer it during training. Specifically, similar to that in Section \ref{enhance}, a matrix is randomly initialized and is updated using the gate mechanism based on the matrix from the last step and generated reports. The layer normalization is designed to integrate the outputs of the memory module into the decoder.

Eight works directly utilized the R2Gen as their decoder. In addition, the design of the memory module and layer normalization inspired subsequent works \citep{99xue2024generating, 53jia2021radiology, 39zhang2023novel}. It is noted that the novel utilization of the memory module by \citet{39zhang2023novel} integrates ground truth reports into the training process, leading to successful outcomes.

\paragraph{Loss functions for decoder-based techniques\label{sec: loss}}
The mainstream loss function for report generation is the cross-entropy loss based on the generated sentences and the ground-truth sentences. Cross-entropy loss can be re-weighted based on term frequency \citep{80gajbhiye2022translating}, TF-IDF \citep{51wang2022automated} or uncertainty \citep{104wang2024trust} to mitigate model bias or handle challenging cases. In addition, \citet{45pandey2021explains} utilized cycle-consistency loss \citep{zhu2017unpaired} to generate reports. The core idea is that a report and its corresponding image share the same information, hence they can be used to generate each other. 

The application of an auxiliary loss function can provide additional supervision signals, further enhancing model performance. Two works \citet{25wang2021self, 46li2022self} applied an additional constraint between features extracted from the generated and real reports. \citet{15wu2023token} applied a penalty to the predicted probabilities of frequent tokens. \citet{22zhang2023semi} created two different versions of an input image by adding noise and feeding them into two networks. An auxiliary loss function ensures the consistency of the outputs produced by the two generators. \citet{98wang2024camanet} obtained two discriminative regions in an image from the generated words and visual classifier separately, then enforced the consistency between them. 

\paragraph{Reinforcement learning for decoder-based techniques \label{sec: Rein}}
Reinforcement learning offers a method to update the model parameters based on non-differentiable reward functions \citep{messina2022survey}. In report generation, the model is typically viewed as an agent interacting with a generation environment. Input features and previously generated tokens are considered as the states of the environment. The parameters of the model define a policy for generating actions (i.e., next token generation) based on the states, with a reward computed at each step based on the action \citep{32qin2022reinforced}. The model is trained to maximize the reward using various strategies such as the policy gradients-based REINFORCE algorithm \citep{williams1992simple}. This reward-based training is often combined with loss-based approaches, either by alternating between them \citep{96kaur2022cadxreport} or first training with loss, followed by rewards \citep{32qin2022reinforced}.  Improvement in the specified evaluation metric (see Section \ref{sec:evaluation}) can be considered as the basis for designing rewards, such as the Consensus-based Image Description Evaluation (CIDEr) \citep{96kaur2022cadxreport}, the Bilingual Evaluation Understudy (BLEU) \citep{32qin2022reinforced, 105gu2024complex}, the Metric for Evaluation of Translation with Explicit ORdering (METEOR) \citep{32qin2022reinforced}, the Recall-Oriented Understudy for Gisting Evaluation (ROUGE) \citep{32qin2022reinforced}, BERTScore \citep{7miura2021improving}, F1 score \citep{7miura2021improving, 15wu2023token}, and accuracy \citep{47hou2021automatic}.  In addition, \citet{47hou2021automatic} trained a language fluency discriminator using the ground truth and generated reports, and then utilized the discriminator to provide rewards.

\paragraph{Curriculum learning for decoder-based techniques \label{sec: TrainingStrategy}}

Data-level curriculum learning is a strategy that organizes training samples in the order of difficulties, allowing the model to learn in a progressive manner \citep{soviany2022curriculum}. Unlike traditional methods, where training data is shuffled and indiscriminately fed into the model, this approach typically involves: (1) ranking the training data based on a predefined strategy; (2) starting the training process with simpler samples and progressively introducing more complex ones; and (3) training the model on the full dataset to reinforce its learning. The key element of this approach is data ranking (i.e., the first step), which mimics the human learning progression. This prevents the model from being overwhelmed by complex samples early in training, thereby reducing the risk of getting stuck in suboptimal solutions. Moreover, the strategy is applicable to both loss-based and reward-based training methods. \citet{4liu2021competence} introduced curriculum learning to the field of report generation by designing specific metrics to evaluate the difficulty of image-report training data. They proposed four metrics named image heuristics, image confidence, text heuristics, and text confidence. Image heuristic evaluates the similarity between input images and normal images. Image confidence indicates the confidence of a classification model. The report heuristic was related to the number of abnormal sentences, and report confidence is evaluated by the negative log-likelihood loss \citep{CurriculumL-2xu2020dynamic}. Instead of directly summing the four metrics, they adaptively select the most suitable metric using dynamic feedback for each training step, and use the selected metric to choose an appropriate training batch. Experimental results show that incorporating this training strategy into various methods leads to improved outcomes.

In conclusion, the Transformer has replaced LSTM as the dominant architecture. Structural enhancements, such as short connections, improve information flow and alleviate gradient vanishing, though they also increase model complexity. The memory matrix allows cross-referencing between cases, but attention must be paid to memory size, as an excessively large memory can introduce redundancy and degrade performance. Another improvement lies in the loss function. It involves either assigning different weights to words or incorporating auxiliary loss functions; however, these additional supervision signals provide limited information compared to those used in image feature extraction. Reinforcement learning offers a method for optimizing non-differentiable objectives, but focusing on specific rewards (e.g., ROUGE-L) may negatively impact performance on other metrics, making reward function design critical. Additionally, curriculum learning presents a promising approach to organizing training data.

\subsubsection{Retrieval-based and template-based techniques for report generation \label{sec: retrieval&temp}}

Retrieval-based techniques generate reports by selecting existing sentences from a large corpus and the selection is typically based on similarity comparison \citep{27endo2021retrieval,12ramesh2022improving,73jeong2024multimodal}. Initially, text and image encoders are trained using a contrastive method, such as the CLIP \citep{CLIPradford2021learning}. The textual features of sentences in a corpus and the visual features of an input image are extracted by the encoders. The visual features are then compared with all textual features in the corpus. The top $K$ sentences with the maximum similarity score are selected for the predicted report. In addition, \citet{73jeong2024multimodal} added a multimodal encoder after the retrieval process to calculate the image-text matching scores between the input image and the retrieved sentences. A filter was applied based on the score to remove entailed or contradicted sentences. 

Other retrieval-based techniques do not follow the above process. \citet{91kong2022transq} treated the report generation in two steps sentence retrieval and selection. They first retrieved a candidate sentence set from the training datasets with far more sentences than a standard medical report and then selected the sentences by a classifier. \citet{70zhang2022category} proposed a retrieval method based on a hashing technique, which mapped multi-modal data with the same label into a shared space.

Template-based methods typically start with the diagnosis of diseases, and then pre-defined sentences are selected based on the diagnosis results. These selected sentences are concatenated to produce reports \citep{49pino2021clinically}. \citet{83abela2022automated} argued that this method was limited by exact labels, therefore they retrieved template sentences by class probabilities and different thresholds corresponding to different descriptions.

In conclusion, retrieval-based and template-based techniques utilize existing data or predefined templates, theoretically enhancing performance, particularly in linguistic fluency. However, challenges such as inadequate feature representation and immature similarity measurement often lead to inferior outcomes. For instance, while inadequate feature representation may only affect specific generated words in decoder-based models, it can result in entirely irrelevant sentence selections in retrieval-based and template-based approaches. Furthermore, these techniques generally exhibit lower flexibility and limited content generation, negatively impacting human-computer interaction.

\section{Large Models in Report Generation\label{sec:lmm}}

Large models, including language models, vision models and multi-modal models, have revolutionized the field of deep learning. Particularly for large language models, their human-like conversational capabilities and ability to respond to free-text queries without task-specific training have been profoundly impactful. However, deploying general large language models in clinical settings remains unfeasible due to insufficient response accuracy and safety concerns \citep{multimoda2yan2023multimodal, 128saab2024capabilities}, which has led to significant attention on their application to medical tasks such as report generation.

In the initial phases, rather than directly generating reports from imaging data, researchers focus on using general large language models to revise reports produced by traditional methods \citep{60selivanov2023medical,54wang2023chatcad}. Traditional deep learning methods were applied to predict report-related information (e.g., lesion regions) and generate preliminary reports from images. Subsequently, pre-trained models (e.g., ChatGPT \citep{54wang2023chatcad} and GPT-3 \citep{GPT3brown2020language, 60selivanov2023medical,54wang2023chatcad}) were employed to improve these reports based on the predicted information, leading to moderate improvements. 

Training a large model on a large medical dataset is a more straightforward solution to achieve better performance. An overview of these models is presented in Table \ref{tab:SumLMM}. We have provided the GitHub repository link for the open-source model. Some links provide only inference APIs \citep{129alkhaldi2024minigpt,124chen2024chexagent}, while others include training code \citep{126zhang2024generalist,127li2024llava,120wang2023r2gengpt,118pellegrini2023radialog,wu2023generalistfoundationmodelradiology}. This process generally comprises two main phases. The first phase is pre-training, where the model learns general feature representations through unsupervised or self-supervised learning on a large, unlabeled medical dataset. This phase enables the model to acquire broad generalization capabilities and adjust its weights to be more aligned with the target medical tasks. The second phase is fine-tuning, which involves refining the model for a specific task (e.g., medical report generation) using a relatively small, high-quality labeled dataset. The common monitoring tool is wandb\footnote{\url{https://wandb.ai/home}}, tracking metrics like loss or validation set results, similar to traditional training. Understanding this process requires grasping three key components:

\begin{table*}[!t]
\caption{\label{tab:SumLMM} Comparison of vision–language models for medical tasks. ``Params" describes the parameter scale, roughly indicating the size of LLMs. ``Foud" indicates if the model can be used for other medical tasks besides report generation. ``RG" shows if it is trained or tested for report generation. ``OS" refers to its open-source status.}
\centering
\begin{tabular}{|>{\raggedright\arraybackslash}p{2cm}|>{\raggedright\arraybackslash}p{1.7cm}|>{\raggedright\arraybackslash}p{2.6cm}|>{\raggedright\arraybackslash}p{1.7cm}|>{\raggedright\arraybackslash}p{3cm}|>{\raggedright\arraybackslash}p{0.7cm}|>{\raggedright\arraybackslash}p{0.7cm}|>{\raggedright\arraybackslash}p{2.5cm}|}
\hline
Name & Params& Visual Encoder &Adapter& LLM&Foud& RG & OS\\
\hline
RaDialog\newline\citep{118pellegrini2023radialog}&7B/13B/33B&BioViL-T&Q-Former&Vicuna-7B/13B/33B&\ding{55}&\ding{51}&\url{https://github.com/ChantalMP/RaDialog}\\
\hline
R2GenGPT\newline \citep{120wang2023r2gengpt}&7B&Swin Transformer&Linear&LLaMA2-7B&\ding{55}&\ding{51}&\url{https://github.com/wang-zhanyu/R2GenGPT}\\
\hline
LLM-CXR\newline \citep{119lee2023llm}&13B&VQ-GAN&--&LLaMA2-13B&\ding{55}&\ding{51}&\ding{55}\\
\hline
MAIRA-1\newline \citep{122hyland2023maira}&7B&RAD-DINO&MLP-4&Vicuna-7B&\ding{55}&\ding{51}&\ding{55}\\
\hline
MAIRA-2\newline \citep{123bannur2024maira}&7B/13B&RAD-DINO&MLP-4&Vicuna-7B/13B v1.5&\ding{55}&\ding{51}&\ding{55}\\
\hline
CheXagent\newline \citep{124chen2024chexagent}&7B&EVA-CLIP&Q-Former&Mistral-7B-v0.1&\ding{51}&\ding{51}&\url{https://github.com/Stanford-AIMI/CheXagent}\\
\hline
LLava-Med\newline \citep{127li2024llava}& 7B&BioMedCLIP&Linear&LLaMA-7B&\ding{51}&\ding{55}&\url{https://github.com/microsoft/LLaVA-Med}\\
\hline
MiniGPT-Med\newline\citep{129alkhaldi2024minigpt}&7B&EVA-CLIP&Linear&LLaMA2-7B&\ding{51}&\ding{51}&\url{https://github.com/Vision-CAIR/MiniGPT-Med}\\
\hline
LLaVA-Ultra\newline \citep{125guo2024llavaultra}&13B&CLIP-ViT-L/14 and SAM-ViT-L &Linear&LLaMA-13B&\ding{51}&\ding{55}&\ding{55}\\
\hline
BiomedGPT-B\newline \citep{126zhang2024generalist}&182M&ResNet-101&--&BERT-style encoder\newline + GPT-style decoder&\ding{51}&\ding{51}&\url{https://github.com/taokz/BiomedGPT}\\
\hline
RadFM \newline \citep{wu2023generalistfoundationmodelradiology}&14B&3D ViT&Transformer +linear&MedLLaMA-13B&\ding{51}&\ding{51}&\url{https://github.com/chaoyi-wu/RadFM}\\
\hline
MedVersa \newline \citep{zhou2024generalistlearnermultifacetedmedical}&--&Swin Transformer /3D UNet&Linear&Llama-2&\ding{51}&\ding{51}&\ding{55}\\
\hline
\end{tabular}
\end{table*}

\begin{itemize}
    \item Acquisition of medical image-text datasets with prompts designed for training.
    \item Except for BiomedGPT \citep{126zhang2024generalist}, the other reviewed papers are based on large language models. Thus, understanding the transition from large language models to large multimodal models, particularly regarding the integration of image data, is essential.
    \item Training large models demands substantial computational resources. For instance, \citet{touvron2023llama} reported that training the famous large language model named LLaMA with 65B-parameter required 2,048 A100 GPUs with 80GB of RAM over approximately 21 days. Such resource demands are beyond the reach of most researchers. Therefore, it is crucial to explore methods for training large medical models with limited resources.
\end{itemize}

We discussed the aforementioned issues in Sections \ref{sec:llm1}, \ref{sec:llm2}, and \ref{sec:llm3}, respectively. 
It is untenable to fully separate large models for generating medical reports from foundational medical models. Thus, in these sections, foundational model methods are introduced first, followed by their application in report generation.

\subsection{Acquisition of medical image-text datasets with prompts \label{sec:llm1}}

A large dataset is required during the pre-training phase. PMC-15M \citep{zhang2023large} is a dataset comprising 15 million biomedical image-text pairs, each consisting of an image and its corresponding caption from biomedical papers. In addition to medical images, it also includes other types of data, such as general biomedical illustrations and microscopy images, without any prompts provided. Therefore, it is impractical to directly utilize the entire dataset for medical image analysis.  \citet{127li2024llava} derived two sub-datasets from PMC-15M for large model training. The first dataset, consisting of 600K image-text pairs, was selected based on caption length. Corresponding prompts were randomly chosen from a predefined list, such as ``Describe the image concisely''. The second dataset consists of multi-round questions and answers for 60K filtered images from the PMC-15M dataset. The image types include X-rays, CT scans, MRIs, histopathology slides, and gross pathology images. In addition to the original text associated with each image, sentences referring to these images were extracted from the original papers to provide extra context. Image, original text, and additional descriptions were then provided to GPT-4, which was tasked with generating multi-round questions and answers as if it could see the images. This novel dataset design enables the Large Language and Vision Assistant for BioMedicine (LLaVa-Med) \citep{127li2024llava} to perform well through pre-training on it and also inspired subsequent works \citep{125guo2024llavaultra}. 

A 60k medical image dataset is relatively small for large model training, and the image-text pair format limits its utility for diverse medical tasks such as disease diagnosis. Consequently, efforts have been made to compile a comprehensive radiology-focused dataset, with the most straightforward approach being the integration of multiple public datasets \citep{129alkhaldi2024minigpt, 126zhang2024generalist, 124chen2024chexagent, wu2023generalistfoundationmodelradiology}. For example, \citet{126zhang2024generalist} aggregated 14 public datasets and \citet{124chen2024chexagent} derived a large-scale instruction-tuning dataset about Chest X-rays from 28 publicly available datasets. Distinct prompt templates were designed for each type of task. For report generation, most studies incorporate MIMIC-CXR into the training set \citep{129alkhaldi2024minigpt}, with a few also utilizing the Indiana University Chest X-Ray Collection (IU-Xray) \citep{126zhang2024generalist}. In addition, most datasets contained only 2D images. \citet{wu2023generalistfoundationmodelradiology} integrated 18 public datasets to construct a radiology image dataset called the Medical Multi-modal Dataset (MedMD), which includes 15.5M 2D scans and 500k 3D scans. They also filtered MedMD, retaining only radiology images. However, no studies have yet conducted ablation experiments to assess how pre-training on different datasets affects model performance. For example, whether using the MIMIC-CXR dataset in both the pre-training and fine-tuning phases improves report generation compared to using it only in fine-tuning.

For the fine-tuning phase, prompts need to be designed based on the existing report generation dataset (e.g., MIMIC-CXR). Most reviewed papers rely on fixed prompts, such as “Generate a comprehensive and detailed diagnosis report for this chest X-ray image” \citep{120wang2023r2gengpt}, or use prompts based on the other sections in the report (i.e., indication, technique, and comparison) to provide the model with additional patient-related information \citep{122hyland2023maira,123bannur2024maira}. \citet{118pellegrini2023radialog} integrated disease classification results into the prompt. However, the classifier's low accuracy yields minimal improvement. Given that prompt design is critical for optimizing model performance \citep{liu2023pre}, further research into developing effective prompts for report-generation tasks is necessary.

\subsection{Evolution from large language models to large multimodal models\label{sec:llm2}}

\begin{figure}[!t]
\centering
\includegraphics[width=0.3\columnwidth]{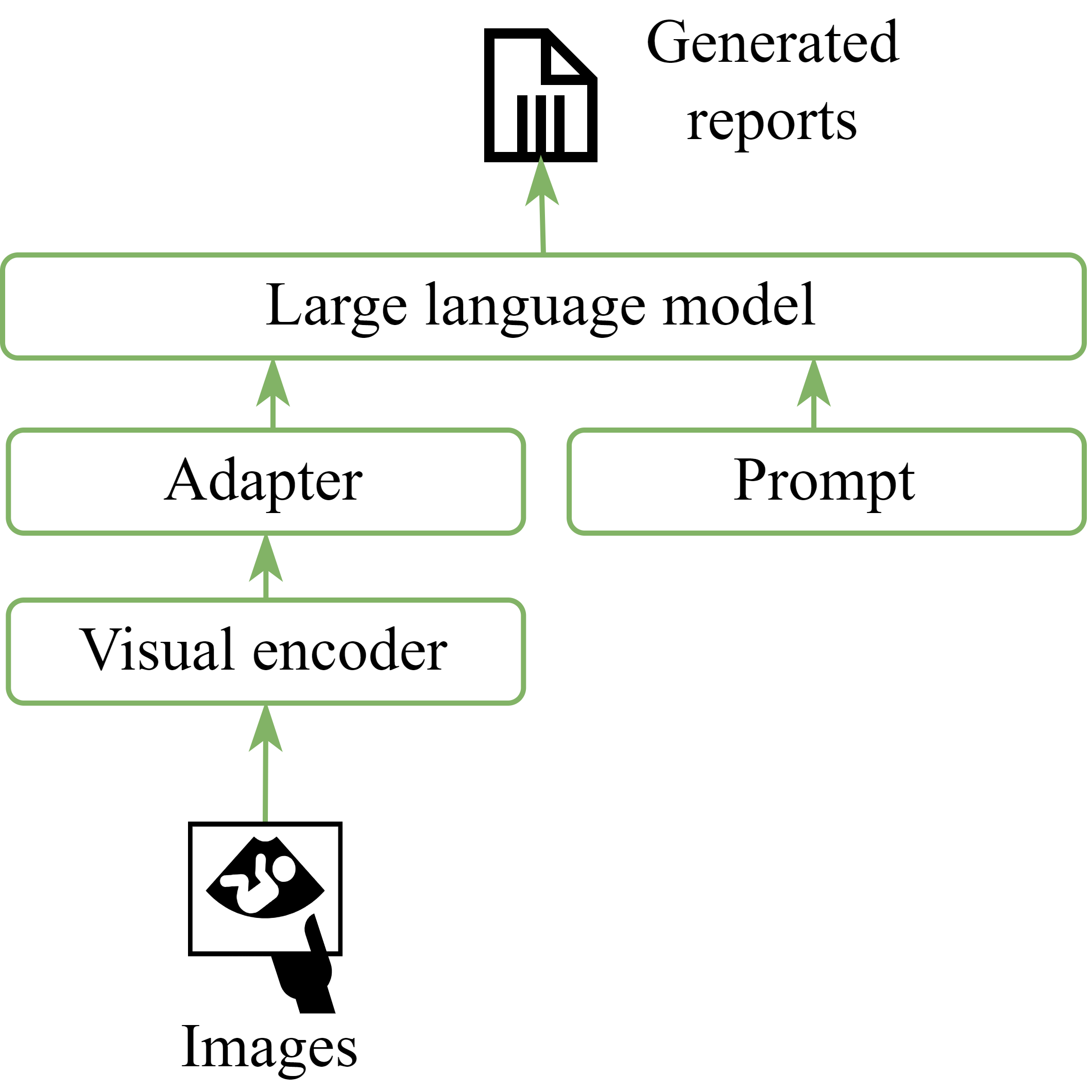}
\caption{\label{fig:LMM}A typical process for transitioning from language models to multimodal models.}
\end{figure}

A common approach for evolving from a large language model to a large multimodal model is to add an adapter between the visual encoder and the language model to align image features with the language model's input, as shown in Figure \ref{fig:LMM}. Table \ref{tab:SumLMM} lists the adapter designs, which are usually linear projections or multilayer perceptrons (MLP). \citet{122hyland2023maira} found that a 4-layer MLP outperforms a 2-layer MLP on report generation tasks, due to the increased non-linearity enhancing the model's expressive power. Additionally, \citet{118pellegrini2023radialog} introduced a more sophisticated adapter (i.e., Q-Former) from Bilp-2 \citep{li2023blip}, but it yielded no significant improvement. 

When training large foundational medical models, task-specific prompts are designed, such as ``Does the patient have {disease} X?" for disease diagnosis, ``Is this image from {modality} X?" for modality recognition, and ``Generate a detailed diagnosis report for this chest x-ray image." for report generation. Typically, cross-entropy loss is applied between the predicted and ground-truth text sequences, helping to standardize the training process across different tasks.

Based on the architecture, the evolution can be achieved in one step by training the visual encoder, adapter, or large language model directly with the target dataset \citep{120wang2023r2gengpt,122hyland2023maira}. The model's initial weights come from general large language models (e.g., Vicuna \citep{chiang2023vicuna}, LLaMA \citep{touvron2023llama}, Mistral \citep{jiang2023mistral}), general visual encoder (e.g., EVA-CLIP \citep{sun2023eva}), and domain-specific visual encoders (e.g., BioMedCLIP \citep{zhang2023large}, BioViL\citep{bannur2023learning}, RAD-DINO \citep{perez2024rad}). \citet{120wang2023r2gengpt} found that for report generation, training only the adapter was ineffective compared to training both the adapter and the visual encoder. \citet{122hyland2023maira} observed that when the visual encoder is frozen and only the adapter and LLM are trained, the choice of a radiology-specific visual encoder can significantly improve model performance. \citet{125guo2024llavaultra} argued that the size imbalance between the visual encoder and the language model causes a bottleneck in feature extraction. To address this, they use two large visual encoders, CLIP and the Segment Anything Model (SAM) \citep{kirillov2023segment}, and combine their features through weighted addition, which outperforms the use of a single encoder.

The one-step approach is highly convenient; however, training the entire visual encoder or LLM on large datasets is costly. To address this, two-step approaches have been proposed for scenarios that require training on larger datasets to achieve better generalization. Inspired by LLaVA \citep{liu2024visual}, LLaVa-Med \citep{127li2024llava} presents a feasible solution. In the initial phase, the 600K dataset from PMC-15M is used to train the adapter for image captioning, with the visual encoder and language model parameters kept frozen. In the subsequent phase, the model is trained on the 60K instruction set from PMC-15M. The visual encoder weights remain fixed, and the projection layer and language model parameters are updated. 

\citet{119lee2023llm} argued that using adapters hinders the interaction between visual and semantic information. To address this, they proposed using frozen Vector Quantised Variational Autoencoder (VQ-GAN) \citep{esser2021taming} to represent image information in a discrete form, incorporating these image tokens as special tokens into the vocabulary of large language models. While experimental results have yet to show improvements in report generation tasks using this method compared to adapters, the VQ-GAN approach remains notable. It introduces a learnable codebook that maps encoded image features to the nearest discrete variable, enabling discretization. This technique is well-established in natural image generation for capturing high-quality discrete image representations but remains underexplored in report generation.

Most medical foundation models are trained directly on multi-task datasets, enabling them to perform a variety of tasks. In contrast, \citet{zhou2024generalistlearnermultifacetedmedical} treat the large model as a planner, predicting specific tokens that activate the corresponding vision modules (e.g., detection and segmentation), which allows it to handle an even wider array of tasks. For instance, when the \texttt{<DET>} token is predicted, the detection module is triggered to generate bounding box information. However, this approach increases the number of models involved, making the pipeline non-end-to-end.

Integrating image features into large language models while preserving their conversational abilities is crucial. However, the current report-generation large models do not achieve this. General medical models are only capable of handling simple conversations. Nonetheless, as shown in Section \ref{sec:comparison}, with accessing to sufficient data, large models demonstrate advantages on natural language-based evaluation metric, highlighting their potential.

\subsection{Training large medical models with limited resources \label{sec:llm3}}

Directly applying a general-purpose LLM into a highly specialized task (e.g. radiology report generation) can not achieve an optimal performance, therefore normally requires LLM fine-tuning using a task-specific dataset. However, fine-tuning the entire LLM is extremely resources demanding. Employing adapters, frozen layers, and training only a subset of the LLM parameters can conserve training resources; however, these techniques often involve a trade-off between efficiency and model performance. Additionally, incorporating extra adapters increases the overall model parameters, further burdening the computational resources required during the inference stage and prolonging the inference time. This section presents two more direct approaches for conserving training resources: Low-Rank Adaptation (LoRA) \citep{hu2021lora}, a classic large model fine-tuning technique widely utilized in report generation task \citep{120wang2023r2gengpt,129alkhaldi2024minigpt,118pellegrini2023radialog,125guo2024llavaultra}, and a lightweight medical foundation model for medical tasks, including report generation.

When fine-tuning large models, the weights $W$ update can be expressed as $W + \Delta W$, where $\Delta W$ denotes the change in $W$ during training. The core idea of LoRA \footnote{\url{https://github.com/microsoft/LoRA}} is that not all parameters need to be changed for a specific target task (e.g., report generation) during fine-tuning. The update weights $\Delta W$ can be efficiently represented using a low-rank matrix and computed via low-rank decomposition. Specifically, given model weights $W \in \mathbb{R}^{d \times k}$, the weight update is assumed to be of low rank $r$, i.e., $W + \Delta W = W + BA$, where $B \in \mathbb{R}^{d \times r}$ and $A \in \mathbb{R}^{r \times k}$, with $r \ll \min(d, k)$. Thus, only the low-rank matrices $A$ and $B$ are trained, and their product $BA$ is added to the original weights. This approach substantially reduces the computational resources required, as only the delta weights $\Delta W$ need to be stored for each new model, thereby conserving disk space. Moreover, experiments conducted on large models (such as GPT-3 175B) have demonstrated that models fine-tuned with LoRA can achieve performance comparable to, or even surpassing, that of full model fine-tuning. However, \citet{120wang2023r2gengpt} observed that when applying this approach to fine-tune the adapter and visual encoder (with 90.9M parameters) for report generation, the performance was inferior to that of full model fine-tuning. This discrepancy arises because, as the model size decreases, the proportion of effective parameters increases, thereby diminishing the advantage of using low-rank matrices to approximate $\Delta W$. The LoRA has been integrated into the State-of-the-art Parameter-Efficient Fine-Tuning (PEFT) library \footnote{\url{https://github.com/huggingface/peft}} and can be directly accessed via its interface. PEFT is designed to efficiently fine-tune large pre-trained models.

A straightforward strategy to conserve training resources is to reduce the model's trainable parameters. However, due to the performance improvements \citep{123bannur2024maira} and enhanced generalization ability with larger models, most reviewed studies still employ models exceeding 7B parameters (see Table \ref{tab:SumLMM}). Recently, \citet{126zhang2024generalist} developed lightweight medical foundation models of 33M/93M/182M parameters, with the 182M BiomedGPT-B achieving state-of-the-art results in 16 out of 25 medical tasks compared to 7B models. Although it has not been compared with other models on the standard report generation benchmark, it offers a valuable insight that relatively lightweight large models can be trained to accomplish report generation. BiomedGPT mainly differs from traditional models by curating a large-scale pretraining corpus and specifically developing instruction-tuning data, emphasizing the importance of a high-quality labeled large training dataset. Although for the report generation task, BiomedGPT-B was tested on only 30 image-report pairs and evaluated manually rather than using natural language or medical correctness metrics, it is still regarded as a significant advancement for lightweight foundation models in medical report generation. 

It is noteworthy that there remains no definitive consensus on the optimal model size and data quantity requirements for training large models for report generation tasks. \citet{118pellegrini2023radialog} fine-tuned models with 7B, 13B, and 33B parameters on the MIMIC dataset, finding no significant performance differences. In contrast, \citet{123bannur2024maira} merged four datasets as training datasets, introducing more diverse inputs and outputs rich in information (as described in Section \ref{sec:exp_auxiliary}). They trained both 7B and 13B models, observing performance improvements as model size increased (see Table \ref{tab:Compare_MIMIC}). Similarly, \citet{126zhang2024generalist} noted consistent performance enhancements with models of 33M, 92M, and 182M parameters. Despite differences in architecture, these findings suggest a bottleneck effect of model size and data usage on model performance, underscoring the need for systematic investigation.

\section{Explainability\label{sec:inter}}
Explainability of a deep learning model is defined as the ability to justify an outcome in understandable terms for a human \citep{doshi2017towards, messina2022survey}. This is especially critical in medical contexts, where model outputs can directly or indirectly influence clinical decision-making. However, most current research focuses on improving report generation model performance, with relatively few studies considering explainability. This section summarizes recent advancements in this area, including visualization, auxiliary tasks, and uncertainty quantification.
\subsection{Visualization for explainability}
Aligned with previous surveys \citep{messina2022survey}, visualization remains the dominant method for interpreting model outputs. It highlights the most crucial part of the inputs relevant to report generation, such as specific areas in images, keywords in text, and certain nodes in knowledge graphs. Visualization allows radiologists to assess whether the model accurately targets the relevant information, thereby fostering greater confidence in the system. The reviewed report generation methods mainly employ two approaches to performing visualization. 

The first approach is a gradient-based method named Gradient-weighted Class Activation Mapping (Grad-CAM) \citep{selvaraju2017grad}. It calculates the importance of each image feature map for a particular prediction by weighting them with the gradients and then performing a weighted sum. Report generation methods often use Grad-CAM in auxiliary classification tasks to identify image regions that influence model decisions \citep{8alfarghaly2021automated,18huang2021deepopht,49pino2021clinically, 96kaur2022cadxreport}, with some methods applying it directly to report generation tasks to determine the image regions relevant to report generation \citep{29najdenkoska2021variational, 29_2najdenkoska2022uncertainty}. 

The second approach utilizes the attention weights. Compared to Grad-CAM, which is limited to images, attention-based visualization is more applicable across various input data modalities, including images \citep{1liu2021exploring, 5chen2021cross, 10wang2023metransformer,22zhang2023semi, 46li2022self,32qin2022reinforced,39zhang2023novel,61moon2022multi,63hou2021ratchet,67cao2023cmt,55lin2023contrastive,50you2022jpg,102li2023unify,91kong2022transq,99xue2024generating}, medical terminology \citep{1liu2021exploring}, retrieved real reports \citep{1liu2021exploring}, and medical knowledge base \citep{1liu2021exploring}. Typically, attention weights are derived from the Scaled Dot-Product Attention mechanism \citep{vaswani2017attention}, which accepts query (Q), key (K), and value (V) vectors as inputs. The formulation is expressed as  $Attention(Q, K, V) = softmax(\frac{QK^T}{\sqrt{d_k}})V$, where $\sqrt{d_k}$ is a scaling factor related to the input dimension. The expression $softmax(\frac{QK^T}{\sqrt{d_k}})$ is interpreted as the attention weights.

Regardless of visualization methods, studies typically provide limited image examples to illustrate that their models can accurately identify relevant regions; however, they are deficient in the systematic evaluation of large-scale datasets. Moreover, there is a lack of quantitative metrics to assess visualization results, possibly due to the absence of ground truth. Non-image-based ground truth can be obtained through text comparisons, while image-based ground truth can be established by collecting physicians' eye-tracking data or using bounding boxes for lesions or anatomical structures.

\subsection{Auxiliary tasks for explainability \label{sec:exp_auxiliary}}
Auxiliary tasks can provide valuable information for image classification or detection (see Section \ref{par:featurelearning_auxiliary_task}), thus enhancing system interpretability by offering radiologists additional insights. However, the potential of auxiliary tasks in improving interpretability has been largely underutilized in previous works \citep{messina2022survey}. \citet{13tanida2023interactive} addressed this gap by linking output results to specific image regions via object detection, enabling users to select areas of interest and receive corresponding explanations. \citet{113dalla2023finding} introduced a multi-label classification head that classifies findings for each detected anatomical region. This model outputs both the bounding box locations and findings as intermediate results, which increases model transparency. \citet{123bannur2024maira} outputs the corresponding bounding box while generating each lesion-containing description. 

\subsection{Uncertainty quantification for explainability}

The uncertainty discussed in this section quantifies the model's confidence in the quality of the generated outputs. \citet{104wang2024trust} measured the uncertainty based on Monte Carlo dropout variational inference \citep{gal2016dropout}. When dropout is applied in the network, performing $T$ stochastic forward passes for a given input allows the variance of the outputs to reflect the model's uncertainty. Greater variance implies greater uncertainty. They defined report-level textual uncertainty and sentence-level textual uncertainty. Report-level uncertainty is measured by the variance of similarities between the $T$ generated reports. For sentence-level uncertainty, they first identify the report with the minimum similarity distance to other reports from the $T$ outputs, termed the average reference report. The uncertainty of each sentence is then represented by the variance of the similarities between the sentences in the average reference report and their corresponding sentences in the $T$ generated reports. Compared to report-level uncertainty, sentence-level uncertainty enables physicians and patients to assign varying trust levels to individual sentences rather than fully affirming or rejecting the entire report, warranting further study.

\section{Datasets\label{sec:datasets}}

Datasets play a crucial role in the development of report-generation models. Abundant and diverse training data can improve the model's accuracy and generalizability. Moreover, a suitable test dataset makes it realistic to test the model's performance in a practical scenario. In this section, we selected 11 public medical image-report datasets utilized in the reviewed articles to provide a comprehensive introduction to popular and newly collected datasets, see Table \ref{tab:SumDataset}. The usage of datasets in each article is shown in Table \ref{tab:overall} in Appendix A. The datasets primarily focus on the lung, including X-ray \citep{IU-XRAYdemner2016preparing, MIMICjohnson2019mimic, MIMIC-CXR-JPGjohnson2019mimic-2} and CT \citep{COV-CTRli2020auxiliary, 24liu2021medical}. In addition, there are also publicly available datasets on eye scans \citep{18huang2021deepopht,55lin2023contrastive, FFA-IRli2021ffa} and breast scans \citep{43yang2021automatic}. The concentration of medical report generation efforts on chest X-rays can be attributed, in part, to the accessibility of large-scale publicly available datasets.

\begin{figure*}[!htbp]
\centering
\includegraphics[width=\columnwidth]{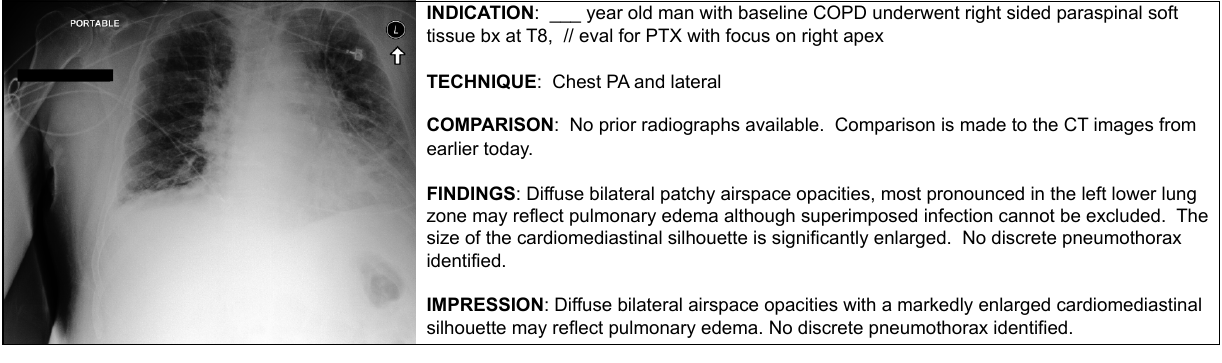}
\caption{\label{fig:DataSample.} A sample from the MIMIC-CXR dataset. The left is the Chest X-ray image and the right is its corresponding radiology report.}
\end{figure*}

The most popular datasets are the IU-Xray \citep{IU-XRAYdemner2016preparing} and the MIMIC-CXR \citep{MIMICjohnson2019mimic, MIMIC-CXR-JPGjohnson2019mimic-2}. The IU-Xray contains 7470 images of frontal and lateral X-rays and 3955 reports with manual annotation based on the MeSH codes\footnote{https://www.nlm.nih.gov/mesh/meshhome.html} and the RadLex codes. These codes encompass standard medical terminology, such as anatomical structures, diseases, pathological signs, foreign objects, and attributes, as defined by authoritative institutions. Manual keyword annotation is invaluable because it helps standardize synonym expressions and clarify linguistic ambiguities, which facilitates the structuring of information in free-text reports \citep{108pellegrini2023rad}. However, IU-Xray annotations are designed to improve retrieval efficiency rather than to structure information, which may result in the overlooking of speculative details when converting free-text reports into a structured format. For example, the annotation for ``Focal airspace disease in the right middle lobe. This is most concerning for pneumonia" is annotated as ``Airspace Disease/lung/middle lobe/right/focal", which omits the concern for pneumonia.  Additionally, the IU-Xray dataset does not provide an official split strategy. Most reviewed studies use a 7:1:2 ratio for training, validation, and test sets. For deep learning methods, the size of the IU-Xray is insufficient, while the collection of MIMIC-CXR alleviates this problem \citep{messina2022survey}. \citet{58chen2020generating} also noted that compared to IU-Xray, MIMIC-CXR may exhibit a greater diversity in report patterns. This is more conducive to training a model with strong generalization performance. MIMIC-CXR consists of 377,110 images and 227,827 reports and has been multi-labeled by automatic tools into 14 categories, including lung-related diseases (e.g., pneumonia), heart-related diseases (e.g., cardiomegaly), pleural-related diseases (e.g., pleural effusion), bone-related conditions (e.g., fracture), support devices, and cases where no abnormal finding is detected. The F1 scores for consolidation, pneumonia, atelectasis, pneumothorax, pleural effusion, and fracture annotation all exceed 0.9, indicating the reliability of these automatically annotated labels. An image-report sample from the MIMIC-CXR dataset is shown in Figure \ref{fig:DataSample.}. The indication, technique, and comparison provide fundamental information for a Chest X-ray test. The automation of radiology report generation targets the findings and impressions sections. The findings section provides a detailed description of the entire image, while the impressions section summarizes these observations. The MIMIC-CXR dataset provides an official data split strategy, consisting of 368,960 training images with 222,758 reports, 2,991 validation images with 1,808 reports, and 5,159 test images with 3,269 reports. To facilitate different methods, many studies have preprocessed this dataset. The most well-known version, introduced by R2Gen \citep{58chen2020generating}, consists of 270,790 training images, 2,130 validation images, and 3,858 test images, and has been widely used in subsequent approaches. However, \citet{58chen2020generating} paired each report with only one image, treating multiple images for the same report (e.g., frontal and lateral views) as distinct cases. This can introduce errors, particularly in contrastive learning, as different views might be seen as negative pairs. Moreover, if a report references findings only visible in the lateral view but is matched with a frontal image, this misalignment could negatively impact the model training. Thus, while this data version is popular, careful selection is necessary when using it. Research on the role of frontal and lateral views in automated report generation has been a focus of previous studies. For instance, \citet{bertrand2019lateral} found that the lateral view alone is less effective than the frontal view, which provides more comprehensive information. However, combining both views can enhance the diagnosis of certain diseases, as they offer complementary data \citep{laserson2018textray}. As discussed in Section 3.3.1.1, \citet{101wang2023self} demonstrated how this combination can be achieved using transformer attention. Nonetheless, utilizing both views increases computational costs and demands more processing power from devices.

The follow-up work provides MIMIC-CXR with richer label information. The Chest ImaGenome dataset \citep{ImaGenomewu2021chest} is based on the anteroposterior and posteroanterior view Chest X-ray images in the MIMIC-CXR dataset. It provides an anatomy-centered scene graph for each image based on sentence-level annotation. The graph includes objects (e.g., lung), attributes (e.g., opacity), and their relationships. Bounding box locations are specified for key anatomical structures. The annotations are generated through two automated pipelines, and a separate dataset of 500 manually annotated cases is also provided for testing. Additionally, using timestamps from MIMIC-CXR, anatomical annotations are compared across sequential exams based on descriptive changes (e.g., decrease). \citet{109dalla2023controllable} incorporated the anatomy information from prior reports into model training, improving most metrics. It should be noted that the ImaGenome dataset uses only frontal images, likely because lateral images have less influence on automatic report generation. Frontal images are generally sufficient for annotating most anatomical structures, and even in the presence of partial obstructions (e.g., up to 15\% of the lung obscured by the heart or diaphragm \citep{raoof2012interpretation}), the annotation boxes remain largely unaffected.  A new dataset, CheXpert Plus, comparable to MIMIC-CXR, was recently released by Stanford University \citep{chambon2024chexpertplusaugmentinglarge}, containing 223,462 images and 187,711 reports. The studies for each patient are also arranged in chronological order. As an extension of the widely used chest X-ray classification dataset named CheXpert  \citep{CheXpertirvin2019chexpert}, it adds corresponding reports, demographic information, automated labels for 14 disease categories, automated graph-based annotations, and a collection of pre-trained models built on this dataset. Although CheXpert Plus has fewer reports than MIMIC, the reports are longer, particularly in the impression section. CheXpert Plus is the largest available dataset for tasks related to the impression section. Additionally, it has been de-identified without altering the structure of the reports, making them more representative of real-world radiology text.

The above datasets contain X-ray images. Besides X-rays, CT is another important imaging modality for the chest. Publicly available datasets for CT report generation were primarily collected for COVID-19 research. The COV-CTR dataset \citep{COV-CTRli2020auxiliary} includes 349 COVID and 379 Non-COVID images, sourced from published papers. One limitation of this dataset is its relatively low image quality. In contrast, the COVID-19 CT dataset \citep{24liu2021medical} contains higher-quality images, which were directly collected from hospitals, offering clearer diagnostic details.

Increasing attention has been paid to the collection of ophthalmic image-report pairs \citep{18huang2021deepopht,FFA-IRli2021ffa,55lin2023contrastive}. Ophthalmic image has lots of modalities, such as Fluorescein Angiography (FA), Fundus Fluorescein Angiography (FFA), Color Fundus photography (CFP), fundus photograph (FP), optical coherence tomography (OCT), fundus autofluorescence (FAF), Indocyanine Green Chorioangiography (ICG), and red-free filtered fundus images. Most retinal datasets are based on one or two modalities and the last released Retina ImBank dataset is the first multi-modality retinal image-text dataset \citep{55lin2023contrastive}. Some datasets provide additional labels for each image, such as lesion boundary \citep{FFA-IRli2021ffa}, lesion category \citep{18huang2021deepopht,FFA-IRli2021ffa}, and keywords \citep{18huang2021deepopht}.

The findings and impressions section of the reports in the aforementioned datasets is in free-text format, which may exhibit expression ambiguity and complicate the assessment of clinical accuracy. Structured reporting presents a potential solution by employing standardized formats for documenting radiological findings and interpretations through predefined templates or checklists, such as ``heart size: normal" \citep{nobel2022structured}. In clinical practice, generating structured reports typically requires doctors to answer a sequence of questions \citep{108pellegrini2023rad}.   \citet{108pellegrini2023rad} released a structured report dataset named Rad-ReStruct based on the IU-Xray. They designed a structured report template with a series of single- or multi-choice questions based on topic existence (e.g., Are there any diseases in the lung?), element existence (e.g., Is there an opacity in the lung?), and attributes (e.g., What is the degree?). They then integrated the IU-Xray report data into the template using its annotated MeSH and RadLex codes. 

The training data exhibits a long-tail distribution. For example, in the IU-Xray dataset, ``normal'' appears 2152 times, while ``degenerated'' appears only once \citep{9yang2021joint}. In MIMIC-CXR and CheXpert Plus, seven out of 14 diseases have positive occurrence rates below 10\%, with one as low as 0.9\% (pleural other in MIMIC-CXR). As a result, the model tends to favor more common words or options from the training set. Overcoming this data imbalance issue is essential.

The training and testing data for most reviewed traditional deep learning models come from the same dataset, which overlooks the importance of testing model generalization --- a critical factor for clinical applications. Given the limited size of the IU-Xray dataset, several studies trained their models using other datasets and then used the entire IU-Xray dataset as an external validation set to assess the model's generalization performance \citep{7miura2021improving,123bannur2024maira}. Moreover, clinical data exists in dynamic environments where patient populations, clinical practices, and operational processes evolve over time \citep{kelly2019key}. Therefore, for a model to be truly applicable in clinical settings, dynamic test datasets and comprehensive calibration processes are necessary to ensure the model can continue to meet requirements as data evolves.

\begin{table*}[!t]
\caption{\label{tab:SumDataset}Public datasets for medical report generation. The report format described (either free-text or structured) refers to the description of the medical condition.}
\centering
\begin{tabular}{|>{\raggedright\arraybackslash}p{2.5cm}|>{\raggedright\arraybackslash}p{1.8cm}|>{\raggedright\arraybackslash}p{3cm}|>{\raggedright\arraybackslash}p{1.3cm}|>{\raggedright\arraybackslash}p{1.1cm}|>{\raggedright\arraybackslash}p{1cm}|>{\raggedright\arraybackslash}p{1.3cm}|>{\raggedright\arraybackslash}p{2.8cm}|}
\hline
Name & Image type & Label type& Images & Reports&Patients& Used by&Featured use cases\\
\hline
IU-Xray \citep{IU-XRAYdemner2016preparing} &Frontal and
lateral chest X-rays  & Free text report, manual/automatic keywords annotation& 7,470 & 3,955&3,955& 71 works&Structured report construction\\
\hline
MIMIC-CXR \citep{MIMICjohnson2019mimic, MIMIC-CXR-JPGjohnson2019mimic-2} & Frontal and lateral chest X-rays & Free text report, automatic 14 diseases annotation& 377,110 & 227,827&65,379& 68 works&Large-scale dataset, diagnosis, knowledge base\\
\hline
Chest ImaGenome \citep{ImaGenomewu2021chest}&Frontal chest X-ray & Free text report, automatic objects (with locations), attributes, and relations annotation&242,072&217,013&--&4 works&Detection, anatomy-based sentence generation, explainability, temporal change assessment\\
\hline
COV-CTR \citep{COV-CTRli2020auxiliary}& Lung CT& Free text report&728&728&--&4 works&COVID-related scenarios\\
\hline
DeepEyeNet \citep{18huang2021deepopht}&FA, CFP&Free text report, manual 265 diseases, and keywords annotation&15,709&15,709&--&3 works&Diagnosis\\
\hline
FFA-IR \citep{FFA-IRli2021ffa}&FFA& Free text report, manual retinal lesion boundary and category annotation&1,048,584&10,790&--&2 works&Detection, explainability\\
\hline
COVID-19 CT \citep{24liu2021medical}&Chest CT& Free text report&1,104&368&96&1 work&COVID-related scenarios\\
BCD2018 \citep{43yang2021automatic}&Breast Ultrasound & Free text report&5,349&5,349&--&1 work&--\\
\hline
Retina ImBank \citep{55lin2023contrastive}&FP, OCT, FFA, FAF,  ICG, and red-free filtered fundus images& Free text report&18,788&18,788&--&1 work&Diverse image modalities\\
\hline
Retina Chinese \citep{55lin2023contrastive}&FP, FFA, and ICG & Free text report&57,498&57,498&--&1 work&--\\
\hline
Rad-ReStruct \citep{108pellegrini2023rad}&Frontal and
lateral chest X-rays & Structured reports&3,720&3,597&3,597&1 work&Structural report generation\\
\hline
CheXpert Plus \citep{chambon2024chexpertplusaugmentinglarge}&Frontal and lateral chest X-rays&Free text report, demographics, automatic 14 diseases annotation, RadGraph annotations&223,462&187,711&64,725&0 work&Large-scale dataset, diagnosis, knowledge base\\
\hline
\end{tabular}
\end{table*}

\section{Evaluation\label{sec:evaluation}}
Accurate assessment of the quality of generated reports is crucial for measuring model performance. The quality of generated reports can be evaluated both quantitatively and qualitatively. Quantitative methods check the text quality and medical correctness of the generated report by natural language evaluation metrics (Section \ref{sec:NLE}) and medical correctness metrics (Section \ref{sec:MCM}), respectively. Qualitative evaluation is normally performed by human experts and they provide overall evaluation for the generated reports (Section \ref{sec:HumanE}). Table \ref{tab:SumMetric} summarizes the description of evaluation methods introduced in this section. The usage of evaluation methods in each article is shown in Table \ref{tab:overall} in Appendix A.

\begin{table*}[!t]
\caption{\label{tab:SumMetric}The quantitative and qualitative evaluation methods for medical report generation. }
\centering
\begin{tabular}{|p{4cm}|p{11.3cm}|p{1.3cm}|}
\hline
Metrics & Description & Used by\\
\hline
\multicolumn{3}{|c|}{Natural language-based metrics}\\
\hline
BLEU \newline\citep{BLEUpapineni2002bleu} & A precision-based metric that measures the n-gram overlapping of the generated text and ground truth text. & 90 works\\
ROUGE-L \newline\citep{Rougelin2004rouge}&A F1-like metric that computes a weighted harmonic mean of precision and recall based on the longest common subsequence.&81 works\\
METEOR \newline\citep{Methorbanerjee2005meteor}&A F1-like metric that computes a weighted harmonic mean of unigram precision and recall. It is an extension of BLEU-1.&63 works\\
CIDEr\newline\citep{Cidervedantam2015cider}&The cosine similarity between generated text and ground truth text based on the TF-IDF.&41 works\\
$S_{emb}$\newline\citep{27endo2021retrieval}&Sending ground truth report and generated report into a textual feature extractor and calculating the cosine similarity between their embeddings from the last layer.&2 works\\
\%Novel\newline\citep{novelvan2018measuring}&The percentage of generated descriptions that are not present in the training data.&1 work\\
\hline
\multicolumn{3}{|c|}{Medical correctness-based metrics}\\
\hline
Clinical Efficacy \newline\citep{58chen2020generating,CEliu2019clinically}&Calculate accuracy, precision, recall, and F1 score based on observations extracted from reference reports and generated reports by automated system.&40 works\\
RadGraph-based metrics\newline\citep{Radgraphjain2021radgraph, delbrouck2022improving}&Calculate precision, recall, and F1 score based on knowledge graphs extracted from reference reports and generated reports by RadGraph series models.&5 works\\
MIRQI\newline\citep{3zhang2020radiology}&Calculate precision, recall, and F1 score based on graph comparison. The ground truth and generated reports are automatically analysed to construct a sub-graph from a defined abnormality graph.&2 works\\
nKTD\newline\citep{17zhou2021visual}&Calculate the Hamming distance based on observations extracted from reference reports and generated reports by the CheXpert Labeller.&1 work\\
\hline
\multicolumn{3}{|c|}{Human-based evaluation}\\
\hline
Comparison&Generate reports by two different models and allow senior radiologists to find which report is better.&12 works\\
Classification&Radiologists categorize the produced reports as accurate, missing details, and false reports.&1 work\\
Error scoring&Radiologists assess the error severity of baseline, model generated, and reference reports. &1 work\\
Grading&
Radiologists need to assign a 5-point scale grade to two types of generated reports, in accordance with clinical standards. &2 works\\
\hline
\multicolumn{3}{|c|}{Large language model-based  evaluation}\\
\hline
GREEN \citep{ostmeier2024green}&A large model takes a reference report and a generated report as input, identifies six types of clinically significant errors, and calculates an overall score.&0 work\\
\hline
\end{tabular}
\end{table*}

\subsection{Natural language-based evaluation metrics\label{sec:NLE}}
Natural language evaluation metrics are from natural language processing tasks and measure the general text quality of generated reports. In the reviewed papers, the most popular metrics are BLEU \citep{BLEUpapineni2002bleu}, ROUGE-L \citep{Rougelin2004rouge}, METEOR \citep{Methorbanerjee2005meteor}, and CIDEr \citep{Cidervedantam2015cider}, which are based on n-gram matching between reference reports and generated reports. The model is deemed superior with an increased number of matches. Among them, the BLEU is the earliest and proposed a modified precision method. When evaluating the quality of radiology report generation, we typically opt for BLEU-1, BLEU-2, BLEU-3, and BLEU-4 metrics. The n in BLEU-n means the calculation is based on n-gram. The METEOR is an extension of BLEU-1 and introduces recall into evaluation. The ROUGE-L also considers precision and recall based on the longest common subsequence between reference and generated text. The CIDEr adopts the TF-IDF. The TF-IDF vectors weigh each n-gram in a sentence, and then the cosine similarity is calculated between the TF-IDF vectors of reference and generated text. When the model consistently produces the most common sentences, it can achieve notable BLEU scores. However, CIDEr can evaluate generated outputs by encouraging the appearance of important terms and punishing high-frequency vocabulary \citep{20li2023dynamic}. CIDEr has a popular variant CIDEr-D, which introduces penalties to generate desired sentence length and remove stemming to ensure the proper usage of word forms.

Other than n-gram matching, \citet{27endo2021retrieval} proposed a new metric $S_{emb}$. A pre-trained feature extractor is applied on both the ground truth and the generated reports, and the cosine similarity between extracted embeddings is calculated. This approach is used to assess whether the semantic information contained in two sentences is consistent. Another natural language metric \%Novel \citep{novelvan2018measuring} is introduced to evaluate the diversity in image captioning.

\subsection{Medical correctness metrics\label{sec:MCM}}
Natural language evaluation metrics evaluate the similarity between produced reports and the ground truth, but cannot accurately measure whether the generated reports contain the required medical facts \citep{referencebabar2021evaluating,messina2022survey}. Hence, medical correctness metrics are proposed to pay attention to the prediction of important medical facts. Generally, an automatic labeler is applied to extract medical facts from generated and reference reports. Then different metrics are applied to these. The mainstream metric is clinical efficacy \citep{CEliu2019clinically,58chen2020generating}, which initially calculates precision, recall, and F1 score, and subsequently extends to accuracy \citep{76babar2021encoder,7miura2021improving,61moon2022multi,42yan2022memory,11yang2022knowledge,60selivanov2023medical,37yang2023radiology} and AUC \citep{20li2023dynamic}. Most of the works utilized CheXpert \citep{CheXpertirvin2019chexpert} as a labeler to extract chest diseases information from ground truth reports, while seven of them \citep{7miura2021improving,42yan2022memory,54wang2023chatcad,98wang2024camanet,113dalla2023finding,109dalla2023controllable,112jin2024promptmrg,118pellegrini2023radialog,122hyland2023maira} utilized a newer labeler CheXbert \citep{CheXbertsmit2020combining}, which has a higher performance. Another popular annotation tool named RadGraph extracts knowledge graphs (i.e., entities and relations) from the reports. It has an upgraded version called RadGraph-XL as mentioned in Section \ref{sec:Inputs}. Most works employed RadGraphF1 and calculated the overlap of entities and relations independently before averaging the results \citep{12ramesh2022improving,73jeong2024multimodal,122hyland2023maira,123bannur2024maira}. In addition, \citet{108pellegrini2023rad} generated structured reports by predicting a series of questions. They utilized macro precision, recall, and F1 score to assess all questions, along with evaluating report-level accuracy. Other metrics calculate the Hamming distance \citep{17zhou2021visual} or perform graph comparison \citep{3zhang2020radiology} based on the extracted results.

\subsection{Human-based evaluation\label{sec:HumanE}}
For qualitative assessment, the most common human-based evaluation method is comparison. In general, a set number of samples (i.e., 100/200/300) are selected from the test dataset and subsequently processed by different generated models. More than one professional clinician is responsible to compare and sort the generated reports. Five works utilized ground truth reports in this process. Two works \citep{7miura2021improving,34cao2022kdtnet} considered the ground truth reports as a reference. Reports were generated by different generators and the radiologists need to select which report is more similar to the reference.  \citet{72dalla2022multimodal} allowed experts to find 5 types of errors (i.e., hallucination, omission, attribute error, impression error, and grammatical error) in different generated reports according to the reference reports.  \citet{35xu2023vision} asked radiologists to rank the ground truth reports and the generated reports. \citet{32qin2022reinforced} invited experts to select the most suitable report from the generated and the ground truth reports according to correctness, language fluency, and content coverage. Other expert evaluation methods are classification \citep{8alfarghaly2021automated}, grading \citep{104wang2024trust}, and error scoring \citep{73jeong2024multimodal}. 

\subsection{Large language model-based evaluation}
Recently, the human-like language comprehension abilities of large language models have achieved impressive performance, leading to their application in evaluation. For instance, LLaVA-Med utilizes GPT-4 to assess the helpfulness, relevance, accuracy, and level of detail of generated responses while assigning an overall score. Similarly, CheXagent provides GPT-4 with a reference report and two generated reports, enabling it to identify the superior report. However, \citet{ostmeier2024green} found that GPT-4 exhibited a low correlation with expert preferences. To address this issue, they developed a specialized model, Generative Radiology Report Evaluation and Error Notation (GREEN), to evaluate report generation. This model takes a reference report and a generated report as input, allowing it to identify and categorize errors into six categories while calculating a score. The strong correlation between this score and expert evaluations underscores its effectiveness.

\section{Benchmark Comparison\label{sec:comparison}}
For model comparison, it is essential to select a benchmark for an impartial evaluation. We choose the MIMIC-CXR \citep{MIMIC-CXR-JPGjohnson2019mimic-2,MIMICjohnson2019mimic}(see Table \ref{tab:Compare_MIMIC}) dataset as a benchmark to compare the model performance for two reasons. First, according to Table \ref{tab:SumDataset}, IU-Xray and MIMIC-CXR are the most popular datasets, but IU-Xray lacks standard training-validation-test splits, leading to less comparable results \citep{messina2022survey}, while MIMIC-CXR has official training-validation-test splits. Second, MIMIC-CXR is the largest image-report dataset, which provides a broader distribution of data, facilitating testing across diverse scenarios, and reducing biases commonly encountered in smaller datasets. This section first introduces the process of paper selection and discusses issues related to benchmark usage. We then review the techniques employed by the top two methods at each step. Finally, we summarize the performance of large model-based approaches and compare them with traditional deep learning methods.

We endeavour to ensure equitable comparisons, but it's important to consider the following three problems when analysing these results. First, we select the methods leveraged the original official splits of the MIMIC-CXR dataset \citep{MIMIC-CXR-JPGjohnson2019mimic-2}, as described in Section \ref{sec:datasets}. Several methodologies highlighted in Section \ref{sec:methods} are excluded due to their differing data utilization strategies. Given the importance of large model methods, we include their results in the comparison despite minor differences from the official test set \citep{120wang2023r2gengpt}. Second, we choose BLEU, ROUGE, METEOR, CIDEr-D, precision, recall, and F1 score metrics as the evaluation metrics. Similar to the previous survey \citep{messina2022survey}, we found that although some metrics have variants, many papers do not specify the particular version used. In that case, we assume they are consistent. Third, the generated report sections vary among different methodologies. Most articles do not explicitly specify the generated report sections, making it challenging to conduct comparative analysis. 

We separate the comparison results based on the generated report section. The completed comparison tables are provided in Table  \ref{tab:ACompare_MIMIC} in Appendix A. Table \ref{tab:Compare_MIMIC} presents the top two performances for each evaluation metric. For input acquisition, three of them \citep{66song2022cross,106liu2023observation,112jin2024promptmrg} utilized multi-modality inputs. While the ablation studies in these papers demonstrate that leveraging multi-modal data yields significant performance enhancements, particularly when real reports are retrieved and used as input (as detailed in Section \ref{sec:data_prepare}) \citep{66song2022cross,112jin2024promptmrg}, the improvements are less pronounced compared to the methods that do not employ multi-modal data. Only \citep{106liu2023observation} achieved an increase of 0.003 on METEOR, and \citep{112jin2024promptmrg} improved precision by 0.02. Thus, the utilization of multi-modal data warrants further enhancement.

\begin{table*}[!t]
\caption{\label{tab:Compare_MIMIC}Comparisons of the model performance on the MIMIC-CXR Dataset. B1, B2, B3, B4, R-L, C-D, P, R, and F represent BLEU-1, BLEU-2, BLEU-3, BLEU-4, ROUGE-L, CIDEr-D, precision, recall, and F1 score, respectively. \textcolor{red}{\textbf{The best}} and \textcolor{blue}{\textbf{second best}} results are highlighted. All values were extracted from their papers.  $\uparrow$ indicates that a higher value for this metric is better. * means large model-based methods.}
\centering
\begin{tabular}{|p{3.8cm}|p{0.9cm}p{0.9cm}p{0.9cm}p{0.9cm}|p{0.9cm}|p{1.5cm}|p{0.9cm}|p{0.9cm}|p{0.9cm}|p{0.9cm}|}
\hline
Paper & B1$\uparrow$ & B2$\uparrow$&B3$\uparrow$&B4$\uparrow$&R-L$\uparrow$&METEOR$\uparrow$&C-D$\uparrow$&P$\uparrow$&R$\uparrow$&F$\uparrow$\\
\hline
\multicolumn{11}{|c|}{Findings Section}\\
\hline
\citet{5chen2021cross}&0.353&\textcolor{blue}{\textbf{0.218}}&\textcolor{blue}{\textbf{0.148}}&0.106&0.278&0.142&--&0.334&0.275&0.278\\
\citet{49pino2021clinically}&-&-&-&-&0.185&--&\textcolor{red}{\textbf{0.238}}&\textcolor{blue}{\textbf{0.381}}&\textcolor{red}{\textbf{0.531}}&\textcolor{blue}{\textbf{0.428}}\\
\citet{66song2022cross}&0.360&\textcolor{red}{\textbf{0.227}}&\textcolor{red}{\textbf{0.156}}&0.117&0.287&0.148&-&\textcolor{red}{\textbf{0.444}}&\textcolor{blue}{\textbf{0.297}}&0.356\\
\citet{123bannur2024maira} 7B*&\textcolor{blue}{\textbf{0.465}}&--&--&\textcolor{blue}{\textbf{0.234}}&\textcolor{blue}{\textbf{0.384}}&\textcolor{blue}{\textbf{0.419}}&--&--&--&0.427\\
\citet{123bannur2024maira} 13B*&\textcolor{red}{\textbf{0.479}}&--&--&\textcolor{red}{\textbf{0.243}}&\textcolor{red}{\textbf{0.391}}&\textcolor{red}{\textbf{0.430}}&--&--&--&\textcolor{red}{\textbf{0.439}}\\
\hline
\multicolumn{11}{|c|}{Impression + Findings Section}\\
\hline
\citet{84wu2022multimodal}&0.340&0.212&0.145&0.103&0.270&\textcolor{blue}{\textbf{0.139}}&0.109&-&-&-\\
\citet{51wang2022automated}&0.351&0.223&\textcolor{blue}{\textbf{0.157}}&0.118&0.287&-&\textcolor{red}{\textbf{0.281}}&-&-&-\\
\citet{69jia2022few}&\textcolor{blue}{\textbf{0.363}}&\textcolor{blue}{\textbf{0.228}}&0.156&0.130&\textcolor{red}{\textbf{0.300}}&-&-&-&-&-\\
\citet{120wang2023r2gengpt}*&\textcolor{red}{\textbf{0.411}}&\textcolor{red}{\textbf{0.267}}&\textcolor{red}{\textbf{0.186}}&\textcolor{blue}{\textbf{0.134}}&\textcolor{blue}{\textbf{0.297}}&\textcolor{red}{\textbf{0.160}}&\textcolor{blue}{\textbf{0.269}}&\textcolor{red}{\textbf{0.392}}&\textcolor{red}{\textbf{0.387}}&\textcolor{red}{\textbf{0.389}}\\
\citet{zhou2024generalistlearnermultifacetedmedical}*&--&--&--&\textcolor{red}{\textbf{0.160}}&--&--&--&--&--&--\\
\hline
\multicolumn{11}{|c|}{Unspecified generated sections}\\
\hline
\citet{119lee2023llm}*&0.092&0.046&0.026&0.015&0.162&0.069&\textcolor{red}{\textbf{0.525}}&-&-&0.211\\
\citet{101wang2023self}&0.363&0.235&0.164&0.118&\textcolor{blue}{\textbf{0.301}}&0.136&-&-&-&-\\
\citet{98wang2024camanet}&0.374&0.230&0.155&0.112&0.279&0.145&0.161&\textcolor{blue}{\textbf{0.483}}&0.323&0.387\\
\citet{15wu2023token}&0.383&0.224&0.146&0.104&0.280&0.147&-&-&-&\textcolor{red}{\textbf{0.758}}\\
\citet{106liu2023observation}&0.391&0.249&\textcolor{blue}{\textbf{0.172}}&\textcolor{blue}{\textbf{0.125}}&\textcolor{red}{\textbf{0.304}}&0.160&-&-&-&-\\
\citet{112jin2024promptmrg}&0.398&-&-&0.112&0.268&0.157&-&\textcolor{red}{\textbf{0.501}}&\textcolor{blue}{\textbf{0.509}}&0.476\\
\citet{92wang2022medical}&\textcolor{blue}{\textbf{0.413}}&\textcolor{red}{\textbf{0.266}}&\textcolor{red}{\textbf{0.186}}&\textcolor{red}{\textbf{0.136}}&0.298&\textcolor{red}{\textbf{0.170}}&\textcolor{blue}{\textbf{0.429}}&-&-&-\\
\citet{91kong2022transq}&\textcolor{red}{\textbf{0.423}}&\textcolor{blue}{\textbf{0.261}}&0.171&0.116&0.286&\textcolor{blue}{\textbf{0.168}}&-&0.482&\textcolor{red}{\textbf{0.563}}&\textcolor{blue}{\textbf{0.519}}\\
\hline
\end{tabular}
\end{table*}

At the feature learning stage, for the architecture of image feature learning, four works utilized a pure Transformer model \citep{92wang2022medical,91kong2022transq,106liu2023observation,51wang2022automated}, another five works employed the CNN-based model \citep{49pino2021clinically,66song2022cross,84wu2022multimodal,69jia2022few,112jin2024promptmrg}, and four works combined CNN and Transformer \citep{5chen2021cross,101wang2023self,98wang2024camanet,15wu2023token}. None of the reviewed studies compared the impact of Transformer and CNN architectures on report generation tasks. In terms of training strategy and enhancement module, eight works \citep{92wang2022medical,112jin2024promptmrg,15wu2023token,98wang2024camanet,51wang2022automated, 84wu2022multimodal,106liu2023observation} leveraged auxiliary tasks and \citet{66song2022cross} designed contrastive attention. The ablation experiments demonstrated their effectiveness. It is apparent that using auxiliary loss for additional supervision is a widely adopted approach for enhancing report generation efficacy during the visual encoder phase. For non-image feature learning, there is a consensus on using transformer-based architectures, and for feature fusion, feature-level operations or attention are employed before inputting them into subsequent networks, lacking specific design considerations. 

At the generation stage, the majority of studies employ a decoder, with only \citet{49pino2021clinically} utilizing template-based approaches and \citet{91kong2022transq} adopting retrieval-based techniques. The method proposed by \citet{91kong2022transq} demonstrates superior performance, especially in lesion detection. It initially retrieves a substantial set of sentence candidates for subsequent selection. However, the lack of ablation studies prevents us from assessing the effectiveness of this design. Among the decoder-based methods, only two employed LSTM architecture \citep{84wu2022multimodal,69jia2022few}, while the remaining methods all utilize Transformer architecture. In comparison to transformer-based approaches, LSTM-based methods exhibited inferior performance, selected primarily due to the limited number of models explicitly designed for generating the ``impression'' and ``finding'' sections. Two works utilized the modified Transformer R2Gen \citep{101wang2023self,98wang2024camanet}, and there is no significant advantage. Three studies enhanced cross-entropy loss by re-weighting \citep{51wang2022automated} or adding auxiliary loss \citep{98wang2024camanet, 15wu2023token}, but did not yield significantly improved results. Notably, the outstanding performance of \citet{15wu2023token}'s work on the F1 metric stems from utilizing F1-based rewards in reinforcement learning, but their results on other metrics were subpar. 

For large model-based methods, R2GenGPT \citep{120wang2023r2gengpt} achieved competitive results by fine-tuning only the visual encoder and adapter on the MIMIC-CXR dataset, underscoring the potential of large models in this domain. At the same time, the usage of a rich training dataset and carefully designed input-output settings enabled MAIRA-2 13B \citep{123bannur2024maira} to achieve the best results in finding section generation.  In addition, LLM-CXR \citep{119lee2023llm} presents contrasting results on natural language metrics, with a low BLEU score (BLEU-1: 0.092) and a high CIDEr score (0.525). Disregarding differences in metric versions (e.g., CIDEr and CIDEr-D), this may suggest the model excels in keyword generation, making CIDEr a more reliable metric than BLEU due to its emphasis on keyword importance. Overall, compared with traditional deep learning-based methods, large model-based methods have significantly improved report generation performance on natural language-based metrics, achieving an improvement of more than 10\% in nearly all metrics. This highlights the superior language generation capabilities of large models. However, claiming a clear advantage of large model-based methods over traditional models is difficult. First, large model-based methods typically require extensive training data. For instance, MAIRA-2 7B/13B was trained on a combination of four image-report datasets, resulting in unfair comparisons. It is essential to train traditional models on large-scale data to assess performance improvements from large model architecture or data volume. Second, advancements in medical correctness metrics are less significant, likely due to the current lack of specialized models for lesion detection in large model-based methodologies. Finally, compared to traditional models, a key advantage of large models lies in their ability to engage in interactive dialogues, which is crucial for applications such as telehealth, virtual medical consultations, and medical education \citep{johri2025evaluation}. However, current methods fail to incorporate interactive data in large model training and lack clear evaluation protocols to assess this capability, preventing the advantage of interactive dialogue from being fully realized.

The ability of some general large models to generate reports has not been evaluated on standard benchmarks, and thus these models are not included in the table. For instance, BiomedGPT evaluated its report generation capability by randomly selecting 30 samples from the MIMIC-CXR test set using human-based evaluation. Overall, the evaluation process for large models in report generation remains inconsistent. Variations in evaluation metrics prevent a clear performance hierarchy between general and domain-specific large models. For example, miniGPT-Med outperforms the domain-specific CheXagent on BERT-Sim metric \citep{zhang2019bertscore} and CheXbert-Sim metric \citep{smit2020chexbert}, while the domain-specific MAIRA-2 surpasses the general Med-PaLM M on most metrics (e.g., ROUGE-L). However, it is difficult to compare the performance of miniGPT-Med and MAIRA-2. Recently, \citet{zhang2024rexrank} proposed ReXrank\footnote{\url{https://rexrank.ai/}}, a public leaderboard designed to evaluate AI-based report generation for chest X-ray images. This platform is essential for standardizing chest X-ray report generation benchmarks and serves as a framework for constructing benchmarks for other types of radiological images. Furthermore, an interface has been made available for users to upload their models.

In conclusion, due to the inconsistent usage of benchmarks, this section can only compare a subset of the reviewed papers. In the five stages, the integration of multimodal inputs has not demonstrated clear advantages and lacks unique designs. Current effective designs focus mainly on image feature extraction, especially through auxiliary loss. During the report generation phase, reinforcement learning effectively enhances specific metrics. However, there are currently no effective designs for loss function. Additionally, large models have shown considerable language-related capabilities, but improvements are needed to generate results with higher medical correctness. Further experiments are required to validate their superiority over traditional methods in report generation when using comparable training datasets.

\section{Challenges and Future Works\label{sec:challenges&futurework}}
Although automated systems offer promising efficiency for clinical workflows, current methods have not produced very high-quality reports. This section evaluates the current progress in automated report generation development and identifies potential areas for improvement.

\subsection{Constructing and utilizing multi-modal data} Considering report generation as a multi-modal problem is more aligned with clinical practice \citep{multimoda1tu2024towards,multimoda2yan2023multimodal}. \citet{76babar2021encoder} have proven the ineffectiveness of simple encoder-decoder report generation models and mentioned that adding prior knowledge can be a promising method. However, the current utilization of multi-modal data remains under-explored. Firstly, the methods for non-image feature extraction and the fusion of multi-modality data are often limited and simplistic, such as using graph encoding for the knowledge base and attention mechanisms for feature fusion. Additionally, there is a lack of comparative analysis on the effectiveness of these methods. For instance, the reviewed papers do not compare two attention fusion approaches: one where concatenated features from different modalities are used as the input for Q, K, and V, and another where one modality is used for Q, and the other modality for K and V. Secondly, there is inadequate utilization of complementary information across input modalities and a lack of targeted design for filtering different types of noise. Thirdly, the construction of the knowledge base is imperfect. The pre-defined graph \citep{3zhang2020radiology} is overly simplistic. Despite Radgraph being a vast knowledge base, it is solely derived from reports, lacking the relationship between images and reports, such as organ recognition or understanding of typical radiological scenarios, which radiologists possess. The Chest ImaGenome dataset \citep{ImaGenomewu2021chest} provides organ recognition annotations, alleviating the problem. Additionally, these publicly available knowledge bases only concentrate on chest X-rays. RadGraph-XL addresses this by expanding knowledge graph construction to other medical imaging modalities and anatomies. This expansion facilitates the development of general medical knowledge databases.

\subsection{Evaluation of medical correctness} 
Evaluating the medical correctness of generated reports is crucial for clinical applications. Compared to previous works \citep{messina2022survey, liao2023deep}, recent works have paid more attention to it, but still have two shortcomings. First, in the reviewed articles, medical correctness evaluation has only been applied to chest X-ray reports. Second, the evaluation is based on the automatic labeler of radiology reports, resulting in a limited scope and reliability of the labels. For instance, the labeler in the medical efficacy metric only targeted 14 types of diseases and the average F1 score is around 0.798 \citep{CheXbertsmit2020combining}. Improving the accuracy and scale of automatic labeling tools can help optimize the evaluation process.

\subsection{Large public datasets and unified benchmark comparison }
As shown in Table \ref{tab:SumDataset}, most public datasets are limited in size. Deep learning-based techniques require a large amount of data. The contemporary prevalence of large models underscores this need for large data volumes. Among the datasets, the MIMIC dataset and recently released CheXpert Plus are relatively large, but only include chest X-ray data. Large datasets targeting other image modalities and diseases need to be constructed. In addition, while MIMIC-CXR is a well-established benchmark compared to IU-Xray, dataset utilization lacks standardization, complicating comparisons. We urge papers using the MIMIC dataset to define their training, validation, and testing partitions, with explicit disclosure of the filtration method, particularly for the testing dataset. Open-sourcing the code can also facilitate the comparison of different methods on a unified test set. The recently introduced ReXrank leaderboard significantly addresses benchmark inconsistency, marking a milestone contribution despite focusing only on chest X-rays.

\subsection{Human-AI interaction and explainability \label{sec:interaction}} 
Effective human-computer interaction can improve the explainability of models. Recent studies have increasingly prioritized human-computer interaction. For traditional deep learning-based methods, \citet{86tanwani2022repsnet} constructed a visual question-answering system for the generation of medical reports, and \citet{13tanida2023interactive} linked the output results to image regions using object detection, allowing users to select areas of interest and receive the corresponding language explanations. However, these systems lack comprehensive language understanding, limiting meaningful interaction with users. The advent of large models with advanced language comprehension presents new opportunities. General dialogue systems, such as GPT-4, have demonstrated the ability to articulate their thought processes and provide rigorous logical answers. In contrast, specialized large models for report generation have yet to achieve comparable advancements. Eight recent studies have explored report-generation tasks using large models. One of these studies produces additional bounding boxes to indicate the areas targeted by the generated descriptions \citep{123bannur2024maira}. However, the design of prompts is still relatively uniform, and none have tested whether these models can articulate their thought processes by responding to queries such as ``How did you determine the location of the lesion?'' Additionally, while some research has incorporated patient clinical histories at the input stage, there has been no exploration of scenarios where users might wish to provide medical history in queries, for example ``I previously had tuberculosis. What impact would that have?"

Providing quantitative results can also improve the explainability of the system and increase user trust. Although measuring uncertainty is prevalent in medical image analysis \citep{lambert2024trustworthy}, only one reviewed paper \citep{104wang2024trust} offered quantitative assessments of system uncertainty, suggesting a promising avenue for future research.

\subsection{Standardized report generation} 
Most existing methods primarily address unstructured report generation. In contrast, structured reporting offers several advantages, including time savings \citep{hong2013content, nobel2022structured}, error reduction, lower communication costs associated with ambiguous language, and improved data consistency and comparability for the development of data records and research. \citet{108pellegrini2023rad} developed a structured template and released a related dataset. This promising start could pave the way for further exploration in this direction. It is noteworthy that effective template design is essential for structural reporting; poorly designed templates can fragment related processes into subcategories, disrupt cohesive synthesis, and hinder comprehensive understanding, especially in multi-compartmental pathological conditions. Currently, research on designing effective structured report templates is limited, making it a valuable area for investigation.

\section{Conclusions\label{sec:conclusions}}
In this study, we have conducted a detailed technical review of 100 papers on automatic medical report generation published in the years 2021, 2022, 2023, and 2024 to showcase both mainstream and novel techniques. Our particular focus lies on the utilization and fusion of multi-modality data. The analysis of methods is structured based on the components of the report generation pipeline, presenting the key techniques for each component. Moreover, we examine recent advancements in large model applications and model explainability and summarize current publicly available datasets and evaluation methods, including both quantitative and qualitative assessments. Subsequently, we compare the results of a subset of the reviewed articles under the same experimental setting. Finally, we outline the current challenges and propose future directions for the generation of medical reports. In general, sustained progress is needed to produce high-quality reports. This survey aims to offer a comprehensive overview of report generation techniques, emphasize critical issues, and assist researchers in quickly grasping recent advances in the field to build stronger systems for clinical practice.

\bibliographystyle{plainnat}
\bibliography{references.bib}

\begin{thebibliography}{188}
\providecommand{\natexlab}[1]{#1}
\providecommand{\url}[1]{\texttt{#1}}
\expandafter\ifx\csname urlstyle\endcsname\relax
  \providecommand{\doi}[1]{doi: #1}\else
  \providecommand{\doi}{doi: \begingroup \urlstyle{rm}\Url}\fi

\bibitem[Abela et~al.(2022)Abela, Abu-Khalaf, Yang, Masek, and Gupta]{83abela2022automated}
Brandon Abela, Jumana Abu-Khalaf, Chi-Wei~Robin Yang, Martin Masek, and Ashu Gupta.
\newblock Automated radiology report generation using a transformer-template system: Improved clinical accuracy and an assessment of clinical safety.
\newblock In \emph{AI 2022: Advances in Artificial Intelligence: 35th Australasian Joint Conference, AI 2022, Perth, WA, Australia, December 5--8, 2022, Proceedings}, pages 530--543. Springer, 2022.

\bibitem[Alfarghaly et~al.(2021)Alfarghaly, Khaled, Elkorany, Helal, and Fahmy]{8alfarghaly2021automated}
Omar Alfarghaly, Rana Khaled, Abeer Elkorany, Maha Helal, and Aly Fahmy.
\newblock Automated radiology report generation using conditioned transformers.
\newblock \emph{Informatics in Medicine Unlocked}, 24:\penalty0 100557, 2021.

\bibitem[Alkhaldi et~al.(2024)Alkhaldi, Alnajim, Alabdullatef, Alyahya, Chen, Zhu, Alsinan, and Elhoseiny]{129alkhaldi2024minigpt}
Asma Alkhaldi, Raneem Alnajim, Layan Alabdullatef, Rawan Alyahya, Jun Chen, Deyao Zhu, Ahmed Alsinan, and Mohamed Elhoseiny.
\newblock Minigpt-med: Large language model as a general interface for radiology diagnosis.
\newblock \emph{arXiv preprint arXiv:2407.04106}, 2024.

\bibitem[Alsentzer et~al.(2019)Alsentzer, Murphy, Boag, Weng, Jindi, Naumann, and McDermott]{alsentzer2019publicly}
Emily Alsentzer, John Murphy, William Boag, Wei-Hung Weng, Di~Jindi, Tristan Naumann, and Matthew McDermott.
\newblock Publicly available clinical bert embeddings.
\newblock In \emph{Proceedings of the 2nd Clinical Natural Language Processing Workshop}, pages 72--78, 2019.

\bibitem[Babar et~al.(2021{\natexlab{a}})Babar, van Laarhoven, and Marchiori]{76babar2021encoder}
Zaheer Babar, Twan van Laarhoven, and Elena Marchiori.
\newblock Encoder-decoder models for chest x-ray report generation perform no better than unconditioned baselines.
\newblock \emph{Plos one}, 16\penalty0 (11):\penalty0 e0259639, 2021{\natexlab{a}}.

\bibitem[Babar et~al.(2021{\natexlab{b}})Babar, van Laarhoven, Zanzotto, and Marchiori]{referencebabar2021evaluating}
Zaheer Babar, Twan van Laarhoven, Fabio~Massimo Zanzotto, and Elena Marchiori.
\newblock Evaluating diagnostic content of ai-generated radiology reports of chest x-rays.
\newblock \emph{Artificial Intelligence in Medicine}, 116:\penalty0 102075, 2021{\natexlab{b}}.

\bibitem[Banerjee and Lavie(2005)]{Methorbanerjee2005meteor}
Satanjeev Banerjee and Alon Lavie.
\newblock Meteor: An automatic metric for mt evaluation with improved correlation with human judgments.
\newblock In \emph{Proceedings of the acl workshop on intrinsic and extrinsic evaluation measures for machine translation and/or summarization}, pages 65--72, 2005.

\bibitem[Bannur et~al.(2023)Bannur, Hyland, Liu, Perez-Garcia, Ilse, Castro, Boecking, Sharma, Bouzid, Thieme, et~al.]{bannur2023learning}
Shruthi Bannur, Stephanie Hyland, Qianchu Liu, Fernando Perez-Garcia, Maximilian Ilse, Daniel~C Castro, Benedikt Boecking, Harshita Sharma, Kenza Bouzid, Anja Thieme, et~al.
\newblock Learning to exploit temporal structure for biomedical vision-language processing.
\newblock In \emph{Proceedings of the IEEE/CVF Conference on Computer Vision and Pattern Recognition}, pages 15016--15027, 2023.

\bibitem[Bannur et~al.(2024)Bannur, Bouzid, Castro, Schwaighofer, Bond-Taylor, Ilse, P{\'e}rez-Garc{\'\i}a, Salvatelli, Sharma, Meissen, et~al.]{123bannur2024maira}
Shruthi Bannur, Kenza Bouzid, Daniel~C Castro, Anton Schwaighofer, Sam Bond-Taylor, Maximilian Ilse, Fernando P{\'e}rez-Garc{\'\i}a, Valentina Salvatelli, Harshita Sharma, Felix Meissen, et~al.
\newblock Maira-2: Grounded radiology report generation.
\newblock \emph{arXiv preprint arXiv:2406.04449}, 2024.

\bibitem[Beddiar et~al.(2023)Beddiar, Oussalah, and Sepp{\"a}nen]{beddiar2023automatic}
Djamila-Romaissa Beddiar, Mourad Oussalah, and Tapio Sepp{\"a}nen.
\newblock Automatic captioning for medical imaging (mic): a rapid review of literature.
\newblock \emph{Artificial Intelligence Review}, 56\penalty0 (5):\penalty0 4019--4076, 2023.

\bibitem[Bertrand et~al.(2019)Bertrand, Hashir, and Cohen]{bertrand2019lateral}
Hadrien Bertrand, Mohammad Hashir, and Joseph~Paul Cohen.
\newblock Do lateral views help automated chest x-ray predictions?
\newblock \emph{arXiv preprint arXiv:1904.08534}, 2019.

\bibitem[Brown et~al.(2020)Brown, Mann, Ryder, Subbiah, Kaplan, Dhariwal, Neelakantan, Shyam, Sastry, Askell, et~al.]{GPT3brown2020language}
Tom Brown, Benjamin Mann, Nick Ryder, Melanie Subbiah, Jared~D Kaplan, Prafulla Dhariwal, Arvind Neelakantan, Pranav Shyam, Girish Sastry, Amanda Askell, et~al.
\newblock Language models are few-shot learners.
\newblock \emph{Advances in neural information processing systems}, 33:\penalty0 1877--1901, 2020.

\bibitem[Cao et~al.(2022)Cao, Cui, Yu, Zhang, Li, Liu, and Xu]{34cao2022kdtnet}
Yiming Cao, Lizhen Cui, Fuqiang Yu, Lei Zhang, Zhen Li, Ning Liu, and Yonghui Xu.
\newblock Kdtnet: medical image report generation via knowledge-driven transformer.
\newblock In \emph{Database Systems for Advanced Applications: 27th International Conference, DASFAA 2022, Virtual Event, April 11--14, 2022, Proceedings, Part III}, pages 117--132. Springer, 2022.

\bibitem[Cao et~al.(2023)Cao, Cui, Zhang, Yu, Cheng, Li, Xu, and Miao]{67cao2023cmt}
Yiming Cao, Lizhen Cui, Lei Zhang, Fuqiang Yu, Ziheng Cheng, Zhen Li, Yonghui Xu, and Chunyan Miao.
\newblock Cmt: Cross-modal memory transformer for medical image report generation.
\newblock In \emph{International Conference on Database Systems for Advanced Applications}, pages 415--424. Springer, 2023.

\bibitem[Chambon et~al.(2024)Chambon, Delbrouck, Sounack, Huang, Chen, Varma, Truong, Chuong, and Langlotz]{chambon2024chexpertplusaugmentinglarge}
Pierre Chambon, Jean-Benoit Delbrouck, Thomas Sounack, Shih-Cheng Huang, Zhihong Chen, Maya Varma, Steven~QH Truong, Chu~The Chuong, and Curtis~P. Langlotz.
\newblock Chexpert plus: Augmenting a large chest x-ray dataset with text radiology reports, patient demographics and additional image formats, 2024.
\newblock URL \url{https://arxiv.org/abs/2405.19538}.

\bibitem[Chen et~al.(2022)Chen, Pan, Zhang, Du, and Cui]{90chen2022vmeknet}
Weipeng Chen, Haiwei Pan, Kejia Zhang, Xin Du, and Qianna Cui.
\newblock Vmeknet: Visual memory and external knowledge based network for medical report generation.
\newblock In \emph{PRICAI 2022: Trends in Artificial Intelligence: 19th Pacific Rim International Conference on Artificial Intelligence, PRICAI 2022, Shanghai, China, November 10--13, 2022, Proceedings, Part I}, pages 188--201. Springer, 2022.

\bibitem[Chen et~al.(2020)Chen, Song, Chang, and Wan]{58chen2020generating}
Zhihong Chen, Yan Song, Tsung-Hui Chang, and Xiang Wan.
\newblock Generating radiology reports via memory-driven transformer.
\newblock In \emph{Proceedings of the 2020 Conference on Empirical Methods in Natural Language Processing (EMNLP)}, pages 1439--1449, 2020.

\bibitem[Chen et~al.(2021)Chen, Shen, Song, and Wan]{5chen2021cross}
Zhihong Chen, Yaling Shen, Yan Song, and Xiang Wan.
\newblock Cross-modal memory networks for radiology report generation.
\newblock In \emph{Proceedings of the 59th Annual Meeting of the Association for Computational Linguistics and the 11th International Joint Conference on Natural Language Processing (Volume 1: Long Papers)}, pages 5904--5914, 2021.

\bibitem[Chen et~al.(2024)Chen, Varma, Delbrouck, Paschali, Blankemeier, Van~Veen, Valanarasu, Youssef, Cohen, Reis, et~al.]{124chen2024chexagent}
Zhihong Chen, Maya Varma, Jean-Benoit Delbrouck, Magdalini Paschali, Louis Blankemeier, Dave Van~Veen, Jeya Maria~Jose Valanarasu, Alaa Youssef, Joseph~Paul Cohen, Eduardo~Pontes Reis, et~al.
\newblock Chexagent: Towards a foundation model for chest x-ray interpretation.
\newblock \emph{arXiv preprint arXiv:2401.12208}, 2024.

\bibitem[Chiang et~al.(2023)Chiang, Li, Lin, Sheng, Wu, Zhang, Zheng, Zhuang, Zhuang, Gonzalez, et~al.]{chiang2023vicuna}
Wei-Lin Chiang, Zhuohan Li, Zi~Lin, Ying Sheng, Zhanghao Wu, Hao Zhang, Lianmin Zheng, Siyuan Zhuang, Yonghao Zhuang, Joseph~E Gonzalez, et~al.
\newblock Vicuna: An open-source chatbot impressing gpt-4 with 90\%* chatgpt quality.
\newblock \emph{See https://vicuna. lmsys. org (accessed 14 April 2023)}, 2\penalty0 (3):\penalty0 6, 2023.

\bibitem[Cornia et~al.(2020)Cornia, Stefanini, Baraldi, and Cucchiara]{M2Transcornia2020meshed}
Marcella Cornia, Matteo Stefanini, Lorenzo Baraldi, and Rita Cucchiara.
\newblock Meshed-memory transformer for image captioning.
\newblock In \emph{Proceedings of the IEEE/CVF conference on computer vision and pattern recognition}, pages 10578--10587, 2020.

\bibitem[Dalla~Serra et~al.(2022)Dalla~Serra, Clackett, MacKinnon, Wang, Deligianni, Dalton, and O’Neil]{72dalla2022multimodal}
Francesco Dalla~Serra, William Clackett, Hamish MacKinnon, Chaoyang Wang, Fani Deligianni, Jeff Dalton, and Alison~Q O’Neil.
\newblock Multimodal generation of radiology reports using knowledge-grounded extraction of entities and relations.
\newblock In \emph{Proceedings of the 2nd Conference of the Asia-Pacific Chapter of the Association for Computational Linguistics and the 12th International Joint Conference on Natural Language Processing}, pages 615--624, 2022.

\bibitem[Dalla~Serra et~al.(2023{\natexlab{a}})Dalla~Serra, Wang, Deligianni, Dalton, and O'Neil]{109dalla2023controllable}
Francesco Dalla~Serra, Chaoyang Wang, Fani Deligianni, Jeff Dalton, and Alison~Q O'Neil.
\newblock Controllable chest x-ray report generation from longitudinal representations.
\newblock In \emph{The 2023 Conference on Empirical Methods in Natural Language Processing}, 2023{\natexlab{a}}.

\bibitem[Dalla~Serra et~al.(2023{\natexlab{b}})Dalla~Serra, Wang, Deligianni, Dalton, and O’Neil]{113dalla2023finding}
Francesco Dalla~Serra, Chaoyang Wang, Fani Deligianni, Jeffrey Dalton, and Alison~Q O’Neil.
\newblock Finding-aware anatomical tokens for chest x-ray automated reporting.
\newblock In \emph{International Workshop on Machine Learning in Medical Imaging}, pages 413--423. Springer, 2023{\natexlab{b}}.

\bibitem[Delbrouck et~al.(2022)Delbrouck, Chambon, Bluethgen, Tsai, Almusa, and Langlotz]{delbrouck2022improving}
Jean-Benoit Delbrouck, Pierre Chambon, Christian Bluethgen, Emily Tsai, Omar Almusa, and Curtis Langlotz.
\newblock Improving the factual correctness of radiology report generation with semantic rewards.
\newblock In \emph{Findings of the Association for Computational Linguistics: EMNLP 2022}, pages 4348--4360, 2022.

\bibitem[Delbrouck et~al.(2024)Delbrouck, Chambon, Chen, Varma, Johnston, Blankemeier, Van~Veen, Bui, Truong, and Langlotz]{delbrouck2024radgraph}
Jean-Benoit Delbrouck, Pierre Chambon, Zhihong Chen, Maya Varma, Andrew Johnston, Louis Blankemeier, Dave Van~Veen, Tan Bui, Steven Truong, and Curtis Langlotz.
\newblock Radgraph-xl: A large-scale expert-annotated dataset for entity and relation extraction from radiology reports.
\newblock In \emph{Findings of the Association for Computational Linguistics ACL 2024}, pages 12902--12915, 2024.

\bibitem[Demner-Fushman et~al.(2016)Demner-Fushman, Kohli, Rosenman, Shooshan, Rodriguez, Antani, Thoma, and McDonald]{IU-XRAYdemner2016preparing}
Dina Demner-Fushman, Marc~D Kohli, Marc~B Rosenman, Sonya~E Shooshan, Laritza Rodriguez, Sameer Antani, George~R Thoma, and Clement~J McDonald.
\newblock Preparing a collection of radiology examinations for distribution and retrieval.
\newblock \emph{Journal of the American Medical Informatics Association}, 23\penalty0 (2):\penalty0 304--310, 2016.

\bibitem[Doshi-Velez and Kim(2017)]{doshi2017towards}
Finale Doshi-Velez and Been Kim.
\newblock Towards a rigorous science of interpretable machine learning.
\newblock \emph{arXiv preprint arXiv:1702.08608}, 2017.

\bibitem[Du et~al.(2022)Du, Pan, Zhang, He, Bian, and Chen]{93du2022automatic}
Xin Du, Haiwei Pan, Kejia Zhang, Shuning He, Xiaofei Bian, and Weipeng Chen.
\newblock Automatic report generation method based on multiscale feature extraction and word attention network.
\newblock In \emph{Asia-Pacific Web (APWeb) and Web-Age Information Management (WAIM) Joint International Conference on Web and Big Data}, pages 520--528. Springer, 2022.

\bibitem[Endo et~al.(2021)Endo, Krishnan, Krishna, Ng, and Rajpurkar]{27endo2021retrieval}
Mark Endo, Rayan Krishnan, Viswesh Krishna, Andrew~Y Ng, and Pranav Rajpurkar.
\newblock Retrieval-based chest x-ray report generation using a pre-trained contrastive language-image model.
\newblock In \emph{Machine Learning for Health}, pages 209--219. PMLR, 2021.

\bibitem[Esser et~al.(2021)Esser, Rombach, and Ommer]{esser2021taming}
Patrick Esser, Robin Rombach, and Bjorn Ommer.
\newblock Taming transformers for high-resolution image synthesis.
\newblock In \emph{Proceedings of the IEEE/CVF conference on computer vision and pattern recognition}, pages 12873--12883, 2021.

\bibitem[Gajbhiye et~al.(2022)Gajbhiye, Nandedkar, and Faye]{80gajbhiye2022translating}
Gaurav~O Gajbhiye, Abhijeet~V Nandedkar, and Ibrahima Faye.
\newblock Translating medical image to radiological report: Adaptive multilevel multi-attention approach.
\newblock \emph{Computer Methods and Programs in Biomedicine}, 221:\penalty0 106853, 2022.

\bibitem[Gal and Ghahramani(2016)]{gal2016dropout}
Yarin Gal and Zoubin Ghahramani.
\newblock Dropout as a bayesian approximation: Representing model uncertainty in deep learning.
\newblock In \emph{international conference on machine learning}, pages 1050--1059. PMLR, 2016.

\bibitem[Gu et~al.(2024)Gu, Liu, Li, and Cai]{105gu2024complex}
Tiancheng Gu, Dongnan Liu, Zhiyuan Li, and Weidong Cai.
\newblock Complex organ mask guided radiology report generation.
\newblock In \emph{Proceedings of the IEEE/CVF Winter Conference on Applications of Computer Vision}, pages 7995--8004, 2024.

\bibitem[Guo et~al.(2024)Guo, Chai, Li, and Wang]{125guo2024llavaultra}
Xuechen Guo, Wenhao Chai, Shi-Yan Li, and Gaoang Wang.
\newblock {LL}a{VA}-ultra: Large chinese language and vision assistant for ultrasound.
\newblock In \emph{ACM Multimedia 2024}, 2024.
\newblock URL \url{https://openreview.net/forum?id=7ZYEoB71Vd}.

\bibitem[Han et~al.(2021)Han, Wei, Xi, Chen, Yin, and Li]{19han2021unifying}
Zhongyi Han, Benzheng Wei, Xiaoming Xi, Bo~Chen, Yilong Yin, and Shuo Li.
\newblock Unifying neural learning and symbolic reasoning for spinal medical report generation.
\newblock \emph{Medical image analysis}, 67:\penalty0 101872, 2021.

\bibitem[He et~al.(2016)He, Zhang, Ren, and Sun]{Resnethe2016deep}
Kaiming He, Xiangyu Zhang, Shaoqing Ren, and Jian Sun.
\newblock Deep residual learning for image recognition.
\newblock In \emph{Proceedings of the IEEE conference on computer vision and pattern recognition}, pages 770--778, 2016.

\bibitem[He et~al.(2020)He, Fan, Wu, Xie, and Girshick]{he2020momentum}
Kaiming He, Haoqi Fan, Yuxin Wu, Saining Xie, and Ross Girshick.
\newblock Momentum contrast for unsupervised visual representation learning.
\newblock In \emph{Proceedings of the IEEE/CVF conference on computer vision and pattern recognition}, pages 9729--9738, 2020.

\bibitem[Hochreiter and Schmidhuber(1997)]{hochreiter1997long}
Sepp Hochreiter and J{\"u}rgen Schmidhuber.
\newblock Long short-term memory.
\newblock \emph{Neural computation}, 9\penalty0 (8):\penalty0 1735--1780, 1997.

\bibitem[Hong and Kahn(2013)]{hong2013content}
Yi~Hong and Charles~E Kahn.
\newblock Content analysis of reporting templates and free-text radiology reports.
\newblock \emph{Journal of digital imaging}, 26:\penalty0 843--849, 2013.

\bibitem[Hosny et~al.(2018)Hosny, Parmar, Quackenbush, Schwartz, and Aerts]{hosny2018artificial}
Ahmed Hosny, Chintan Parmar, John Quackenbush, Lawrence~H Schwartz, and Hugo~JWL Aerts.
\newblock Artificial intelligence in radiology.
\newblock \emph{Nature Reviews Cancer}, 18\penalty0 (8):\penalty0 500--510, 2018.

\bibitem[Hou et~al.(2021{\natexlab{a}})Hou, Kaissis, Summers, and Kainz]{63hou2021ratchet}
Benjamin Hou, Georgios Kaissis, Ronald~M Summers, and Bernhard Kainz.
\newblock Ratchet: Medical transformer for chest x-ray diagnosis and reporting.
\newblock In \emph{Medical Image Computing and Computer Assisted Intervention--MICCAI 2021: 24th International Conference, Strasbourg, France, September 27--October 1, 2021, Proceedings, Part VII 24}, pages 293--303. Springer, 2021{\natexlab{a}}.

\bibitem[Hou et~al.(2021{\natexlab{b}})Hou, Zhao, Liu, Chang, and Hu]{47hou2021automatic}
Daibing Hou, Zijian Zhao, Yuying Liu, Faliang Chang, and Sanyuan Hu.
\newblock Automatic report generation for chest x-ray images via adversarial reinforcement learning.
\newblock \emph{IEEE Access}, 9:\penalty0 21236--21250, 2021{\natexlab{b}}.

\bibitem[Hou et~al.(2023)Hou, Xu, Cheng, Li, and Liu]{115hou2023organ}
Wenjun Hou, Kaishuai Xu, Yi~Cheng, Wenjie Li, and Jiang Liu.
\newblock Organ: observation-guided radiology report generation via tree reasoning.
\newblock \emph{arXiv preprint arXiv:2306.06466}, 2023.

\bibitem[Hu et~al.(2021)Hu, Shen, Wallis, Allen-Zhu, Li, Wang, Wang, and Chen]{hu2021lora}
Edward~J Hu, Yelong Shen, Phillip Wallis, Zeyuan Allen-Zhu, Yuanzhi Li, Shean Wang, Lu~Wang, and Weizhu Chen.
\newblock Lora: Low-rank adaptation of large language models.
\newblock \emph{arXiv preprint arXiv:2106.09685}, 2021.

\bibitem[Huang et~al.(2017)Huang, Liu, Van Der~Maaten, and Weinberger]{Densenethuang2017densely}
Gao Huang, Zhuang Liu, Laurens Van Der~Maaten, and Kilian~Q Weinberger.
\newblock Densely connected convolutional networks.
\newblock In \emph{Proceedings of the IEEE conference on computer vision and pattern recognition}, pages 4700--4708, 2017.

\bibitem[Huang et~al.(2021{\natexlab{a}})Huang, Wu, Yang, and Worring]{71huang2021deep}
Jia-Hong Huang, Ting-Wei Wu, Chao-Han~Huck Yang, and Marcel Worring.
\newblock Deep context-encoding network for retinal image captioning.
\newblock In \emph{2021 IEEE International Conference on Image Processing (ICIP)}, pages 3762--3766. IEEE, 2021{\natexlab{a}}.

\bibitem[Huang et~al.(2021{\natexlab{b}})Huang, Yang, Liu, Tian, Liu, Wu, Lin, Wang, Morikawa, Chang, et~al.]{18huang2021deepopht}
Jia-Hong Huang, C-H~Huck Yang, Fangyu Liu, Meng Tian, Yi-Chieh Liu, Ting-Wei Wu, I~Lin, Kang Wang, Hiromasa Morikawa, Hernghua Chang, et~al.
\newblock Deepopht: medical report generation for retinal images via deep models and visual explanation.
\newblock In \emph{Proceedings of the IEEE/CVF winter conference on applications of computer vision}, pages 2442--2452, 2021{\natexlab{b}}.

\bibitem[Huang et~al.(2022)Huang, Wu, Yang, Shi, Lin, Tegner, Worring, et~al.]{62huang2022non}
Jia-Hong Huang, Ting-Wei Wu, C-H~Huck Yang, Zenglin Shi, I~Lin, Jesper Tegner, Marcel Worring, et~al.
\newblock Non-local attention improves description generation for retinal images.
\newblock In \emph{Proceedings of the IEEE/CVF winter conference on applications of computer vision}, pages 1606--1615, 2022.

\bibitem[Huang et~al.(2023)Huang, Zhang, and Zhang]{21huang2023kiut}
Zhongzhen Huang, Xiaofan Zhang, and Shaoting Zhang.
\newblock Kiut: Knowledge-injected u-transformer for radiology report generation.
\newblock In \emph{Proceedings of the IEEE/CVF Conference on Computer Vision and Pattern Recognition}, pages 19809--19818, 2023.

\bibitem[Hyland et~al.(2023)Hyland, Bannur, Bouzid, Castro, Ranjit, Schwaighofer, P{\'e}rez-Garc{\'\i}a, Salvatelli, Srivastav, Thieme, et~al.]{122hyland2023maira}
Stephanie~L Hyland, Shruthi Bannur, Kenza Bouzid, Daniel~C Castro, Mercy Ranjit, Anton Schwaighofer, Fernando P{\'e}rez-Garc{\'\i}a, Valentina Salvatelli, Shaury Srivastav, Anja Thieme, et~al.
\newblock Maira-1: A specialised large multimodal model for radiology report generation.
\newblock \emph{arXiv preprint arXiv:2311.13668}, 2023.

\bibitem[Irvin et~al.(2019)Irvin, Rajpurkar, Ko, Yu, Ciurea-Ilcus, Chute, Marklund, Haghgoo, Ball, Shpanskaya, et~al.]{CheXpertirvin2019chexpert}
Jeremy Irvin, Pranav Rajpurkar, Michael Ko, Yifan Yu, Silviana Ciurea-Ilcus, Chris Chute, Henrik Marklund, Behzad Haghgoo, Robyn Ball, Katie Shpanskaya, et~al.
\newblock Chexpert: A large chest radiograph dataset with uncertainty labels and expert comparison.
\newblock In \emph{Proceedings of the AAAI conference on artificial intelligence}, volume~33, pages 590--597, 2019.

\bibitem[Jain et~al.(2021)Jain, Agrawal, Saporta, Truong, Duong, Bui, Chambon, Zhang, Lungren, Ng, et~al.]{Radgraphjain2021radgraph}
Saahil Jain, Ashwin Agrawal, Adriel Saporta, Steven~QH Truong, Du~Nguyen Duong, Tan Bui, Pierre Chambon, Yuhao Zhang, Matthew~P Lungren, Andrew~Y Ng, et~al.
\newblock Radgraph: Extracting clinical entities and relations from radiology reports.
\newblock \emph{arXiv preprint arXiv:2106.14463}, 2021.

\bibitem[Jeong et~al.(2024)Jeong, Tian, Li, Hartung, Adithan, Behzadi, Calle, Osayande, Pohlen, and Rajpurkar]{73jeong2024multimodal}
Jaehwan Jeong, Katherine Tian, Andrew Li, Sina Hartung, Subathra Adithan, Fardad Behzadi, Juan Calle, David Osayande, Michael Pohlen, and Pranav Rajpurkar.
\newblock Multimodal image-text matching improves retrieval-based chest x-ray report generation.
\newblock In \emph{Medical Imaging with Deep Learning}, pages 978--990. PMLR, 2024.

\bibitem[Jia et~al.(2021)Jia, Xiong, Zhang, Zhang, Suzanne, Zhu, and Tang]{53jia2021radiology}
Xing Jia, Yun Xiong, Jiawei Zhang, Yao Zhang, Blackley Suzanne, Yangyong Zhu, and Chunlei Tang.
\newblock Radiology report generation for rare diseases via few-shot transformer.
\newblock In \emph{IEEE International Conference on Bioinformatics and Biomedicine (BIBM)}, pages 1347--1352. IEEE, 2021.

\bibitem[Jia et~al.(2022)Jia, Xiong, Zhang, Zhang, Zhu, and Philip]{69jia2022few}
Xing Jia, Yun Xiong, Jiawei Zhang, Yao Zhang, Yangyong Zhu, and S~Yu Philip.
\newblock Few-shot radiology report generation via knowledge transfer and multi-modal alignment.
\newblock In \emph{2022 IEEE International Conference on Bioinformatics and Biomedicine (BIBM)}, pages 1574--1579. IEEE, 2022.

\bibitem[Jiang et~al.(2023)Jiang, Sablayrolles, Mensch, Bamford, Chaplot, Casas, Bressand, Lengyel, Lample, Saulnier, et~al.]{jiang2023mistral}
Albert~Q Jiang, Alexandre Sablayrolles, Arthur Mensch, Chris Bamford, Devendra~Singh Chaplot, Diego de~las Casas, Florian Bressand, Gianna Lengyel, Guillaume Lample, Lucile Saulnier, et~al.
\newblock Mistral 7b.
\newblock \emph{arXiv preprint arXiv:2310.06825}, 2023.

\bibitem[Jin et~al.(2024)Jin, Che, Lin, and Chen]{112jin2024promptmrg}
Haibo Jin, Haoxuan Che, Yi~Lin, and Hao Chen.
\newblock Promptmrg: Diagnosis-driven prompts for medical report generation.
\newblock In \emph{Proceedings of the AAAI Conference on Artificial Intelligence}, volume~38, pages 2607--2615, 2024.

\bibitem[Johnson et~al.(2019{\natexlab{a}})Johnson, Pollard, Berkowitz, Greenbaum, Lungren, Deng, Mark, and Horng]{MIMICjohnson2019mimic}
Alistair~EW Johnson, Tom~J Pollard, Seth~J Berkowitz, Nathaniel~R Greenbaum, Matthew~P Lungren, Chih-ying Deng, Roger~G Mark, and Steven Horng.
\newblock Mimic-cxr, a de-identified publicly available database of chest radiographs with free-text reports.
\newblock \emph{Scientific data}, 6\penalty0 (1):\penalty0 317, 2019{\natexlab{a}}.

\bibitem[Johnson et~al.(2019{\natexlab{b}})Johnson, Pollard, Greenbaum, Lungren, Deng, Peng, Lu, Mark, Berkowitz, and Horng]{MIMIC-CXR-JPGjohnson2019mimic-2}
Alistair~EW Johnson, Tom~J Pollard, Nathaniel~R Greenbaum, Matthew~P Lungren, Chih-ying Deng, Yifan Peng, Zhiyong Lu, Roger~G Mark, Seth~J Berkowitz, and Steven Horng.
\newblock Mimic-cxr-jpg, a large publicly available database of labeled chest radiographs.
\newblock \emph{arXiv preprint arXiv:1901.07042}, 2019{\natexlab{b}}.

\bibitem[Johri et~al.(2025)Johri, Jeong, Tran, Schlessinger, Wongvibulsin, Barnes, Zhou, Cai, Van~Allen, Kim, et~al.]{johri2025evaluation}
Shreya Johri, Jaehwan Jeong, Benjamin~A Tran, Daniel~I Schlessinger, Shannon Wongvibulsin, Leandra~A Barnes, Hong-Yu Zhou, Zhuo~Ran Cai, Eliezer~M Van~Allen, David Kim, et~al.
\newblock An evaluation framework for clinical use of large language models in patient interaction tasks.
\newblock \emph{Nature Medicine}, pages 1--10, 2025.

\bibitem[Kalil et~al.(2016)Kalil, Metersky, Klompas, Muscedere, Sweeney, Palmer, Napolitano, O'Grady, Bartlett, Carratal{\`a}, et~al.]{kalil2016management}
Andre~C Kalil, Mark~L Metersky, Michael Klompas, John Muscedere, Daniel~A Sweeney, Lucy~B Palmer, Lena~M Napolitano, Naomi~P O'Grady, John~G Bartlett, Jordi Carratal{\`a}, et~al.
\newblock Management of adults with hospital-acquired and ventilator-associated pneumonia: 2016 clinical practice guidelines by the infectious diseases society of america and the american thoracic society.
\newblock \emph{Clinical infectious diseases}, 63\penalty0 (5):\penalty0 e61--e111, 2016.

\bibitem[Kaur and Mittal(2022{\natexlab{a}})]{64kaur2022radiobert}
Navdeep Kaur and Ajay Mittal.
\newblock Radiobert: A deep learning-based system for medical report generation from chest x-ray images using contextual embeddings.
\newblock \emph{Journal of Biomedical Informatics}, 135:\penalty0 104220, 2022{\natexlab{a}}.

\bibitem[Kaur and Mittal(2022{\natexlab{b}})]{96kaur2022cadxreport}
Navdeep Kaur and Ajay Mittal.
\newblock Cadxreport: Chest x-ray report generation using co-attention mechanism and reinforcement learning.
\newblock \emph{Computers in Biology and Medicine}, 145:\penalty0 105498, 2022{\natexlab{b}}.

\bibitem[Kaur et~al.(2022)Kaur, Mittal, and Singh]{kaur2022methods}
Navdeep Kaur, Ajay Mittal, and Gurprem Singh.
\newblock Methods for automatic generation of radiological reports of chest radiographs: a comprehensive survey.
\newblock \emph{Multimedia Tools and Applications}, 81\penalty0 (10):\penalty0 13409--13439, 2022.

\bibitem[Kelly et~al.(2019)Kelly, Karthikesalingam, Suleyman, Corrado, and King]{kelly2019key}
Christopher~J Kelly, Alan Karthikesalingam, Mustafa Suleyman, Greg Corrado, and Dominic King.
\newblock Key challenges for delivering clinical impact with artificial intelligence.
\newblock \emph{BMC medicine}, 17:\penalty0 1--9, 2019.

\bibitem[Kingma and Welling(2013)]{kingma2013auto}
Diederik~P Kingma and Max Welling.
\newblock Auto-encoding variational bayes.
\newblock \emph{arXiv preprint arXiv:1312.6114}, 2013.

\bibitem[Kirillov et~al.(2023)Kirillov, Mintun, Ravi, Mao, Rolland, Gustafson, Xiao, Whitehead, Berg, Lo, et~al.]{kirillov2023segment}
Alexander Kirillov, Eric Mintun, Nikhila Ravi, Hanzi Mao, Chloe Rolland, Laura Gustafson, Tete Xiao, Spencer Whitehead, Alexander~C Berg, Wan-Yen Lo, et~al.
\newblock Segment anything.
\newblock In \emph{Proceedings of the IEEE/CVF International Conference on Computer Vision}, pages 4015--4026, 2023.

\bibitem[Kong et~al.(2022)Kong, Huang, Kuang, Zhu, and Wu]{91kong2022transq}
Ming Kong, Zhengxing Huang, Kun Kuang, Qiang Zhu, and Fei Wu.
\newblock Transq: Transformer-based semantic query for medical report generation.
\newblock In \emph{Medical Image Computing and Computer Assisted Intervention--MICCAI 2022: 25th International Conference, Singapore, September 18--22, 2022, Proceedings, Part VIII}, pages 610--620. Springer, 2022.

\bibitem[Lambert et~al.(2024)Lambert, Forbes, Doyle, Dehaene, and Dojat]{lambert2024trustworthy}
Benjamin Lambert, Florence Forbes, Senan Doyle, Harmonie Dehaene, and Michel Dojat.
\newblock Trustworthy clinical ai solutions: a unified review of uncertainty quantification in deep learning models for medical image analysis.
\newblock \emph{Artificial Intelligence in Medicine}, page 102830, 2024.

\bibitem[Langlotz(2006)]{RadLexlanglotz2006radlex}
Curtis~P Langlotz.
\newblock Radlex: a new method for indexing online educational materials, 2006.

\bibitem[Laserson et~al.(2018)Laserson, Lantsman, Cohen-Sfady, Tamir, Goz, Brestel, Bar, Atar, and Elnekave]{laserson2018textray}
Jonathan Laserson, Christine~Dan Lantsman, Michal Cohen-Sfady, Itamar Tamir, Eli Goz, Chen Brestel, Shir Bar, Maya Atar, and Eldad Elnekave.
\newblock Textray: Mining clinical reports to gain a broad understanding of chest x-rays.
\newblock In \emph{Medical Image Computing and Computer Assisted Intervention--MICCAI 2018: 21st International Conference, Granada, Spain, September 16-20, 2018, Proceedings, Part II 11}, pages 553--561. Springer, 2018.

\bibitem[Lee et~al.(2022)Lee, Cho, Park, Chae, and Kim]{56lee2022cross}
Hojun Lee, Hyunjun Cho, Jieun Park, Jinyeong Chae, and Jihie Kim.
\newblock Cross encoder-decoder transformer with global-local visual extractor for medical image captioning.
\newblock \emph{Sensors}, 22\penalty0 (4):\penalty0 1429, 2022.

\bibitem[Lee et~al.(2020)Lee, Yoon, Kim, Kim, Kim, So, and Kang]{BioBertlee2020biobert}
Jinhyuk Lee, Wonjin Yoon, Sungdong Kim, Donghyeon Kim, Sunkyu Kim, Chan~Ho So, and Jaewoo Kang.
\newblock Biobert: a pre-trained biomedical language representation model for biomedical text mining.
\newblock \emph{Bioinformatics}, 36\penalty0 (4):\penalty0 1234--1240, 2020.

\bibitem[Lee et~al.(2023)Lee, Kim, Chang, and Ye]{119lee2023llm}
Suhyeon Lee, Won~Jun Kim, Jinho Chang, and Jong~Chul Ye.
\newblock Llm-cxr: Instruction-finetuned llm for cxr image understanding and generation.
\newblock \emph{arXiv preprint arXiv:2305.11490}, 2023.

\bibitem[Li et~al.(2024)Li, Wong, Zhang, Usuyama, Liu, Yang, Naumann, Poon, and Gao]{127li2024llava}
Chunyuan Li, Cliff Wong, Sheng Zhang, Naoto Usuyama, Haotian Liu, Jianwei Yang, Tristan Naumann, Hoifung Poon, and Jianfeng Gao.
\newblock Llava-med: Training a large language-and-vision assistant for biomedicine in one day.
\newblock \emph{Advances in Neural Information Processing Systems}, 36, 2024.

\bibitem[Li et~al.(2022{\natexlab{a}})Li, Li, Hu, and Tao]{46li2022self}
Jun Li, Shibo Li, Ying Hu, and Huiren Tao.
\newblock A self-guided framework for radiology report generation.
\newblock In \emph{Medical Image Computing and Computer Assisted Intervention--MICCAI 2022: 25th International Conference, Singapore, September 18--22, 2022, Proceedings, Part VIII}, pages 588--598. Springer, 2022{\natexlab{a}}.

\bibitem[Li et~al.(2023{\natexlab{a}})Li, Li, Savarese, and Hoi]{li2023blip}
Junnan Li, Dongxu Li, Silvio Savarese, and Steven Hoi.
\newblock Blip-2: Bootstrapping language-image pre-training with frozen image encoders and large language models.
\newblock In \emph{International conference on machine learning}, pages 19730--19742. PMLR, 2023{\natexlab{a}}.

\bibitem[Li et~al.(2020)Li, Wang, Chang, and Liang]{COV-CTRli2020auxiliary}
Mingjie Li, Fuyu Wang, Xiaojun Chang, and Xiaodan Liang.
\newblock Auxiliary signal-guided knowledge encoder-decoder for medical report generation.
\newblock \emph{arXiv preprint arXiv:2006.03744}, 2020.

\bibitem[Li et~al.(2021)Li, Cai, Liu, Weng, Zhao, Wang, Chen, Liu, Pan, Li, et~al.]{FFA-IRli2021ffa}
Mingjie Li, Wenjia Cai, Rui Liu, Yuetian Weng, Xiaoyun Zhao, Cong Wang, Xin Chen, Zhong Liu, Caineng Pan, Mengke Li, et~al.
\newblock Ffa-ir: Towards an explainable and reliable medical report generation benchmark.
\newblock In \emph{Thirty-fifth Conference on Neural Information Processing Systems Datasets and Benchmarks Track (Round 2)}, 2021.

\bibitem[Li et~al.(2022{\natexlab{b}})Li, Cai, Verspoor, Pan, Liang, and Chang]{31li2022cross}
Mingjie Li, Wenjia Cai, Karin Verspoor, Shirui Pan, Xiaodan Liang, and Xiaojun Chang.
\newblock Cross-modal clinical graph transformer for ophthalmic report generation.
\newblock In \emph{Proceedings of the IEEE/CVF Conference on Computer Vision and Pattern Recognition}, pages 20656--20665, 2022{\natexlab{b}}.

\bibitem[Li et~al.(2023{\natexlab{b}})Li, Lin, Chen, Lin, Liang, and Chang]{20li2023dynamic}
Mingjie Li, Bingqian Lin, Zicong Chen, Haokun Lin, Xiaodan Liang, and Xiaojun Chang.
\newblock Dynamic graph enhanced contrastive learning for chest x-ray report generation.
\newblock In \emph{Proceedings of the IEEE/CVF Conference on Computer Vision and Pattern Recognition}, pages 3334--3343, 2023{\natexlab{b}}.

\bibitem[Li et~al.(2023{\natexlab{c}})Li, Liu, Wang, Chang, and Liang]{14li2023auxiliary}
Mingjie Li, Rui Liu, Fuyu Wang, Xiaojun Chang, and Xiaodan Liang.
\newblock Auxiliary signal-guided knowledge encoder-decoder for medical report generation.
\newblock \emph{World Wide Web}, 26\penalty0 (1):\penalty0 253--270, 2023{\natexlab{c}}.

\bibitem[Li et~al.(2023{\natexlab{d}})Li, Xu, Yuan, Chen, Zhang, Feng, Zhang, and Gao]{110li2023enhanced}
Qingqiu Li, Jilan Xu, Runtian Yuan, Mohan Chen, Yuejie Zhang, Rui Feng, Xiaobo Zhang, and Shang Gao.
\newblock Enhanced knowledge injection for radiology report generation.
\newblock In \emph{2023 IEEE International Conference on Bioinformatics and Biomedicine (BIBM)}, pages 2053--2058. IEEE, 2023{\natexlab{d}}.

\bibitem[Li et~al.(2023{\natexlab{e}})Li, Yang, Cheng, Zhu, Li, and Zou]{102li2023unify}
Yaowei Li, Bang Yang, Xuxin Cheng, Zhihong Zhu, Hongxiang Li, and Yuexian Zou.
\newblock Unify, align and refine: Multi-level semantic alignment for radiology report generation.
\newblock In \emph{Proceedings of the IEEE/CVF International Conference on Computer Vision}, pages 2863--2874, 2023{\natexlab{e}}.

\bibitem[Liao et~al.(2023)Liao, Liu, and Spasi{\'c}]{liao2023deep}
Yuxiang Liao, Hantao Liu, and Irena Spasi{\'c}.
\newblock Deep learning approaches to automatic radiology report generation: A systematic review.
\newblock \emph{Informatics in Medicine Unlocked}, page 101273, 2023.

\bibitem[Lin(2004)]{Rougelin2004rouge}
Chin-Yew Lin.
\newblock Rouge: A package for automatic evaluation of summaries.
\newblock In \emph{Text summarization branches out}, pages 74--81, 2004.

\bibitem[Lin et~al.(2023)Lin, Zhang, Shi, Xu, Tao, Wu, He, and Ge]{55lin2023contrastive}
Zhihong Lin, Donghao Zhang, Danli Shi, Renjing Xu, Qingyi Tao, Lin Wu, Mingguang He, and Zongyuan Ge.
\newblock Contrastive pre-training and linear interaction attention-based transformer for universal medical reports generation.
\newblock \emph{Journal of Biomedical Informatics}, page 104281, 2023.

\bibitem[Liu et~al.(2023{\natexlab{a}})Liu, Tian, and Song]{liu2023systematic}
Chang Liu, Yuanhe Tian, and Yan Song.
\newblock A systematic review of deep learning-based research on radiology report generation.
\newblock \emph{arXiv preprint arXiv:2311.14199}, 2023{\natexlab{a}}.

\bibitem[Liu et~al.(2021{\natexlab{a}})Liu, Ge, and Wu]{4liu2021competence}
Fenglin Liu, Shen Ge, and Xian Wu.
\newblock Competence-based multimodal curriculum learning for medical report generation.
\newblock In \emph{Proceedings of the 59th Annual Meeting of the Association for Computational Linguistics and the 11th International Joint Conference on Natural Language Processing (Volume 1: Long Papers)}, pages 3001--3012, 2021{\natexlab{a}}.

\bibitem[Liu et~al.(2021{\natexlab{b}})Liu, Wu, Ge, Fan, and Zou]{1liu2021exploring}
Fenglin Liu, Xian Wu, Shen Ge, Wei Fan, and Yuexian Zou.
\newblock Exploring and distilling posterior and prior knowledge for radiology report generation.
\newblock In \emph{Proceedings of the IEEE/CVF conference on computer vision and pattern recognition}, pages 13753--13762, 2021{\natexlab{b}}.

\bibitem[Liu et~al.(2021{\natexlab{c}})Liu, Yin, Wu, Ge, Zhang, and Sun]{52ma2021contrastive}
Fenglin Liu, Changchang Yin, Xian Wu, Shen Ge, Ping Zhang, and Xu~Sun.
\newblock Contrastive attention for automatic chest x-ray report generation.
\newblock In \emph{Findings of the Association for Computational Linguistics: ACL-IJCNLP}, page 269–280, 2021{\natexlab{c}}.

\bibitem[Liu et~al.(2021{\natexlab{d}})Liu, You, Wu, Ge, Sun, et~al.]{2liu2021auto}
Fenglin Liu, Chenyu You, Xian Wu, Shen Ge, Xu~Sun, et~al.
\newblock Auto-encoding knowledge graph for unsupervised medical report generation.
\newblock \emph{Advances in Neural Information Processing Systems}, 34:\penalty0 16266--16279, 2021{\natexlab{d}}.

\bibitem[Liu et~al.(2021{\natexlab{e}})Liu, Liao, Wang, Zhang, Zhang, Liang, Wan, Li, Li, Zhang, et~al.]{24liu2021medical}
Guangyi Liu, Yinghong Liao, Fuyu Wang, Bin Zhang, Lu~Zhang, Xiaodan Liang, Xiang Wan, Shaolin Li, Zhen Li, Shuixing Zhang, et~al.
\newblock Medical-vlbert: Medical visual language bert for covid-19 ct report generation with alternate learning.
\newblock \emph{IEEE Transactions on Neural Networks and Learning Systems}, 32\penalty0 (9):\penalty0 3786--3797, 2021{\natexlab{e}}.

\bibitem[Liu et~al.(2019)Liu, Hsu, McDermott, Boag, Weng, Szolovits, and Ghassemi]{CEliu2019clinically}
Guanxiong Liu, Tzu-Ming~Harry Hsu, Matthew McDermott, Willie Boag, Wei-Hung Weng, Peter Szolovits, and Marzyeh Ghassemi.
\newblock Clinically accurate chest x-ray report generation.
\newblock In \emph{Machine Learning for Healthcare Conference}, pages 249--269. PMLR, 2019.

\bibitem[Liu et~al.(2024)Liu, Li, Wu, and Lee]{liu2024visual}
Haotian Liu, Chunyuan Li, Qingyang Wu, and Yong~Jae Lee.
\newblock Visual instruction tuning.
\newblock \emph{Advances in neural information processing systems}, 36, 2024.

\bibitem[Liu et~al.(2023{\natexlab{b}})Liu, Yuan, Fu, Jiang, Hayashi, and Neubig]{liu2023pre}
Pengfei Liu, Weizhe Yuan, Jinlan Fu, Zhengbao Jiang, Hiroaki Hayashi, and Graham Neubig.
\newblock Pre-train, prompt, and predict: A systematic survey of prompting methods in natural language processing.
\newblock \emph{ACM Computing Surveys}, 55\penalty0 (9):\penalty0 1--35, 2023{\natexlab{b}}.

\bibitem[Liu et~al.(2023{\natexlab{c}})Liu, Zhu, Zheng, Zhao, He, and Zhao]{106liu2023observation}
Zhizhe Liu, Zhenfeng Zhu, Shuai Zheng, Yawei Zhao, Kunlun He, and Yao Zhao.
\newblock From observation to concept: A flexible multi-view paradigm for medical report generation.
\newblock \emph{IEEE Transactions on Multimedia}, 2023{\natexlab{c}}.

\bibitem[McDonald et~al.(2015)McDonald, Schwartz, Eckel, Diehn, Hunt, Bartholmai, Erickson, and Kallmes]{mcdonald2015effects}
Robert~J McDonald, Kara~M Schwartz, Laurence~J Eckel, Felix~E Diehn, Christopher~H Hunt, Brian~J Bartholmai, Bradley~J Erickson, and David~F Kallmes.
\newblock The effects of changes in utilization and technological advancements of cross-sectional imaging on radiologist workload.
\newblock \emph{Academic radiology}, 22\penalty0 (9):\penalty0 1191--1198, 2015.

\bibitem[Messina et~al.(2022)Messina, Pino, Parra, Soto, Besa, Uribe, And{\'\i}a, Tejos, Prieto, and Capurro]{messina2022survey}
Pablo Messina, Pablo Pino, Denis Parra, Alvaro Soto, Cecilia Besa, Sergio Uribe, Marcelo And{\'\i}a, Cristian Tejos, Claudia Prieto, and Daniel Capurro.
\newblock A survey on deep learning and explainability for automatic report generation from medical images.
\newblock \emph{ACM Computing Surveys (CSUR)}, 54\penalty0 (10s):\penalty0 1--40, 2022.

\bibitem[Mirikharaji et~al.(2023)Mirikharaji, Abhishek, Bissoto, Barata, Avila, Valle, Celebi, and Hamarneh]{mirikharaji2023survey}
Zahra Mirikharaji, Kumar Abhishek, Alceu Bissoto, Catarina Barata, Sandra Avila, Eduardo Valle, M~Emre Celebi, and Ghassan Hamarneh.
\newblock A survey on deep learning for skin lesion segmentation.
\newblock \emph{Medical Image Analysis}, page 102863, 2023.

\bibitem[Miura et~al.(2021)Miura, Zhang, Tsai, Langlotz, and Jurafsky]{7miura2021improving}
Yasuhide Miura, Yuhao Zhang, Emily Tsai, Curtis Langlotz, and Dan Jurafsky.
\newblock Improving factual completeness and consistency of image-to-text radiology report generation.
\newblock In \emph{Proceedings of the 2021 Conference of the North American Chapter of the Association for Computational Linguistics: Human Language Technologies}, pages 5288--5304, 2021.

\bibitem[Mohsan et~al.(2022)Mohsan, Akram, Rasool, Alghamdi, Baqai, and Abbas]{38mohsan2022vision}
Mashood~Mohammad Mohsan, Muhammad~Usman Akram, Ghulam Rasool, Norah~Saleh Alghamdi, Muhammad Abdullah~Aamer Baqai, and Muhammad Abbas.
\newblock Vision transformer and language model based radiology report generation.
\newblock \emph{IEEE Access}, 11:\penalty0 1814--1824, 2022.

\bibitem[Moon et~al.(2022)Moon, Lee, Shin, Kim, and Choi]{61moon2022multi}
Jong~Hak Moon, Hyungyung Lee, Woncheol Shin, Young-Hak Kim, and Edward Choi.
\newblock Multi-modal understanding and generation for medical images and text via vision-language pre-training.
\newblock \emph{IEEE Journal of Biomedical and Health Informatics}, 26\penalty0 (12):\penalty0 6070--6080, 2022.

\bibitem[Najdenkoska et~al.(2021)Najdenkoska, Zhen, Worring, and Shao]{29najdenkoska2021variational}
Ivona Najdenkoska, Xiantong Zhen, Marcel Worring, and Ling Shao.
\newblock Variational topic inference for chest x-ray report generation.
\newblock In \emph{Medical Image Computing and Computer Assisted Intervention--MICCAI 2021: 24th International Conference, Strasbourg, France, September 27--October 1, 2021, Proceedings, Part III 24}, pages 625--635. Springer, 2021.

\bibitem[Najdenkoska et~al.(2022)Najdenkoska, Zhen, Worring, and Shao]{29_2najdenkoska2022uncertainty}
Ivona Najdenkoska, Xiantong Zhen, Marcel Worring, and Ling Shao.
\newblock Uncertainty-aware report generation for chest x-rays by variational topic inference.
\newblock \emph{Medical Image Analysis}, 82:\penalty0 102603, 2022.

\bibitem[Neumann et~al.(2019)Neumann, King, Beltagy, and Ammar]{neumann2019scispacy}
Mark Neumann, Daniel King, Iz~Beltagy, and Waleed Ammar.
\newblock Scispacy: fast and robust models for biomedical natural language processing.
\newblock \emph{arXiv preprint arXiv:1902.07669}, 2019.

\bibitem[Nobel et~al.(2022)Nobel, van Geel, and Robben]{nobel2022structured}
J~Martijn Nobel, Koos van Geel, and Simon~GF Robben.
\newblock Structured reporting in radiology: a systematic review to explore its potential.
\newblock \emph{European radiology}, pages 1--18, 2022.

\bibitem[Ostmeier et~al.(2024)Ostmeier, Xu, Chen, Varma, Blankemeier, Bluethgen, Michalson, Moseley, Langlotz, Chaudhari, et~al.]{ostmeier2024green}
Sophie Ostmeier, Justin Xu, Zhihong Chen, Maya Varma, Louis Blankemeier, Christian Bluethgen, Arne~Edward Michalson, Michael Moseley, Curtis Langlotz, Akshay~S Chaudhari, et~al.
\newblock Green: Generative radiology report evaluation and error notation.
\newblock \emph{arXiv preprint arXiv:2405.03595}, 2024.

\bibitem[Pahwa et~al.(2021)Pahwa, Mehta, Kapadia, Jain, and Luthra]{16pahwa2021medskip}
Esha Pahwa, Dwij Mehta, Sanjeet Kapadia, Devansh Jain, and Achleshwar Luthra.
\newblock Medskip: Medical report generation using skip connections and integrated attention.
\newblock In \emph{Proceedings of the IEEE/CVF International Conference on Computer Vision}, pages 3409--3415, 2021.

\bibitem[Pandey et~al.(2021)Pandey, Paliwal, Dhall, Subramanian, and Mahapatra]{45pandey2021explains}
Abhineet Pandey, Bhawna Paliwal, Abhinav Dhall, Ramanathan Subramanian, and Dwarikanath Mahapatra.
\newblock This explains that: Congruent image--report generation for explainable medical image analysis with cyclic generative adversarial networks.
\newblock In \emph{Interpretability of Machine Intelligence in Medical Image Computing, and Topological Data Analysis and Its Applications for Medical Data: 4th International Workshop, iMIMIC 2021, and 1st International Workshop, TDA4MedicalData 2021, Held in Conjunction with MICCAI 2021, Strasbourg, France, September 27, 2021, Proceedings 4}, pages 34--43. Springer, 2021.

\bibitem[Pang et~al.(2023)Pang, Li, and Zhao]{pang2023survey}
Ting Pang, Peigao Li, and Lijie Zhao.
\newblock A survey on automatic generation of medical imaging reports based on deep learning.
\newblock \emph{BioMedical Engineering OnLine}, 22\penalty0 (1):\penalty0 48, 2023.

\bibitem[Papineni et~al.(2002)Papineni, Roukos, Ward, and Zhu]{BLEUpapineni2002bleu}
Kishore Papineni, Salim Roukos, Todd Ward, and Wei-Jing Zhu.
\newblock Bleu: a method for automatic evaluation of machine translation.
\newblock In \emph{Proceedings of the 40th annual meeting of the Association for Computational Linguistics}, pages 311--318, 2002.

\bibitem[Pellegrini et~al.(2023{\natexlab{a}})Pellegrini, Keicher, {\"O}zsoy, and Navab]{108pellegrini2023rad}
Chantal Pellegrini, Matthias Keicher, Ege {\"O}zsoy, and Nassir Navab.
\newblock Rad-restruct: A novel vqa benchmark and method for structured radiology reporting.
\newblock In \emph{International Conference on Medical Image Computing and Computer-Assisted Intervention}, pages 409--419. Springer, 2023{\natexlab{a}}.

\bibitem[Pellegrini et~al.(2023{\natexlab{b}})Pellegrini, {\"O}zsoy, Busam, Navab, and Keicher]{118pellegrini2023radialog}
Chantal Pellegrini, Ege {\"O}zsoy, Benjamin Busam, Nassir Navab, and Matthias Keicher.
\newblock Radialog: A large vision-language model for radiology report generation and conversational assistance.
\newblock \emph{arXiv preprint arXiv:2311.18681}, 2023{\natexlab{b}}.

\bibitem[P{\'e}rez-Garc{\'\i}a et~al.(2024)P{\'e}rez-Garc{\'\i}a, Sharma, Bond-Taylor, Bouzid, Salvatelli, Ilse, Bannur, Castro, Schwaighofer, Lungren, et~al.]{perez2024rad}
Fernando P{\'e}rez-Garc{\'\i}a, Harshita Sharma, Sam Bond-Taylor, Kenza Bouzid, Valentina Salvatelli, Maximilian Ilse, Shruthi Bannur, Daniel~C Castro, Anton Schwaighofer, Matthew~P Lungren, et~al.
\newblock Rad-dino: Exploring scalable medical image encoders beyond text supervision.
\newblock \emph{arXiv preprint arXiv:2401.10815}, 2024.

\bibitem[Pino et~al.(2021)Pino, Parra, Besa, and Lagos]{49pino2021clinically}
Pablo Pino, Denis Parra, Cecilia Besa, and Claudio Lagos.
\newblock Clinically correct report generation from chest x-rays using templates.
\newblock In \emph{Machine Learning in Medical Imaging: 12th International Workshop, MLMI 2021, Held in Conjunction with MICCAI 2021, Strasbourg, France, September 27, 2021, Proceedings 12}, pages 654--663. Springer, 2021.

\bibitem[Qin and Song(2022)]{32qin2022reinforced}
Han Qin and Yan Song.
\newblock Reinforced cross-modal alignment for radiology report generation.
\newblock In \emph{Findings of the Association for Computational Linguistics: ACL 2022}, pages 448--458, 2022.

\bibitem[Radford et~al.(2021)Radford, Kim, Hallacy, Ramesh, Goh, Agarwal, Sastry, Askell, Mishkin, Clark, et~al.]{CLIPradford2021learning}
Alec Radford, Jong~Wook Kim, Chris Hallacy, Aditya Ramesh, Gabriel Goh, Sandhini Agarwal, Girish Sastry, Amanda Askell, Pamela Mishkin, Jack Clark, et~al.
\newblock Learning transferable visual models from natural language supervision.
\newblock In \emph{International conference on machine learning}, pages 8748--8763. PMLR, 2021.

\bibitem[Ramesh et~al.(2021)Ramesh, Pavlov, Goh, Gray, Voss, Radford, Chen, and Sutskever]{ramesh2021zero}
Aditya Ramesh, Mikhail Pavlov, Gabriel Goh, Scott Gray, Chelsea Voss, Alec Radford, Mark Chen, and Ilya Sutskever.
\newblock Zero-shot text-to-image generation.
\newblock In \emph{International conference on machine learning}, pages 8821--8831. Pmlr, 2021.

\bibitem[Ramesh et~al.(2022)Ramesh, Chi, and Rajpurkar]{12ramesh2022improving}
Vignav Ramesh, Nathan~A Chi, and Pranav Rajpurkar.
\newblock Improving radiology report generation systems by removing hallucinated references to non-existent priors.
\newblock In \emph{Machine Learning for Health}, pages 456--473. PMLR, 2022.

\bibitem[Raoof et~al.(2012)Raoof, Feigin, Sung, Raoof, Irugulpati, and Rosenow~III]{raoof2012interpretation}
Suhail Raoof, David Feigin, Arthur Sung, Sabiha Raoof, Lavanya Irugulpati, and Edward~C Rosenow~III.
\newblock Interpretation of plain chest roentgenogram.
\newblock \emph{Chest}, 141\penalty0 (2):\penalty0 545--558, 2012.

\bibitem[Rimmer(2017)]{rimmer2017radiologist}
Abi Rimmer.
\newblock Radiologist shortage leaves patient care at risk, warns royal college.
\newblock \emph{BMJ: British Medical Journal (Online)}, 359, 2017.

\bibitem[Saab et~al.(2024)Saab, Tu, Weng, Tanno, Stutz, Wulczyn, Zhang, Strother, Park, Vedadi, et~al.]{128saab2024capabilities}
Khaled Saab, Tao Tu, Wei-Hung Weng, Ryutaro Tanno, David Stutz, Ellery Wulczyn, Fan Zhang, Tim Strother, Chunjong Park, Elahe Vedadi, et~al.
\newblock Capabilities of gemini models in medicine.
\newblock \emph{arXiv preprint arXiv:2404.18416}, 2024.

\bibitem[Selivanov et~al.(2023)Selivanov, Rogov, Chesakov, Shelmanov, Fedulova, and Dylov]{60selivanov2023medical}
Alexander Selivanov, Oleg~Y Rogov, Daniil Chesakov, Artem Shelmanov, Irina Fedulova, and Dmitry~V Dylov.
\newblock Medical image captioning via generative pretrained transformers.
\newblock \emph{Scientific Reports}, 13\penalty0 (1):\penalty0 4171, 2023.

\bibitem[Selvaraju et~al.(2017)Selvaraju, Cogswell, Das, Vedantam, Parikh, and Batra]{selvaraju2017grad}
Ramprasaath~R Selvaraju, Michael Cogswell, Abhishek Das, Ramakrishna Vedantam, Devi Parikh, and Dhruv Batra.
\newblock Grad-cam: Visual explanations from deep networks via gradient-based localization.
\newblock In \emph{Proceedings of the IEEE international conference on computer vision}, pages 618--626, 2017.

\bibitem[Shamshad et~al.(2023)Shamshad, Khan, Zamir, Khan, Hayat, Khan, and Fu]{shamshad2023transformers}
Fahad Shamshad, Salman Khan, Syed~Waqas Zamir, Muhammad~Haris Khan, Munawar Hayat, Fahad~Shahbaz Khan, and Huazhu Fu.
\newblock Transformers in medical imaging: A survey.
\newblock \emph{Medical Image Analysis}, page 102802, 2023.

\bibitem[Shetty et~al.(2023)Shetty, VS, and Mahale]{68shetty2023cross}
Shashank Shetty, Ananthanarayana VS, and Ajit Mahale.
\newblock Cross-modal deep learning-based clinical recommendation system for radiology report generation from chest x-rays.
\newblock \emph{International Journal of Engineering}, 2023.

\bibitem[Singh et~al.(2021)Singh, Karimi, Ho-Shon, and Hamey]{87singh2021show}
Sonit Singh, Sarvnaz Karimi, Kevin Ho-Shon, and Len Hamey.
\newblock Show, tell and summarise: learning to generate and summarise radiology findings from medical images.
\newblock \emph{Neural Computing and Applications}, 33:\penalty0 7441--7465, 2021.

\bibitem[Sirshar et~al.(2022)Sirshar, Paracha, Akram, Alghamdi, Zaidi, and Fatima]{23sirshar2022attention}
Mehreen Sirshar, Muhammad Faheem~Khalil Paracha, Muhammad~Usman Akram, Norah~Saleh Alghamdi, Syeda Zainab~Yousuf Zaidi, and Tatheer Fatima.
\newblock Attention based automated radiology report generation using cnn and lstm.
\newblock \emph{Plos one}, 17\penalty0 (1):\penalty0 e0262209, 2022.

\bibitem[Smit et~al.(2020{\natexlab{a}})Smit, Jain, Rajpurkar, Pareek, Ng, and Lungren]{CheXbertsmit2020combining}
Akshay Smit, Saahil Jain, Pranav Rajpurkar, Anuj Pareek, Andrew~Y Ng, and Matthew Lungren.
\newblock Combining automatic labelers and expert annotations for accurate radiology report labeling using bert.
\newblock In \emph{Proceedings of the 2020 Conference on Empirical Methods in Natural Language Processing (EMNLP)}, pages 1500--1519, 2020{\natexlab{a}}.

\bibitem[Smit et~al.(2020{\natexlab{b}})Smit, Jain, Rajpurkar, Pareek, Ng, and Lungren]{smit2020chexbert}
Akshay Smit, Saahil Jain, Pranav Rajpurkar, Anuj Pareek, Andrew~Y Ng, and Matthew~P Lungren.
\newblock Chexbert: combining automatic labelers and expert annotations for accurate radiology report labeling using bert.
\newblock \emph{arXiv preprint arXiv:2004.09167}, 2020{\natexlab{b}}.

\bibitem[Song et~al.(2022)Song, Zhang, Ji, Liu, and Wei]{66song2022cross}
Xiao Song, Xiaodan Zhang, Junzhong Ji, Ying Liu, and Pengxu Wei.
\newblock Cross-modal contrastive attention model for medical report generation.
\newblock In \emph{Proceedings of the 29th International Conference on Computational Linguistics}, pages 2388--2397, 2022.

\bibitem[Soviany et~al.(2022)Soviany, Ionescu, Rota, and Sebe]{soviany2022curriculum}
Petru Soviany, Radu~Tudor Ionescu, Paolo Rota, and Nicu Sebe.
\newblock Curriculum learning: A survey.
\newblock \emph{International Journal of Computer Vision}, 130\penalty0 (6):\penalty0 1526--1565, 2022.

\bibitem[Sun et~al.(2022)Sun, Wei, Wang, and Zheng]{48sun2022lesion}
Jinghan Sun, Dong Wei, Liansheng Wang, and Yefeng Zheng.
\newblock Lesion guided explainable few weak-shot medical report generation.
\newblock In \emph{Medical Image Computing and Computer Assisted Intervention--MICCAI 2022: 25th International Conference, Singapore, September 18--22, 2022, Proceedings, Part V}, pages 615--625. Springer, 2022.

\bibitem[Sun et~al.(2023)Sun, Fang, Wu, Wang, and Cao]{sun2023eva}
Quan Sun, Yuxin Fang, Ledell Wu, Xinlong Wang, and Yue Cao.
\newblock Eva-clip: Improved training techniques for clip at scale.
\newblock \emph{arXiv preprint arXiv:2303.15389}, 2023.

\bibitem[Sun et~al.(2019)Sun, Deng, Nie, and Tang]{Rotatesun2019rotate}
Zhiqing Sun, Zhi-Hong Deng, Jian-Yun Nie, and Jian Tang.
\newblock Rotate: Knowledge graph embedding by relational rotation in complex space.
\newblock \emph{arXiv preprint arXiv:1902.10197}, 2019.

\bibitem[Tanida et~al.(2023)Tanida, M{\"u}ller, Kaissis, and Rueckert]{13tanida2023interactive}
Tim Tanida, Philip M{\"u}ller, Georgios Kaissis, and Daniel Rueckert.
\newblock Interactive and explainable region-guided radiology report generation.
\newblock In \emph{Proceedings of the IEEE/CVF Conference on Computer Vision and Pattern Recognition}, pages 7433--7442, 2023.

\bibitem[Tanwani et~al.(2022)Tanwani, Barral, and Freedman]{86tanwani2022repsnet}
Ajay~K Tanwani, Joelle Barral, and Daniel Freedman.
\newblock Repsnet: Combining vision with language for automated medical reports.
\newblock In \emph{Medical Image Computing and Computer Assisted Intervention--MICCAI 2022: 25th International Conference, Singapore, September 18--22, 2022, Proceedings, Part V}, pages 714--724. Springer, 2022.

\bibitem[Topol(2019)]{I1topol2019deep}
Eric Topol.
\newblock \emph{Deep medicine: how artificial intelligence can make healthcare human again}.
\newblock Hachette UK, 2019.

\bibitem[Touvron et~al.(2023)Touvron, Martin, Stone, Albert, Almahairi, Babaei, Bashlykov, Batra, Bhargava, Bhosale, et~al.]{touvron2023llama}
Hugo Touvron, Louis Martin, Kevin Stone, Peter Albert, Amjad Almahairi, Yasmine Babaei, Nikolay Bashlykov, Soumya Batra, Prajjwal Bhargava, Shruti Bhosale, et~al.
\newblock Llama 2: Open foundation and fine-tuned chat models.
\newblock \emph{arXiv preprint arXiv:2307.09288}, 2023.

\bibitem[Tu et~al.(2024)Tu, Azizi, Driess, Schaekermann, Amin, Chang, Carroll, Lau, Tanno, Ktena, et~al.]{multimoda1tu2024towards}
Tao Tu, Shekoofeh Azizi, Danny Driess, Mike Schaekermann, Mohamed Amin, Pi-Chuan Chang, Andrew Carroll, Charles Lau, Ryutaro Tanno, Ira Ktena, et~al.
\newblock Towards generalist biomedical ai.
\newblock \emph{NEJM AI}, 1\penalty0 (3):\penalty0 AIoa2300138, 2024.

\bibitem[Van~Miltenburg et~al.(2018)Van~Miltenburg, Elliott, and Vossen]{novelvan2018measuring}
Emiel Van~Miltenburg, Desmond Elliott, and Piek Vossen.
\newblock Measuring the diversity of automatic image descriptions.
\newblock In \emph{Proceedings of the 27th International Conference on Computational Linguistics}, pages 1730--1741, 2018.

\bibitem[Vaswani et~al.(2017)Vaswani, Shazeer, Parmar, Uszkoreit, Jones, Gomez, Kaiser, and Polosukhin]{vaswani2017attention}
Ashish Vaswani, Noam Shazeer, Niki Parmar, Jakob Uszkoreit, Llion Jones, Aidan~N Gomez, {\L}ukasz Kaiser, and Illia Polosukhin.
\newblock Attention is all you need.
\newblock \emph{Advances in neural information processing systems}, 30, 2017.

\bibitem[Vedantam et~al.(2015)Vedantam, Lawrence~Zitnick, and Parikh]{Cidervedantam2015cider}
Ramakrishna Vedantam, C~Lawrence~Zitnick, and Devi Parikh.
\newblock Cider: Consensus-based image description evaluation.
\newblock In \emph{Proceedings of the IEEE conference on computer vision and pattern recognition}, pages 4566--4575, 2015.

\bibitem[Wang et~al.(2022{\natexlab{a}})Wang, Bhalerao, and He]{33wang2022cross}
Jun Wang, Abhir Bhalerao, and Yulan He.
\newblock Cross-modal prototype driven network for radiology report generation.
\newblock In \emph{Computer Vision--ECCV 2022: 17th European Conference, Tel Aviv, Israel, October 23--27, 2022, Proceedings, Part XXXV}, pages 563--579. Springer, 2022{\natexlab{a}}.

\bibitem[Wang et~al.(2024{\natexlab{a}})Wang, Bhalerao, Yin, See, and He]{98wang2024camanet}
Jun Wang, Abhir Bhalerao, Terry Yin, Simon See, and Yulan He.
\newblock Camanet: class activation map guided attention network for radiology report generation.
\newblock \emph{IEEE Journal of Biomedical and Health Informatics}, 2024{\natexlab{a}}.

\bibitem[Wang et~al.(2022{\natexlab{b}})Wang, Ning, Lu, Wei, Zheng, and Chen]{82wang2022inclusive}
Lin Wang, Munan Ning, Donghuan Lu, Dong Wei, Yefeng Zheng, and Jie Chen.
\newblock An inclusive task-aware framework for radiology report generation.
\newblock In \emph{Medical Image Computing and Computer Assisted Intervention--MICCAI 2022: 25th International Conference, Singapore, September 18--22, 2022, Proceedings, Part VIII}, pages 568--577. Springer, 2022{\natexlab{b}}.

\bibitem[Wang et~al.(2023{\natexlab{a}})Wang, Zhao, Ouyang, Wang, and Shen]{54wang2023chatcad}
Sheng Wang, Zihao Zhao, Xi~Ouyang, Qian Wang, and Dinggang Shen.
\newblock Chatcad: Interactive computer-aided diagnosis on medical image using large language models.
\newblock \emph{arXiv preprint arXiv:2302.07257}, 2023{\natexlab{a}}.

\bibitem[Wang et~al.(2023{\natexlab{b}})Wang, Peng, Liu, and Peng]{107wang2023fine}
Siyuan Wang, Bo~Peng, Yichao Liu, and Qi~Peng.
\newblock Fine-grained medical vision-language representation learning for radiology report generation.
\newblock In \emph{Proceedings of the 2023 Conference on Empirical Methods in Natural Language Processing}, pages 15949--15956, 2023{\natexlab{b}}.

\bibitem[Wang et~al.(2022{\natexlab{c}})Wang, Tang, Lin, Shih, Ding, and Peng]{30wang2022prior}
Song Wang, Liyan Tang, Mingquan Lin, George Shih, Ying Ding, and Yifan Peng.
\newblock Prior knowledge enhances radiology report generation.
\newblock In \emph{AMIA Annual Symposium Proceedings}, volume 2022, page 486. American Medical Informatics Association, 2022{\natexlab{c}}.

\bibitem[Wang et~al.(2024{\natexlab{b}})Wang, Lin, Xu, Dong, Luo, Tian, Shi, Huang, Zhang, Fan, et~al.]{104wang2024trust}
Yixin Wang, Zihao Lin, Zhe Xu, Haoyu Dong, Jie Luo, Jiang Tian, Zhongchao Shi, Lifu Huang, Yang Zhang, Jianping Fan, et~al.
\newblock Trust it or not: Confidence-guided automatic radiology report generation.
\newblock \emph{Neurocomputing}, page 127374, 2024{\natexlab{b}}.

\bibitem[Wang et~al.(2023{\natexlab{c}})Wang, Wang, Liu, Gao, Zhang, and Wang]{101wang2023self}
Yuhao Wang, Kai Wang, Xiaohong Liu, Tianrun Gao, Jingyue Zhang, and Guangyu Wang.
\newblock Self adaptive global-local feature enhancement for radiology report generation.
\newblock In \emph{2023 IEEE International Conference on Image Processing (ICIP)}, pages 2275--2279. IEEE, 2023{\natexlab{c}}.

\bibitem[Wang et~al.(2021)Wang, Zhou, Wang, and Li]{25wang2021self}
Zhanyu Wang, Luping Zhou, Lei Wang, and Xiu Li.
\newblock A self-boosting framework for automated radiographic report generation.
\newblock In \emph{Proceedings of the IEEE/CVF Conference on Computer Vision and Pattern Recognition}, pages 2433--2442, 2021.

\bibitem[Wang et~al.(2022{\natexlab{d}})Wang, Han, Wang, Li, and Zhou]{51wang2022automated}
Zhanyu Wang, Hongwei Han, Lei Wang, Xiu Li, and Luping Zhou.
\newblock Automated radiographic report generation purely on transformer: A multicriteria supervised approach.
\newblock \emph{IEEE Transactions on Medical Imaging}, 41\penalty0 (10):\penalty0 2803--2813, 2022{\natexlab{d}}.

\bibitem[Wang et~al.(2022{\natexlab{e}})Wang, Tang, Wang, Li, and Zhou]{92wang2022medical}
Zhanyu Wang, Mingkang Tang, Lei Wang, Xiu Li, and Luping Zhou.
\newblock A medical semantic-assisted transformer for radiographic report generation.
\newblock In \emph{Medical Image Computing and Computer Assisted Intervention--MICCAI 2022: 25th International Conference, Singapore, September 18--22, 2022, Proceedings, Part III}, pages 655--664. Springer, 2022{\natexlab{e}}.

\bibitem[Wang et~al.(2023{\natexlab{d}})Wang, Liu, Wang, and Zhou]{10wang2023metransformer}
Zhanyu Wang, Lingqiao Liu, Lei Wang, and Luping Zhou.
\newblock Metransformer: Radiology report generation by transformer with multiple learnable expert tokens.
\newblock In \emph{Proceedings of the IEEE/CVF Conference on Computer Vision and Pattern Recognition}, pages 11558--11567, 2023{\natexlab{d}}.

\bibitem[Wang et~al.(2023{\natexlab{e}})Wang, Liu, Wang, and Zhou]{120wang2023r2gengpt}
Zhanyu Wang, Lingqiao Liu, Lei Wang, and Luping Zhou.
\newblock R2gengpt: Radiology report generation with frozen llms.
\newblock \emph{Meta-Radiology}, 1\penalty0 (3):\penalty0 100033, 2023{\natexlab{e}}.

\bibitem[Williams(1992)]{williams1992simple}
Ronald~J Williams.
\newblock Simple statistical gradient-following algorithms for connectionist reinforcement learning.
\newblock \emph{Machine learning}, 8:\penalty0 229--256, 1992.

\bibitem[Wu et~al.(2023{\natexlab{a}})Wu, Zhang, Zhang, Wang, and Xie]{wu2023generalistfoundationmodelradiology}
Chaoyi Wu, Xiaoman Zhang, Ya~Zhang, Yanfeng Wang, and Weidi Xie.
\newblock Towards generalist foundation model for radiology by leveraging web-scale 2d\&3d medical data, 2023{\natexlab{a}}.
\newblock URL \url{https://arxiv.org/abs/2308.02463}.

\bibitem[Wu et~al.(2021{\natexlab{a}})Wu, Agu, Lourentzou, Sharma, Paguio, Yao, Dee, Mitchell, Kashyap, Giovannini, et~al.]{ImaGenomewu2021chest}
Joy Wu, Nkechinyere Agu, Ismini Lourentzou, Arjun Sharma, Joseph Paguio, Jasper~Seth Yao, Edward~Christopher Dee, William Mitchell, Satyananda Kashyap, Andrea Giovannini, et~al.
\newblock Chest imagenome dataset.
\newblock \emph{Physio Net}, 2021{\natexlab{a}}.

\bibitem[Wu et~al.(2021{\natexlab{b}})Wu, Syed, Ahmad, Pillai, Gur, Jadhav, Gruhl, Kato, Moradi, and Syeda-Mahmood]{wu2021ai}
Joy~T Wu, Ali Syed, Hassan Ahmad, Anup Pillai, Yaniv Gur, Ashutosh Jadhav, Daniel Gruhl, Linda Kato, Mehdi Moradi, and Tanveer Syeda-Mahmood.
\newblock Ai accelerated human-in-the-loop structuring of radiology reports.
\newblock In \emph{AMIA Annual Symposium Proceedings}, volume 2020, page 1305, 2021{\natexlab{b}}.

\bibitem[Wu et~al.(2022)Wu, Li, Wang, and Qian]{84wu2022multimodal}
Xing Wu, Jingwen Li, Jianjia Wang, and Quan Qian.
\newblock Multimodal contrastive learning for radiology report generation.
\newblock \emph{Journal of Ambient Intelligence and Humanized Computing}, pages 1--10, 2022.

\bibitem[Wu et~al.(2023{\natexlab{b}})Wu, Huang, and Huang]{15wu2023token}
Yuexin Wu, I-Chan Huang, and Xiaolei Huang.
\newblock Token imbalance adaptation for radiology report generation.
\newblock In \emph{Conference on Health, Inference, and Learning}, pages 72--85. PMLR, 2023{\natexlab{b}}.

\bibitem[Xu et~al.(2020)Xu, Hu, Jiang, Feng, Wang, Huang, Ju, Xiao, and Zhu]{CurriculumL-2xu2020dynamic}
Chen Xu, Bojie Hu, Yufan Jiang, Kai Feng, Zeyang Wang, Shen Huang, Qi~Ju, Tong Xiao, and Jingbo Zhu.
\newblock Dynamic curriculum learning for low-resource neural machine translation.
\newblock In \emph{Proceedings of the 28th International Conference on Computational Linguistics}, pages 3977--3989, 2020.

\bibitem[Xu et~al.(2023)Xu, Zhu, Huang, Jin, Ding, Li, and Ran]{35xu2023vision}
Dexuan Xu, Huashi Zhu, Yu~Huang, Zhi Jin, Weiping Ding, Hang Li, and Menglong Ran.
\newblock Vision-knowledge fusion model for multi-domain medical report generation.
\newblock \emph{Information Fusion}, 97:\penalty0 101817, 2023.

\bibitem[Xue et~al.(2024)Xue, Tan, Tan, Qin, and Xiang]{99xue2024generating}
Youyuan Xue, Yun Tan, Ling Tan, Jiaohua Qin, and Xuyu Xiang.
\newblock Generating radiology reports via auxiliary signal guidance and a memory-driven network.
\newblock \emph{Expert Systems with Applications}, 237:\penalty0 121260, 2024.

\bibitem[Yan et~al.(2022)Yan, Pei, Zhao, Shan, and Tian]{74yan2022prior}
Bin Yan, Mingtao Pei, Meng Zhao, Caifeng Shan, and Zhaoxing Tian.
\newblock Prior guided transformer for accurate radiology reports generation.
\newblock \emph{IEEE Journal of Biomedical and Health Informatics}, 26\penalty0 (11):\penalty0 5631--5640, 2022.

\bibitem[Yan(2022)]{42yan2022memory}
Sixing Yan.
\newblock Memory-aligned knowledge graph for clinically accurate radiology image report generation.
\newblock In \emph{Proceedings of the 21st Workshop on Biomedical Language Processing}, pages 116--122, 2022.

\bibitem[Yan et~al.(2023)Yan, Zhang, Zhou, He, Li, and Sun]{multimoda2yan2023multimodal}
Zhiling Yan, Kai Zhang, Rong Zhou, Lifang He, Xiang Li, and Lichao Sun.
\newblock Multimodal chatgpt for medical applications: an experimental study of gpt-4v.
\newblock \emph{arXiv preprint arXiv:2310.19061}, 2023.

\bibitem[Yang et~al.(2021{\natexlab{a}})Yang, Niu, Wu, Wang, Liu, and Li]{43yang2021automatic}
Shaokang Yang, Jianwei Niu, Jiyan Wu, Yong Wang, Xuefeng Liu, and Qingfeng Li.
\newblock Automatic ultrasound image report generation with adaptive multimodal attention mechanism.
\newblock \emph{Neurocomputing}, 427:\penalty0 40--49, 2021{\natexlab{a}}.

\bibitem[Yang et~al.(2022)Yang, Wu, Ge, Zhou, and Xiao]{11yang2022knowledge}
Shuxin Yang, Xian Wu, Shen Ge, S~Kevin Zhou, and Li~Xiao.
\newblock Knowledge matters: Chest radiology report generation with general and specific knowledge.
\newblock \emph{Medical Image Analysis}, 80:\penalty0 102510, 2022.

\bibitem[Yang et~al.(2023)Yang, Wu, Ge, Zheng, Zhou, and Xiao]{37yang2023radiology}
Shuxin Yang, Xian Wu, Shen Ge, Zhuozhao Zheng, S~Kevin Zhou, and Li~Xiao.
\newblock Radiology report generation with a learned knowledge base and multi-modal alignment.
\newblock \emph{Medical Image Analysis}, 86:\penalty0 102798, 2023.

\bibitem[Yang et~al.(2021{\natexlab{b}})Yang, Yu, Zhang, Han, Jiang, and Huang]{9yang2021joint}
Yan Yang, Jun Yu, Jian Zhang, Weidong Han, Hanliang Jiang, and Qingming Huang.
\newblock Joint embedding of deep visual and semantic features for medical image report generation.
\newblock \emph{IEEE Transactions on Multimedia}, 2021{\natexlab{b}}.

\bibitem[You et~al.(2021)You, Liu, Ge, Xie, Zhang, and Wu]{6you2021aligntransformer}
Di~You, Fenglin Liu, Shen Ge, Xiaoxia Xie, Jing Zhang, and Xian Wu.
\newblock Aligntransformer: Hierarchical alignment of visual regions and disease tags for medical report generation.
\newblock In \emph{Medical Image Computing and Computer Assisted Intervention--MICCAI 2021: 24th International Conference, Strasbourg, France, September 27--October 1, 2021, Proceedings, Part III 24}, pages 72--82. Springer, 2021.

\bibitem[You et~al.(2022)You, Li, Okumura, and Suzuki]{50you2022jpg}
Jingyi You, Dongyuan Li, Manabu Okumura, and Kenji Suzuki.
\newblock Jpg-jointly learn to align: Automated disease prediction and radiology report generation.
\newblock In \emph{Proceedings of the 29th International Conference on Computational Linguistics}, pages 5989--6001, 2022.

\bibitem[Zhang et~al.(2023{\natexlab{a}})Zhang, Shen, Wan, Goudos, Wu, Cheng, and Zhang]{39zhang2023novel}
Junsan Zhang, Xiuxuan Shen, Shaohua Wan, Sotirios~K Goudos, Jie Wu, Ming Cheng, and Weishan Zhang.
\newblock A novel deep learning model for medical report generation by inter-intra information calibration.
\newblock \emph{IEEE Journal of Biomedical and Health Informatics}, 2023{\natexlab{a}}.

\bibitem[Zhang et~al.(2024{\natexlab{a}})Zhang, Zhou, Adhikarla, Yan, Liu, Yu, Liu, Chen, Davison, Ren, et~al.]{126zhang2024generalist}
Kai Zhang, Rong Zhou, Eashan Adhikarla, Zhiling Yan, Yixin Liu, Jun Yu, Zhengliang Liu, Xun Chen, Brian~D Davison, Hui Ren, et~al.
\newblock A generalist vision--language foundation model for diverse biomedical tasks.
\newblock \emph{Nature Medicine}, pages 1--13, 2024{\natexlab{a}}.

\bibitem[Zhang et~al.(2023{\natexlab{b}})Zhang, Jiang, Zhang, Huang, Fan, Yu, and Han]{22zhang2023semi}
Ke~Zhang, Hanliang Jiang, Jian Zhang, Qingming Huang, Jianping Fan, Jun Yu, and Weidong Han.
\newblock Semi-supervised medical report generation via graph-guided hybrid feature consistency.
\newblock \emph{IEEE Transactions on Multimedia}, 2023{\natexlab{b}}.

\bibitem[Zhang et~al.(2023{\natexlab{c}})Zhang, Xu, Usuyama, Bagga, Tinn, Preston, Rao, Wei, Valluri, Wong, et~al.]{zhang2023large}
Sheng Zhang, Yanbo Xu, Naoto Usuyama, Jaspreet Bagga, Robert Tinn, Sam Preston, Rajesh Rao, Mu~Wei, Naveen Valluri, Cliff Wong, et~al.
\newblock Large-scale domain-specific pretraining for biomedical vision-language processing.
\newblock \emph{arXiv preprint arXiv:2303.00915}, 2\penalty0 (3):\penalty0 6, 2023{\natexlab{c}}.

\bibitem[Zhang et~al.(2019)Zhang, Kishore, Wu, Weinberger, and Artzi]{zhang2019bertscore}
Tianyi Zhang, Varsha Kishore, Felix Wu, Kilian~Q Weinberger, and Yoav Artzi.
\newblock Bertscore: Evaluating text generation with bert.
\newblock In \emph{International Conference on Learning Representations}, 2019.

\bibitem[Zhang et~al.(2024{\natexlab{b}})Zhang, Zhou, Yang, Banerjee, Acosta, Miller, Huang, and Rajpurkar]{zhang2024rexrank}
Xiaoman Zhang, Hong-Yu Zhou, Xiaoli Yang, Oishi Banerjee, Juli{\'a}n~N Acosta, Josh Miller, Ouwen Huang, and Pranav Rajpurkar.
\newblock Rexrank: A public leaderboard for ai-powered radiology report generation.
\newblock \emph{arXiv preprint arXiv:2411.15122}, 2024{\natexlab{b}}.

\bibitem[Zhang et~al.(2020)Zhang, Wang, Xu, Yu, Yuille, and Xu]{3zhang2020radiology}
Yixiao Zhang, Xiaosong Wang, Ziyue Xu, Qihang Yu, Alan Yuille, and Daguang Xu.
\newblock When radiology report generation meets knowledge graph.
\newblock In \emph{Proceedings of the AAAI Conference on Artificial Intelligence}, volume~34, pages 12910--12917, 2020.

\bibitem[Zhang et~al.(2022)Zhang, Ou, Zhang, and Deng]{70zhang2022category}
Yong Zhang, Weihua Ou, Jiacheng Zhang, and Jiaxin Deng.
\newblock Category supervised cross-modal hashing retrieval for chest x-ray and radiology reports.
\newblock \emph{Computers \& Electrical Engineering}, 98:\penalty0 107673, 2022.

\bibitem[Zhou et~al.(2016)Zhou, Khosla, Lapedriza, Oliva, and Torralba]{zhou2016learning}
Bolei Zhou, Aditya Khosla, Agata Lapedriza, Aude Oliva, and Antonio Torralba.
\newblock Learning deep features for discriminative localization.
\newblock In \emph{Proceedings of the IEEE conference on computer vision and pattern recognition}, pages 2921--2929, 2016.

\bibitem[Zhou et~al.(2024)Zhou, Adithan, Acosta, Topol, and Rajpurkar]{zhou2024generalistlearnermultifacetedmedical}
Hong-Yu Zhou, Subathra Adithan, Julián~Nicolás Acosta, Eric~J. Topol, and Pranav Rajpurkar.
\newblock A generalist learner for multifaceted medical image interpretation, 2024.
\newblock URL \url{https://arxiv.org/abs/2405.07988}.

\bibitem[Zhou et~al.(2021)Zhou, Huang, Zhou, Fu, and Shao]{17zhou2021visual}
Yi~Zhou, Lei Huang, Tao Zhou, Huazhu Fu, and Ling Shao.
\newblock Visual-textual attentive semantic consistency for medical report generation.
\newblock In \emph{Proceedings of the IEEE/CVF International Conference on Computer Vision}, pages 3985--3994, 2021.

\bibitem[Zhu et~al.(2017)Zhu, Park, Isola, and Efros]{zhu2017unpaired}
Jun-Yan Zhu, Taesung Park, Phillip Isola, and Alexei~A Efros.
\newblock Unpaired image-to-image translation using cycle-consistent adversarial networks.
\newblock In \emph{Proceedings of the IEEE international conference on computer vision}, pages 2223--2232, 2017.

\end{thebibliography}

\newpage
\section*{Appendix A}

\setcounter{table}{0}
\renewcommand{\thetable}{A\arabic{table}}
{
\footnotesize
\setlength\LTcapwidth{\linewidth}
    \begin{longtable}{|p{0.6cm}|p{1cm}|p{1.1cm}|p{0.8cm}|p{0.8cm}|p{0.8cm}|p{0.8cm}|p{0.6cm}|p{0.8cm}|p{1cm}|p{1cm}|p{0.4cm}|p{1cm}|p{0.8cm}|}
    \caption{Summary of papers in the survey. The overview is based on the findings of the survey. The dataset and metrics section focuses on mainstream datasets and metrics. '--' indicates that the paper either does not detail this process or does not include the key techniques summarized in this survey. The following abbreviations are used: I-Archi: the architecture of image feature learning, I-Module: the enhancement module of image feature learning, NI: the feature learning of non-image data, GEI: gastrointestinal endoscope image, RetiI: retinal image, Term: terminology, KnowB: knowledge base, RealR: real report, ClinicalI: clinical information, InsT: instructions, FreFilter: frequency-based filtering, ConAtoL: converting all tokens to lowercase, RemoNAT: removing non-alphabetic tokens, DenN: DenseNet, IncepV3: InceptionV3, EffNet: EfficientNet, TranF: Transformer, AT: auxiliary task, ContL: Contrastive learning, ContM: Contrastive module, Embed: general embedding techniques, MM: memory metric, GraA: graph attention, AttenT: attention, FeatO: feature-level operation, OptimS: optimization strategies, H-LSTM: hierarchical LSTM, ReLoss: re-weighted loss function, ReinL: reinforcement learning, R-L: Rouge-L, MET: METEOR, C-D: CIDEr-D, CE: clinical efficacy, R\&T: retrieval-based and template-based techniques, RGF1: RadGraphF1, Com: comparison, Clas: classification, Escore: error scoring, ECI: extracting case-related information, IU: IU-Xray, MIMIC: MIMIC-CXR, ImaGeno: Chest ImaGenome, COV: COV-CTR, DeepEye: DeepEyeNet, CCT: COVID-19 CT, Ret-I: Retina ImBank, and Ret-C: Retina Chinese. In addition, AT-Graph, AT-Class, AT-EC, and AT-DS mean the graph-based, classification, embedding comparison, and detection/segmentation auxiliary tasks.}\label{tab:overall}\\
    \hline \multirow{3}*{Paper} & \multirow{3}*{Input Data}& \multirow{3}*{\shortstack[l]{Data \\Preparation}} & \multicolumn{5}{c|}{Feature Learning}& \multirow{3}*{\shortstack[l]{Feature\\Fusion}}& \multicolumn{3}{c|}{Generation}&\multirow{3}*{Datasets}& \multirow{3}*{Metrics}\\
         \cline{4-8}\cline{10-12}
         ~ & ~ & ~ & \multicolumn{3}{c|}{Image}&\multicolumn{2}{c|}{Non-image}&~&\multicolumn{2}{c|}{Decoder-based method}&\multirow{2}*{R\&T}&~&~\\ 
         \cline{4-8}\cline{10-11}
         ~ & ~ & ~ &Archi&A\_L&E\_M&Archi&E\_M&~&Archi&T\_S&~&~&\\ \hline
         
	\endfirsthead	
 
     \multicolumn{11}{c}%
    {{\bfseries \tablename\ \thetable{} -- continued from previous page}} \\
        \hline \multirow{3}*{Paper} & \multirow{3}*{Input Data}& \multirow{3}*{\shortstack[l]{Data \\Preparation}} & \multicolumn{5}{c|}{Feature Learning}& \multirow{3}*{\shortstack[l]{Feature\\Fusion}}& \multicolumn{3}{c|}{Generation}&\multirow{3}*{Datasets}& \multirow{3}*{Metrics}\\
         \cline{4-8}\cline{10-12}
         ~ & ~ & ~ & \multicolumn{3}{c|}{Image}&\multicolumn{2}{c|}{Non-image}&~&\multicolumn{2}{c|}{Decoder-based method}&\multirow{2}*{R\&T}&~&~\\ 
         \cline{4-8}\cline{10-11}
         ~ & ~ & ~ &Archi&A\_L&E\_M&Archi&E\_M&~&Archi&T\_S&~&~&\\ \hline

	\endhead
        \hline \multicolumn{14}{|r|}{{Continued on next page}} \\ \hline
	\endfoot
        
	\endlastfoot
        \multicolumn{14}{|c|}{Traditional deep learning methods}\\
        \hline 
         \citet{1liu2021exploring}&Chest X-ray,\newline Term,\newline RealR&Tokenizing,\newline ConAtoL,\newline RemoNAT,\newline FreFilter&ResNet&AT-Graph&--&Embed,\newline TranF&--&FeatO&TranF&--&--&IU,\newline MIMIC&BLEU,\newline R-L,\newline MET,\newline C-D\\
         \hline
         \citet{2liu2021auto}&Chest X-ray&--&ResNet&AT-Graph&--&--&--&--&TranF&--&--&IU,\newline MIMIC&BLEU,\newline R-L,\newline MET,\newline CE,\newline Com\\
         \hline
         \citet{4liu2021competence}&Chest X-ray&Tokenizing,\newline ConAtoL,\newline RemoNAT&CNNs&--&--&--&--&--&LSTMs&Curriculum learning&--&IU,\newline MIMIC&BLEU,\newline R-L,\newline MET,\newline Com\\
         \hline
         \citet{5chen2021cross}&Chest X-ray&--&ResNet+\newline TranF&--&--&--&--&MM&TranF&--&--&IU,\newline MIMIC&BLEU,\newline R-L,\newline MET,\newline CE\\
         \hline
         \citet{6you2021aligntransformer}&Chest X-ray&--&ResNet&--&--&--&--&AttenT&TranF&--&--&IU,\newline MIMIC&BLEU,\newline R-L,\newline MET,\newline Com\\
         \hline
         \citet{7miura2021improving}&Chest X-ray&--&DenN+\newline TranF&--&--&--&--&--&TranF&ReinL&--&MIMIC&BLEU,\newline C-D,\newline CE,\newline Com\\
         \hline
         \citet{8alfarghaly2021automated}&Chest X-ray&Resizing&DenN&AT-Class&--&--&--&AttenT&TranF&--&--&IU&BLEU,\newline R-L, \newline MET,\newline C-D,\newline Clas\\
         \hline
         \citet{9yang2021joint}&Chest X-ray&--&ResNet&AT-EC&--&--&--&--&H-LSTM&--&--&IU,\newline MIMIC&BLEU,\newline R-L,\newline C-D\\
         \hline
         \citet{16pahwa2021medskip}&Chest X-ray&Resizing, \newline Cropping, \newline Flipping&HRNet+\newline TranF&--&--&--&--&--&R2Gen&--&--&IU&BLEU,\newline R-L,\newline MET\\
         \hline
         \citet{17zhou2021visual}&Chest X-ray,\newline ClinicalI&Tokenizing,\newline FreFilter&DenN&AT-Class,\newline AT-EC &--&One-hot, \newline Embed&--&AttenT&H-LSTM&--&--&IU,\newline MIMIC&BLEU,\newline R-L,\newline MET,\newline C-D,\newline nKTD\\
         \hline
         \citet{18huang2021deepopht}&RetiI, \newline Term&--&CNNs&--&--&Embed&--&FeatO&LSTM&--&--&DeepEye&BLEU,\newline R-L,\newline C-D\\
         \hline
         \citet{19han2021unifying}&Spine MRI&--&Self-design &AT-DS&--&--&--&--&Reasoning&--&--&--&--\\
         \hline
         \citet{24liu2021medical}&Chest X-ray,\newline Chest CT,\newline Term&--&DenN&AT-Class&--&Embed,\newline BERT&--&FeatO,\newline AttenT&TranF&AT&--&CCT &BLEU,\newline R-L,\newline C-D,\newline Com\\
         \hline
         \citet{25wang2021self}&Chest X-ray,\newline Chest CT&Tokenizing&ResNet+\newline TranF&AT-EC&--&--&--&--&H-LSTM&AT&--&IU,\newline COV&BLEU,\newline R-L,\newline C-D\\
         \hline
         \citet{27endo2021retrieval}&Chest X-ray&--&ResNet&--&--&--&--&--&--&--&R&MIMIC&BLEU,\newline $S_{emb}$,\newline CE\\
         \hline
         \citet{29najdenkoska2021variational}&Chest X-ray&Resizing,\newline Tokenizing,\newline ConAtoL,\newline RemoNAT,\newline FreFilter&DenN+\newline TranF&AT-EC&--&--&--&--&LSTM&--&--&IU,\newline MIMIC&BLEU,\newline R-L,\newline MET,\newline CE\\
         \hline
         \citet{43yang2021automatic}&Breast \newline ultrasound&Tokenizing,\newline FreFilter&ResNet&AT-Class&--&--&--&FeatO&LSTM&--&--&BCD2018&BLEU,\newline R-L,\newline MET,\newline C-D\\
         \hline
         \citet{45pandey2021explains}&Chest X-ray&Resizing&VGG&--&--&--&--&--&H-LSTM&AT&--&IU&BLEU,\newline R-L\\
         \hline
         \citet{47hou2021automatic}&Chest X-ray&Resizing,\newline Tokenizing,\newline FreFilter&ResNet&AT-Class&--&--&--&AttenT &H-LSTM&ReinL&--&IU,\newline MIMIC&BLEU,\newline R-L,\newline MET,\newline C-D,\newline CE\\
         \hline
         \citet{49pino2021clinically}&Chest X-ray&--&DenN&--&--&--&--&--&--&--&T&IU,\newline MIMIC&BLEU,\newline R-L,\newline C-D,\newline CE,\newline MIRQI\\
         \hline
         \citet{52ma2021contrastive}&Chest X-ray&Tokenizing,\newline ConAtoL,\newline FreFilter&ResNet&--&ContM&--&--&--&H-LSTM&--&--&IU,\newline MIMIC&BLEU,\newline R-L,\newline MET,\newline CE,\newline Com\\
         \hline
         \citet{53jia2021radiology}&Chest X-ray&--&ResNet+\newline TranF&--&--&--&--&--&TranF&--&--&IU,\newline MIMIC&BLEU,\newline R-L\\
         \hline
         \citet{63hou2021ratchet}&Chest X-ray&Resizing,\newline Data\newline augmentation&DenN&--&--&--&--&--&TranF&--&--&MIMIC&BLEU,\newline R-L,\newline MET,\newline CE\\
         \hline
         \citet{71huang2021deep}&RetiI, \newline Term&Tokenizing,\newline ConAtoL,\newline RemoNAT,\newline FreFilter&CNNs&--&--&Embed&--&LSTM &LSTM&--&--&DeepEye&BLEU,\newline R-L,\newline C-D\\
         \hline
         \citet{76babar2021encoder}&Chest X-ray&Tokenizing,\newline ConAtoL,\newline RemoNAT&--&--&--&--&--&--&\multicolumn{3}{c|}{Unconditional generation}&IU,\newline MIMIC&BLEU,\newline R-L,\newline MET,\newline C-D,\newline CE\\
         \hline
         \citet{87singh2021show}&Chest X-ray&Resizing,\newline Tokenizing,\newline ConAtoL,\newline RemoNAT,\newline FreFilter&IncepV3&AT-Class&--&--&--&--&LSTM&--&--&IU,\newline MIMIC&BLEU,\newline R-L,\newline MET,\newline C-D\\
         \hline
         \citet{11yang2022knowledge}&Chest X-ray,\newline KnowB,\newline RealR&Resizing,\newline Tokenizing,\newline ConAtoL,\newline FreFilter,\newline ECI&ResNet&--&--&Embed,\newline RotatE, \newline BERT&--&AttenT&TranF&--&--&IU,\newline MIMIC&BLEU,\newline R-L,\newline C-D,\newline CE\\
         \hline
         \citet{12ramesh2022improving}&Chest X-ray&Tokenizing, Filtering&ResNet/\newline TranF&--&--&--&--&--&--&--&R&MIMIC&$S_{emb}$, \newline RGF1\\
         \hline
         \citet{23sirshar2022attention}&Chest X-ray&Tokenizing,\newline RemoNAT,\newline ConAtoL&VGG&--&--&--&--&--&LSTM&--&--&IU,\newline MIMIC&BLEU\\
         \hline
         \citet{29_2najdenkoska2022uncertainty}&Chest X-ray&Resizing,\newline RemoNAT,\newline FreFilter&DenN+\newline TranF&AT-EC&--&--&--&--&TranF, \newline LSTM&--&--&IU,\newline MIMIC&BLEU,\newline R-L,\newline MET,\newline \%Novel,\newline CE\\
         \hline
         \citet{30wang2022prior}&Chest X-ray&Cropping,\newline Tokenizing,\newline ConAtoL,\newline FreFilter&DenN&AT-Graph&--&--&--&--&H-LSTM&--&--&IU&BLEU,\newline R-L,\newline C-D\\
         \hline
         \citet{31li2022cross}&RetiI&Resizing,\newline Tokenizing,\newline ConAtoL,\newline FreFilter&I3D+\newline TranF&AT-Graph&--&--&--&--&TranF&--&--&FFA-IR &BLEU,\newline R-L,\newline MET,\newline C-D,\newline Com\\
         \hline
         \citet{32qin2022reinforced}&Chest X-ray&--&ResNet+\newline TranF&--&--&--&--&MM&TranF&ReinL&--&IU,\newline MIMIC&BLEU,\newline R-L,\newline MET,\newline CE,\newline Com\\
         \hline
         \citet{33wang2022cross}&Chest X-ray&Resizing,\newline Cropping&ResNet+\newline TranF&ContL&--&--&--&MM&TranF&--&--&IU,\newline MIMIC&BLEU,\newline R-L,\newline MET,\newline C-D\\
         \hline
         \citet{34cao2022kdtnet}&Chest X-ray,\newline GEI, \newline Term&--&DenN+\newline TranF&AT-Graph&--&Embed,\newline BERT&--&FeatO,\newline AttenT&TranF&--&--&IU&BLEU,\newline R-L,\newline C-D,\newline Com\\
         \hline
         \citet{38mohsan2022vision}&Chest X-ray&Tokenizing,\newline ConAtoL&TranF&--&--&--&--&--&TranF&--&--&IU&BLEU,\newline R-L,\newline MET,\newline C-D\\
         \hline
         \citet{42yan2022memory}&Chest X-ray&--&DenN+\newline TranF&AT-Graph&MM&--&--&--&TranF&--&--&IU,\newline MIMIC&BLEU,\newline R-L,\newline MET,\newline C-D,\newline CE,\newline MIRQI\\
         \hline
         \citet{46li2022self}&Chest X-ray&--&ResNet+\newline TranF&AT-Class&--&--&--&--&TranF&AT&--&IU&BLEU,\newline R-L,\newline MET\\
         \hline
         \citet{48sun2022lesion}&RetiI&Resizing&ResNet+\newline Faster-RCNN&AT-EC,\newline AT-DS&--&--&--&--&TranF&--&--&FFA-IR &BLEU,\newline R-L,\newline MET,\newline C-D\\
         \hline
         \citet{50you2022jpg}&Chest X-ray&Tokenizing,\newline ConAtoL,\newline RemoNAT&ResNet+\newline TranF&AT-Class&--&--&--&AttenT,\newline MM&TranF&--&--&IU&BLEU,\newline R-L,\newline MET\\
         \hline
         \citet{51wang2022automated}&Chest X-ray&Resizing&TranF&AT-Class,\newline AT-EC&--&--&--&--&TranF&ReLoss&--&IU,\newline MIMIC&BLEU,\newline R-L,\newline C-D\\
         \hline
         \citet{56lee2022cross}&Chest X-ray&Resizing,\newline Cropping,\newline Flipping&ResNet+\newline TranF&--&--&--&--&--&R2Gen&--&--&IU&BLEU,\newline R-L,\newline MET\\
         \hline
         \citet{61moon2022multi}&Chest X-ray&Resizing,\newline Cropping, \newline Tokenizing&ResNet+\newline TranF&AT&--&--&--&--&TranF&--&--&IU,\newline MIMIC&BLEU,\newline CE\\
         \hline
         \citet{62huang2022non}&RetiI, \newline Term&Resizing,\newline Tokenizing,\newline ConAtoL,\newline RemoNAT,\newline FreFilter&CNNs&--&--&Embed&--&AttenT&LSTM&--&--&DeepEye&BLEU,\newline R-L,\newline C-D\\
         \hline
         \citet{64kaur2022radiobert}&Chest X-ray&Resizing, \newline Convert image to\newline grayscale, \newline Flipping,\newline Tokenizing,\newline ConAtoL,\newline RemoNAT&VGG&--&--&--&--&--&H-LSTM&--&--&IU&BLEU,\newline R-L,\newline C-D\\
         \hline
         \citet{66song2022cross}&Chest X-ray,\newline RealR&--&DenN&--&ContM&--&--&FeatO, \newline AttenT&TranF&--&--&IU,\newline MIMIC&BLEU,\newline R-L,\newline MET,\newline CE\\
         \hline
         \citet{69jia2022few}&Chest X-ray&--&DenN&--&--&--&--&AttenT&H-LSTM&--&--&IU,\newline MIMIC&BLEU,\newline R-L\\
         \hline
         \citet{70zhang2022category}&Chest X-ray&Resizing&VGG&--&--&--&--&--&--&--&R&MIMIC&Precision\\
         \hline
         \citet{72dalla2022multimodal}&Chest X-ray,\newline ClinicalI&Tokenizing,\newline Resizing,\newline Flipping,\newline Rotation,\newline Cropping&ResNet+\newline TranF&AT-Graph&--&Embed&--&AttenT&TranF&--&--&MIMIC&BLEU,\newline R-L,\newline MET,\newline CE,\newline Com\\
         \hline
         \citet{74yan2022prior}&Chest X-ray&Tokenizing,\newline ConAtoL,\newline FreFilter&ResNet+\newline TranF&AT&--&--&--&AttenT&TranF&--&--&IU,\newline MIMIC&BLEU,\newline R-L,\newline MET,\newline C-D,\newline CE\\
         \hline
         \citet{80gajbhiye2022translating}&Chest X-ray&Tokenizing,\newline Removing irrelevant elements, ConAtoL,\newline FreFilter&DenN&AT-Class&--&--&--&--&LSTM&ReLoss&--&IU&BLEU,\newline R-L,\newline MET,\newline C-D\\
         \hline
         \citet{82wang2022inclusive}&Chest X-ray&Grouping&ResNet+\newline TranF&AT-Class&--&--&--&--&R2Gen&--&--&IU,\newline MIMIC&BLEU,\newline R-L,\newline MET\\
         \hline
         \citet{83abela2022automated}&Chest X-ray&--&DenN&--&--&--&--&--&--&--&T&MIMIC&BLEU,\newline CE\\
         \hline
         \citet{84wu2022multimodal}&Chest X-ray&--&ResNet&ContL&--&--&--&--&LSTM&--&--&IU,\newline MIMIC&BLEU,\newline R-L,\newline MET,\newline C-D\\
         \hline
         \citet{86tanwani2022repsnet}&Chest X-ray,\newline InsT&Resizing,\newline Image transformations,\newline Tokenizing&ResNeXt&AT-Class,\newline ContL&--&Embed,\newline BERT&--&AttenT&TranF&--&--&IU&BLEU\\
         \hline
         \citet{90chen2022vmeknet}&Chest X-ray&Resizing,\newline Tokenizing,\newline ConAtoL,\newline RemoNAT&ResNet+\newline TranF&AT-EC&MM&--&--&--&R2Gen&--&--&IU&BLEU,\newline R-L,\newline MET\\
         \hline
         \citet{91kong2022transq}&Chest X-ray&Resizing&TranF&--&--&--&--&--&--&--&R&IU,\newline MIMIC&BLEU,\newline R-L,\newline MET,\newline CE\\
         \hline
         \citet{92wang2022medical}&Chest X-ray&Tokenizing,\newline ConAtoL,\newline RemoNAT,\newline FreFilter&TranF&AT-Class,\newline ContL&--&--&--&FeatO&TranF&--&--&MIMIC&BLEU,\newline R-L,\newline MET,\newline C-D\\
         \hline
         \citet{93du2022automatic}&Chest X-ray&--&ResNet&AT-Class&--&--&--&--&H-LSTM&--&--&IU&BLEU,\newline R-L,\newline MET\\
         \hline
         \citet{96kaur2022cadxreport}&Chest X-ray&Resizing,\newline Tokenizing,\newline ConAtoL,\newline RemoNAT,\newline FreFilter&VGG&AT-Class&--&--&--&FeatO&H-LSTM&ReinL&--&IU&BLEU,\newline R-L,\newline C-D\\
         \hline
         \citet{13tanida2023interactive}&Chest X-ray&Resizing,\newline Removing redundant whitespaces&ResNet+\newline Faster-RCNN&AT-DS&--&--&--&--&TranF&--&--&MIMIC,\newline ImaGeno&BLEU,\newline R-L,\newline MET,\newline C-D,\newline CE\\
         \hline
         \citet{10wang2023metransformer}&Chest X-ray&Tokenizing&TranF&AT&--&--&--&--&TranF&--&--&IU,\newline MIMIC&BLEU,\newline R-L,\newline MET,\newline C-D,\newline CE\\
         \hline
         \citet{14li2023auxiliary}&Chest X-ray,\newline Chest CT&Resizing,\newline Tokenizing,\newline  FreFilter&DenN&AT-Graph&--&--&--&AttenT&TranF&--&--&IU,\newline COV &BLEU,\newline R-L,\newline C-D,\newline Com\\
         \hline
         \citet{15wu2023token}&Chest X-ray&Tokenizing,\newline ConAtoL,\newline Removing irrelevant elements,\newline  FreFilter&ResNet+\newline TranF&--&--&--&--&--&TranF&ReinL,\newline AT&--&IU,\newline MIMIC&BLEU,\newline R-L,\newline MET\\
         \hline
         \citet{20li2023dynamic}&Chest X-ray,\newline KnowB,\newline RealR&Tokenizing, \newline ECI&TranF&AT-Class,\newline ContL&--&Embed,\newline BERT&GraA&AttenT&TranF&--&--&IU,\newline MIMIC&BLEU,\newline R-L,\newline MET,\newline C-D,\newline CE\\
         \hline
         \citet{21huang2023kiut}&Chest X-ray,\newline KnowB&Tokenizing,\newline ConAtoL,\newline RemoNAT,\newline FreFilter,\newline ECI&ResNet+\newline TranF&--&--&Embed,\newline BERT&GraA&FeatO, \newline AttenT&TranF&--&--&IU,\newline MIMIC&BLEU,\newline R-L,\newline MET,\newline CE\\
         \hline
         \citet{22zhang2023semi}&Chest X-ray&--&DenN+\newline TranF&AT-Graph,\newline AT-Class,\newline AT-EC&--&--&--&--&TranF&AT&--&IU,\newline MIMIC&BLEU,\newline R-L,\newline MET,\newline CE\\
         \hline
         \citet{35xu2023vision}&Chest X-ray,\newline Dermoscopy,\newline KnowB&ECI&DenN+\newline TranF&AT-Class&--&Embed,\newline BERT&GraA&AttenT&R2Gen&--&--&IU&BLEU,\newline R-L,\newline MET,\newline Com\\
         \hline
         \citet{37yang2023radiology}&Chest X-ray,\newline KnowB&Resizing,\newline Tokenizing,\newline ConAtoL,\newline FreFilter&ResNet&AT-Class,\newline AT-EC&--&--&--&AttenT&TranF&--&--&IU,\newline MIMIC&BLEU,\newline C-D,\newline CE\\
         \hline
         \citet{39zhang2023novel}&Chest X-ray,\newline Chest CT&Resizing&ResNet+\newline TranF&--&--&--&--&--&TranF+\newline MM&--&--&IU,\newline MIMIC,\newline COV&BLEU,\newline R-L,\newline MET\\
         \hline
         \citet{54wang2023chatcad}&Chest X-ray&--&ResNet+\newline TranF&AT-Class,\newline AT-DS&--&--&--&--&R2Gen\newline+LLM&--&--&MIMIC&CE\\
         \hline
         \citet{55lin2023contrastive}&Chest X-ray,\newline RetiI&Resizing,\newline Tokenizing,\newline FreFilter&ResNet+\newline TranF&ContL&--&--&--&--&TranF&--&--&IU,\newline MIMIC,\newline Ret-I,\newline Ret-C&BLEU,\newline R-L,\newline MET\\
         \hline
         \citet{60selivanov2023medical}&Chest X-ray&Resizing,\newline Tokenizing,\newline ConAtoL,\newline RemoNAT&DenN&--&--&--&--&--&TranF&--&--&IU,\newline MIMIC&BLEU,\newline R-L,\newline C-D,\newline CE\\
         \hline
         \citet{67cao2023cmt}&Chest X-ray,\newline GEI, \newline Term&--&DenN+\newline TranF&--&--&Embed,\newline BERT&--&FeatO, \newline AttenT, \newline MM&TranF&--&--&IU,\newline MIMIC&BLEU,\newline R-L,\newline MET,\newline C-D\\
         \hline
         \citet{68shetty2023cross}&Chest X-ray&--&Self-design&--&--&--&--&--&LSTM&--&--&IU&BLEU\\
         \hline
         \citet{101wang2023self}&Chest X-ray&-&ResNet+\newline Faster-RCNN+\newline TranF&--&--&--&--&--&R2Gen&--&--&IU,\newline MIMIC,\newline ImaGeno&BLEU,\newline R-L,\newline MET\\
         \hline
         \citet{102li2023unify}&Chest X-ray&Resizing, \newline Cropping,\newline Tokenizing,\newline FreFilter,\newline ConAtoL,\newline RemoNAT&TranF&--&--&--&--&MM\newline+Self-design&TranF&AT&--&IU,\newline MIMIC&BLEU,\newline R-L,\newline MET,\newline C-D\\
         \hline
         \citet{106liu2023observation}&Chest X-ray,\newline Term,\newline RealR&FreFilter,\newline ConAtol&DenN&ContL&--&Embed,\newline TranF&--&FeatO, \newline AttenT&TranF&--&--&IU,\newline MIMIC&BLEU,\newline R-L,\newline MET\\
         \hline
         \citet{107wang2023fine}&Chest X-ray&--&ResNet+\newline TranF&ContL&--&--&--&--&R2Gen&--&--&IU,\newline MIMIC&BLEU,\newline R-L,\newline MET,\newline C-D\\
         \hline
         \citet{108pellegrini2023rad}&Chest X-ray, \newline InsT&--&EffNet&--&--&Embed,\newline BERT&--&AttenT&\multicolumn{3}{c|}{\makecell{Classification for\\ structural report generation}}&Rad-ReStruct&CE\\
         \hline
         \citet{109dalla2023controllable}&Chest X-ray,\newline ClinicalI&Resizing, \newline Cropping,\newline Grouping&ResNet+\newline Faster-RCNN&AT-DS&--&Embed&--&AttenT&TranF&--&--&MIMIC,\newline ImaGeno&BLEU,\newline R-L,\newline MET,\newline CE\\
         \hline
         \citet{110li2023enhanced}&Chest X-ray,\newline RealR,\newline Term&Resizing,\newline ECI&ResNet+\newline TranF&--&--&Embed,\newline BERT&Rewei-ghting&AttenT&TranF&-&--&IU,\newline MIMIC&BLEU,\newline R-L,\newline MET,\newline C-D\\
         \hline
         \citet{113dalla2023finding}&Chest X-ray,\newline ClinicalI&Resizing, \newline Cropping&ResNet+\newline Faster-RCNN&AT-DS,\newline AT-Graph&--&Embed&--&AttenT&TranF&-&--&MIMIC,\newline ImaGeno&BLEU,\newline R-L,\newline MET,\newline CE\\
         \hline
         \citet{115hou2023organ}&Chest X-ray&--&ResNet+\newline TranF&AT-Class&--&--&--&--&TranF&--&--&IU,\newline MIMIC&BLEU,\newline R-L,\newline MET,\newline CE\\
         \hline
         \citet{73jeong2024multimodal}&Chest X-ray&Resizing&TranF&--&--&--&--&--&--&--&R& MIMIC&BLEU,\newline Escore, \newline RGF1\\
         \hline
         \citet{105gu2024complex}&Chest X-ray,\newline Term&Resizing, \newline Cropping,\newline Flipping,\newline FreFilter,\newline RemoNAT&ResNet&AT-DS,\newline AT&--&Embed,\newline TranF&--&FeatO, \newline AttenT&TranF&ReinL&--&IU,\newline MIMIC&BLEU,\newline R-L,\newline MET,\newline CE\\
         \hline
         \citet{104wang2024trust}&Chest X-ray,\newline Chest CT&--&ResNet+\newline TranF&AT&--&--&--&--&TranF&ReLoss&--&IU,\newline COV&BLEU,\newline R-L,\newline MET,\newline Grading\\
         \hline
         \citet{99xue2024generating}&Chest X-ray, \newline Term&-&ResNet+\newline TranF&--&--&Embed,\newline TranF&--&FeatO, \newline AttenT&TranF&-&--&IU,\newline MIMIC&BLEU,\newline R-L,\newline MET\\
         \hline
         \citet{98wang2024camanet}&Chest X-ray&Resizing, \newline Cropping&DenN+\newline TranF&AT-Class&--&--&--&--&R2Gen&AT&--&IU,\newline MIMIC&BLEU,\newline R-L,\newline MET,\newline C-D,\newline CE\\
         \hline
         \citet{112jin2024promptmrg}&Chest X-ray, \newline RealR&--&ResNet&AT-Class&--&Embed,\newline TranF&--&FeatO, \newline AttenT&TranF&--&--&IU,\newline MIMIC&BLEU,\newline R-L,\newline MET,\newline CE\\
         \hline
         \multicolumn{14}{|c|}{Large model-based methods}\\
         \hline
        \citet{118pellegrini2023radialog}&Chest X-ray, \newline InsT&--&ResNet+\newline TranF&--&--&Embed&--&FeatO, \newline AttenT&TranF&--&--&MIMIC&BLEU,\newline R-L,\newline MET,\newline CE\\
         \hline
         \citet{119lee2023llm}&Chest X-ray, \newline InsT&--&CNNs&--&--&Embed&--&FeatO, \newline AttenT&TranF&--&--&MIMIC&BLEU,\newline R-L,\newline MET,\newline C-D,\newline CE\\
         \hline
         \citet{120wang2023r2gengpt}&Chest X-ray, \newline InsT&--&TranF&--&--&Embed&--&FeatO, \newline AttenT&TranF&--&--&IU,\newline MIMIC&BLEU,\newline R-L,\newline MET,\newline C-D,\newline CE\\
         \hline
         \citet{122hyland2023maira}&Chest X-ray, \newline InsT, \newline ClinicalI&--&TranF&--&--&Embed&--&FeatO, \newline AttenT&TranF&--&--&MIMIC&BLEU, \newline R-L, \newline MET, \newline CE, \newline RGF1\\
         \hline
         \citet{123bannur2024maira}&Chest X-ray, \newline InsT, \newline ClinicalI, \newline RealR&--&TranF&--&--&Embed&--&FeatO, \newline AttenT&TranF&--&--&IU,\newline MIMIC&BLEU, \newline R-L, \newline MET, \newline CE, \newline RGF1\\
         \hline
         \citet{124chen2024chexagent}&Chest X-ray, \newline InsT&--&TranF&--&--&Embed&--&FeatO, \newline AttenT&TranF&--&--&MIMIC&RG-based, \newline GPT-4-based, \newline Grading\\
         \hline
         \citet{125guo2024llavaultra}&Ultrasound, \newline InsT&--&TranF&--&--&Embed&--&FeatO, \newline AttenT&TranF&--&--&--&--\\
         \hline
         \citet{126zhang2024generalist}&Chest X-ray, \newline InsT&--&ResNet&--&--&Embed&--&FeatO, \newline AttenT&TranF&--&--&IU,\newline MIMIC&Human-based evaluation\\
         \hline
         \citet{127li2024llava}&Chest X-ray, \newline InsT&--&TranF&--&--&Embed&--&FeatO, \newline AttenT&TranF&--&--&--&GPT-4-based\\
         \hline
         \citet{129alkhaldi2024minigpt}&Chest X-ray, \newline InsT&--&TranF&--&--&Embed&--&FeatO, \newline AttenT&TranF&--&--&MIMIC&--\\
         \hline
         \citet{wu2023generalistfoundationmodelradiology}&Chest X-ray, \newline InsT&--&TranF&--&--&Embed&--&FeatO, \newline AttenT&TranF&--&--&IU,\newline MIMIC&BLEU, \newline R-L,\newline CE\\
         \hline
         \citet{zhou2024generalistlearnermultifacetedmedical}&Chest X-ray, \newline InsT&--&TranF&--&--&Embed&--&FeatO, \newline AttenT&TranF&--&--&IU,\newline MIMIC&BLEU, CE,  RGF\\
         \hline
    \end{longtable}
    }

\begin{table*}[!t]
\caption{\label{tab:ACompare_MIMIC}Comparisons of the model performance on the MIMIC-CXR Dataset. B1, B2, B3, B4, R-L, C-D, P, R, and F represent BLEU-1, BLEU-2, BLEU-3, BLEU-4, ROUGE-L, CIDEr-D, precision, recall, and F1 score, respectively. \textcolor{red}{\textbf{The best}} and \textcolor{blue}{\textbf{second best}} results are highlighted. All values were extracted from their papers. * means large model-based methods.}
\centering
\begin{tabular}{|p{3.8cm}|p{0.9cm}p{0.9cm}p{0.9cm}p{0.9cm}|p{0.9cm}|p{1.5cm}|p{0.9cm}|p{0.9cm}|p{0.9cm}|p{0.9cm}|}
\hline
Paper & B1$\uparrow$ & B2$\uparrow$&B3$\uparrow$&B4$\uparrow$&R-L$\uparrow$&METEOR$\uparrow$&C-D$\uparrow$&P$\uparrow$&R$\uparrow$&F$\uparrow$\\
\hline
\multicolumn{11}{|c|}{Findings Section}\\
\hline
\citet{5chen2021cross}&0.353&\textcolor{blue}{\textbf{0.218}}&\textcolor{blue}{\textbf{0.148}}&0.106&0.278&0.142&--&0.334&0.275&0.278\\
\citet{49pino2021clinically}&-&-&-&-&0.185&--&\textcolor{red}{\textbf{0.238}}&\textcolor{blue}{\textbf{0.381}}&\textcolor{red}{\textbf{0.531}}&\textcolor{blue}{\textbf{0.428}}\\
\citet{66song2022cross}&0.360&\textcolor{red}{\textbf{0.227}}&\textcolor{red}{\textbf{0.156}}&0.117&0.287&0.148&-&\textcolor{red}{\textbf{0.444}}&\textcolor{blue}{\textbf{0.297}}&0.356\\
\citet{118pellegrini2023radialog}*&0.346&--&--&0.095&0.271&0.140&--&--&--&0.394\\
\citet{122hyland2023maira}*&0.392&--&--&0.142&0.289&0.333&--&--&--&0.386\\
\citet{zhou2024generalistlearnermultifacetedmedical}*&--&--&--&0.178&--&--&--&--&--&--\\
\citet{123bannur2024maira} 7B*&\textcolor{blue}{\textbf{0.465}}&--&--&\textcolor{blue}{\textbf{0.234}}&\textcolor{blue}{\textbf{0.384}}&\textcolor{blue}{\textbf{0.419}}&--&--&--&0.427\\
\citet{123bannur2024maira} 13B*&\textcolor{red}{\textbf{0.479}}&--&--&\textcolor{red}{\textbf{0.243}}&\textcolor{red}{\textbf{0.391}}&\textcolor{red}{\textbf{0.430}}&--&--&--&\textcolor{red}{\textbf{0.439}}\\
\hline
\multicolumn{11}{|c|}{Impression + Findings Section}\\
\hline
\citet{84wu2022multimodal}&0.340&0.212&0.145&0.103&0.270&\textcolor{blue}{\textbf{0.139}}&0.109&-&-&-\\
\citet{51wang2022automated}&0.351&0.223&\textcolor{blue}{\textbf{0.157}}&0.118&0.287&-&\textcolor{red}{\textbf{0.281}}&-&-&-\\
\citet{69jia2022few}&\textcolor{blue}{\textbf{0.363}}&\textcolor{blue}{\textbf{0.228}}&0.156&0.130&\textcolor{red}{\textbf{0.300}}&-&-&-&-&-\\
\citet{120wang2023r2gengpt}*&\textcolor{red}{\textbf{0.411}}&\textcolor{red}{\textbf{0.267}}&\textcolor{red}{\textbf{0.186}}&\textcolor{blue}{\textbf{0.134}}&\textcolor{blue}{\textbf{0.297}}&\textcolor{red}{\textbf{0.160}}&\textcolor{blue}{\textbf{0.269}}&\textcolor{red}{\textbf{0.392}}&\textcolor{red}{\textbf{0.387}}&\textcolor{red}{\textbf{0.389}}\\
\citet{zhou2024generalistlearnermultifacetedmedical}*&--&--&--&\textcolor{red}{\textbf{0.160}}&--&--&--&--&--&--\\
\hline
\multicolumn{11}{|c|}{Unspecified generated sections}\\
\hline
\citet{119lee2023llm}*&0.092&0.046&0.026&0.015&0.162&0.069&\textcolor{red}{\textbf{0.525}}&-&-&0.211\\
\citet{wu2023generalistfoundationmodelradiology}*&0.128&--&--&--&--&--&--&--&--&--\\
\citet{4liu2021competence}&0.344&0.217&0.140&0.097&0.218&0.133&-&-&-&-\\
\citet{52ma2021contrastive}&0.350&0.219&0.152&0.109&0.283&0.151&-&0.352&0.298&0.303\\
\citet{1liu2021exploring}&0.360&0.224&0.149&0.106&0.284&0.149&-&-&-&-\\
\citet{110li2023enhanced}&0.360&0.231&0.162&0.119&0.298&0.153&0.217&-&-&-\\
\citet{55lin2023contrastive}&0.362&0.227&0.155&0.113&0.283&0.142&-&-&-&-\\
\citet{22zhang2023semi}&0.362&0.229&0.157&0.113&0.284&0.153&-&0.380&0.342&0.335\\
\citet{11yang2022knowledge}&0.363&0.228&0.156&0.115&0.284&-&0.203&0.458&0.348&0.371\\
\citet{101wang2023self}&0.363&0.235&0.164&0.118&\textcolor{blue}{\textbf{0.301}}&0.136&-&-&-&-\\
\citet{2liu2021auto}&0.369&0.231&0.156&0.118&0.295&0.153&-&0.389&0.362&0.355\\
\citet{99xue2024generating}&0.372&0.233&0.154&0.112&0.286&0.152&-&-&-&-\\
\citet{98wang2024camanet}&0.374&0.230&0.155&0.112&0.279&0.145&0.161&\textcolor{blue}{\textbf{0.483}}&0.323&0.387\\
\citet{39zhang2023novel}&0.376&0.233&0.157&0.113&0.276&0.144&-&-&-&-\\
\citet{6you2021aligntransformer}&0.378&0.235&0.156&0.112&0.283&0.158&-&-&-&-\\
\citet{32qin2022reinforced}&0.381&0.232&0.155&0.109&0.287&0.151&-&0.342&0.294&0.292\\
\citet{15wu2023token}&0.383&0.224&0.146&0.104&0.280&0.147&-&-&-&\textcolor{red}{\textbf{0.758}}\\
\citet{37yang2023radiology}&0.386&0.237&0.157&0.111&0.274&-&0.111&0.420&0.339&0.352\\
\citet{107wang2023fine}&-&-&-&0.119&0.286&0.158&0.259&-&-&-\\
\citet{10wang2023metransformer}&0.386&0.250&0.169&0.124&0.291&0.152&0.362&0.364&0.309&0.311\\
\citet{106liu2023observation}&0.391&0.249&\textcolor{blue}{\textbf{0.172}}&\textcolor{blue}{\textbf{0.125}}&\textcolor{red}{\textbf{0.304}}&0.160&-&-&-&-\\
\citet{21huang2023kiut}&0.393&0.243&0.159&0.113&0.285&0.160&-&0.371&0.318&0.321\\
\citet{82wang2022inclusive}&0.395&0.253&0.170&0.121&0.284&0.147&-&-&-&-\\
\citet{112jin2024promptmrg}&0.398&-&-&0.112&0.268&0.157&-&\textcolor{red}{\textbf{0.501}}&\textcolor{blue}{\textbf{0.509}}&0.476\\
\citet{115hou2023organ}&0.407&0.256&0.172&0.123&0.293&0.162&--&0.416&0.418&0.385\\
\citet{92wang2022medical}&\textcolor{blue}{\textbf{0.413}}&\textcolor{red}{\textbf{0.266}}&\textcolor{red}{\textbf{0.186}}&\textcolor{red}{\textbf{0.136}}&0.298&\textcolor{red}{\textbf{0.170}}&\textcolor{blue}{\textbf{0.429}}&-&-&-\\
\citet{91kong2022transq}&\textcolor{red}{\textbf{0.423}}&\textcolor{blue}{\textbf{0.261}}&0.171&0.116&0.286&\textcolor{blue}{\textbf{0.168}}&-&0.482&\textcolor{red}{\textbf{0.563}}&\textcolor{blue}{\textbf{0.519}}\\
\hline
\end{tabular}
\end{table*}

\end{document}